\documentclass[journal]{IEEEtran}

\usepackage{moreverb,url}
\usepackage[colorlinks,bookmarksopen,bookmarksnumbered,citecolor=red,urlcolor=red]{hyperref}
\usepackage{siunitx}
\usepackage{xspace}
\usepackage{tabularx}
\usepackage{xcolor,colortbl}
\usepackage{xparse}
\usepackage{amsthm}
\usepackage{mathtools}
\usepackage{url}
\usepackage{multicol}
\usepackage{soul}
\usepackage{color}
\usepackage{times}
\usepackage{graphicx}
\usepackage{lipsum}
\usepackage{setspace}
\usepackage{epsfig}
\usepackage{amsmath}
\usepackage{bm}
\usepackage{amssymb}
\usepackage{tabularx}
\usepackage{mathtools}
\usepackage{color,soul}
\usepackage{paralist}
\usepackage{tikz}
\usepackage{float}
\usepackage{pdfpages}
\usepackage{makecell}
\usepackage{array}
\usepackage{xcolor}
\usepackage{subcaption}
\usepackage{enumerate}
\usepackage{url}
\usepackage{relsize}
\usepackage{pgfgantt}
\usepackage{amsfonts}
\usepackage{textcomp}
\usepackage{multirow,diagbox}
\usepackage{threeparttable}
\usepackage{booktabs}
\usepackage{bm}

\usepackage{etoolbox} % for '\patchcmd' macro
\patchcmd{\subequations}{{0}}{{-1}}{}{}       % decrement the equation counter
\patchcmd{\subequations}{\alph}{.\arabic}{}{} % change display format of eq. counter

\renewcommand\IEEEkeywordsname{Keywords}

\usepackage{xcolor}  % Needed for colours - load before mdframed

\newcommand{\etal}{\textit{et al}.~}

\def\secref#1{Sec.~\ref{#1}}
\def\figref#1{Fig.~\ref{#1}}
\def\tabref#1{Table~\ref{#1}}

\ExplSyntaxOn
\NewDocumentCommand{\splitverb}{v}
 {
  \group_begin:
  \use:c { verbatim@font }
  \seq_set_split:Nnn \l_tmpa_seq {} {#1}
  \seq_use:Nn \l_tmpa_seq { \hspace{0pt plus 0.1pt} }
  \group_end:
 }
\ExplSyntaxOff

\title{\LARGE \bf
DynoSAM: Open-Source Smoothing and Mapping Framework for Dynamic SLAM
}

\author{Jesse Morris\mbox{*}\thanks{\mbox{*} Corresponding author}, Yiduo~Wang, Mikolaj Kliniewski and~Viorela~Ila% <-this % stops a space
\thanks{Jesse Morris, Yiduo Wang, Mikolaj Kliniewski and Viorela Ila are with the Australian Centre For Robotics (ACFR), School of Aerospace, Mechanical and Mechatronic Engineering (AMME), University of Sydney, 2006 Sydney, Australia.
	{\tt \{jesse.morris, yiduo.wang, mikolaj.kliniewski, viorela.ila\}@sydney.edu.au}}%
}

\begin{document}

% include macris
% Types
\newcolumntype{L}[1]{>{\raggedright\let\newline\\\arraybackslash\hspace{0pt}}m{#1}}
\newcolumntype{C}[1]{>{\centering\let\newline\\\arraybackslash\hspace{0pt}}m{#1}}
\newcolumntype{R}[1]{>{\raggedleft\let\newline\\\arraybackslash\hspace{0pt}}m{#1}}
\newcommand\todos[1]{\textcolor{red}{#1}} % TODO

% Commands
\newcommand{\inv}{^{-1}}
\newcommand{\tr}{^{\!\top}}
\newcommand{\invtr}{^{-\!\top}}
\newcommand{\invtrs}{^{-\!\top\slash2}}
\newcommand{\invs}{^{-1\slash2}}
\newcommand{\argmax}{\operatornamewithlimits{argmax}}
\newcommand{\argmin}{\operatornamewithlimits{argmin}}
\newcommand{\diag}{\mathit{diag}}
\newcommand{\se}{\mathrm{se}(3)}
\newcommand{\SE}{\mathrm{SE}(3)}
\newcommand{\SO}{SO(3)}
\newcommand{\R} {{\rm I\!R}}
\newcommand{\E} {{\rm I\!E}}
\newcommand{\bl}{{\bar l}}
\newcommand{\eqdef}{\vcentcolon=}
\newcommand{\algrule}[1][.5pt]{\par\vskip.5\baselineskip\hrule height #1\par\vskip.5\baselineskip}

\newcommand{\rot}[4]{\prescript{#1}{#3}{\mathbf{R}}^{#2}_{#4}}
\newcommand{\tran}[4]{\prescript{#1}{#3}{\mathbf{t}}^{#2}_{#4}}

\newcommand{\campose}[2]{\prescript{#1}{}{\mathbf{X}}_{#2}}
\newcommand{\objpose}[2]{\prescript{#1}{}{\mathbf{L}}_{#2}}
\newcommand{\worldf}{W}
\newcommand{\cammotion}[3]{\prescript{#1}{#2}{\mathbf{T}}_{#3}}
\newcommand{\objmotion}[3]{\prescript{#1}{#2}{\mathbf{H}}_{#3}}
\newcommand{\othmotion}[4]{\prescript{#1}{#2}{\mathbf{#3}}_{#4}}
\newcommand{\objf}{L}
\newcommand{\camf}{X}
\newcommand{\imgf}{I}
\newcommand{\mpoint}[2]{\prescript{#1}{}{\mathbf{m}}_{#2}}
\newcommand{\nhpoint}[2]{\prescript{#1}{}{\tilde{\mathbf{m}}}_{#2}}
\newcommand{\ppoint}[2]{\prescript{#1}{}{\mathbf{p}}_{#2}}
\newcommand{\ipoint}[2]{\prescript{#1}{}{\mathbf{p}}_{#2}}
\newcommand{\icorre}[2]{\prescript{#1}{}{\tilde{\mathbf{p}}}_{#2}}
\newcommand{\opflow}[1]{\prescript{#1}{}{\bm{\phi}}}

\newcommand{\suchthat}{\;\ifnum\currentgrouptype=16 \middle\fi|\;}

\newcommand{\factor}[2]{\lVert {#1} \rVert^2_{\Sigma_{{#2}}}}

\newcommand{\ztwod}{\mathbf{z}_{\text{2D}}}
\newcommand{\zthreed}{\mathbf{z}_{\text{3D}}}

%Commands for comments
\newcommand{\comments}[1]{\textcolor{red}{#1}}
\newcommand{\Jesse}[1] { \textcolor{orange}{Jesse: #1}}
\newcommand{\Yiduo}[1] { \textcolor{blue}{YW: #1}}
\newcommand{\hlb}[1]{\textcolor{blue}{#1}}
\newcommand\Viorela[1]{\textcolor{red}{Viorela: #1}}
\newcommand{\Mik}[1]{\textcolor{magenta}{Mik: #1}}

% $R_k^\mathcal{I}$
\newcommand{\rgbi}{$I^R_k$~}
\newcommand{\depthi}{$I^D_k$~}
\newcommand{\flowi}{$I^F_k$~}
\newcommand{\semantici}{$I^S_k$~}

%Commands for special colours
\definecolor{niceorange}{RGB}{230,159, 0}
\definecolor{niceskyblue}{RGB}{86,180, 233}
\definecolor{nicebluishgreen}{RGB}{0,158, 115}
\definecolor{niceyellow}{RGB}{240,228, 66}
\definecolor{niceblue}{RGB}{0,114, 178}
\definecolor{nicevermillion}{RGB}{213,94, 0}
\definecolor{nicereddishpurple}{RGB}{204,121, 167}

\maketitle
\thispagestyle{empty}
\pagestyle{empty}

\begin{abstract}
Traditional Visual Simultaneous Localization and Mapping systems focus solely on static scene structures, overlooking dynamic elements in the environment.
Although effective for accurate visual odometry in complex scenarios, these methods discard crucial information about moving objects.
By incorporating this information into a Dynamic SLAM framework, the motion of dynamic entities can be estimated, enhancing navigation whilst ensuring accurate localization. 
However, the fundamental formulation of Dynamic SLAM remains an open challenge, with no consensus on the optimal approach for accurate motion estimation within a SLAM pipeline.

Therefore, we developed \textit{DynoSAM}, an open-source framework for Dynamic Objects SLAM that enables the efficient implementation, testing, and comparison of various Dynamic SLAM optimization formulations. 
We further propose a novel formulation that encodes rigid-body motion model in object pose estimation as well as an error metric agnostic to object frame definition. 
\textit{DynoSAM} integrates static and dynamic measurements into a unified optimization problem solved using factor graphs, simultaneously estimating camera poses, static scene, object motion or poses, and object structures.
We evaluate \textit{DynoSAM} across diverse simulated and real-world datasets, achieving state-of-the-art motion estimation in indoor and outdoor environments, with substantial improvements over existing systems. 
Additionally, we demonstrate \textit{DynoSAM}'s contributions to downstream applications, including 3D reconstruction of dynamic scenes and trajectory prediction, thereby showcasing potential for advancing dynamic object-aware SLAM systems.
Code is open-sourced at \url{https://github.com/ACFR-RPG/DynOSAM}
\end{abstract}

\begin{IEEEkeywords}
Dynamic SLAM, Mapping, RGBD Perception
\end{IEEEkeywords}

%%%%%%%%%%%%%%%%%%%%%%%%%%%%%%%%%%%%%%%%%%%%%%%%%%%%%%%%%%%%%%%%%%%%%%%%%%%%%%%%
%%%%%%%%%%%%%%%%%%%%%%%%%%%%%%%%%%%%%%%%%%%%%%%%%%%%%%%%%%%%%%%%%%%%%%%%%%%%%%%%%%%%%%%%%%%%%%%%%%%%%%%%%%%%%%%

\section{Introduction}
\label{sec:intro}

Simultaneous Localization and Mapping (SLAM) has been a key research area for over three decades~\cite{Rosen2021annurev}. 
Despite significant advancements, most SLAM systems are designed with the assumption of a predominantly static environment~\cite{Newcombe2011ismar,mur2017orb,Campos2021tro}. 
This limitation presents challenges in real-world scenarios, 
as dynamic objects are common and must be accounted for.

\begin{figure}[t]
    \centering
    \includegraphics[width=\linewidth]{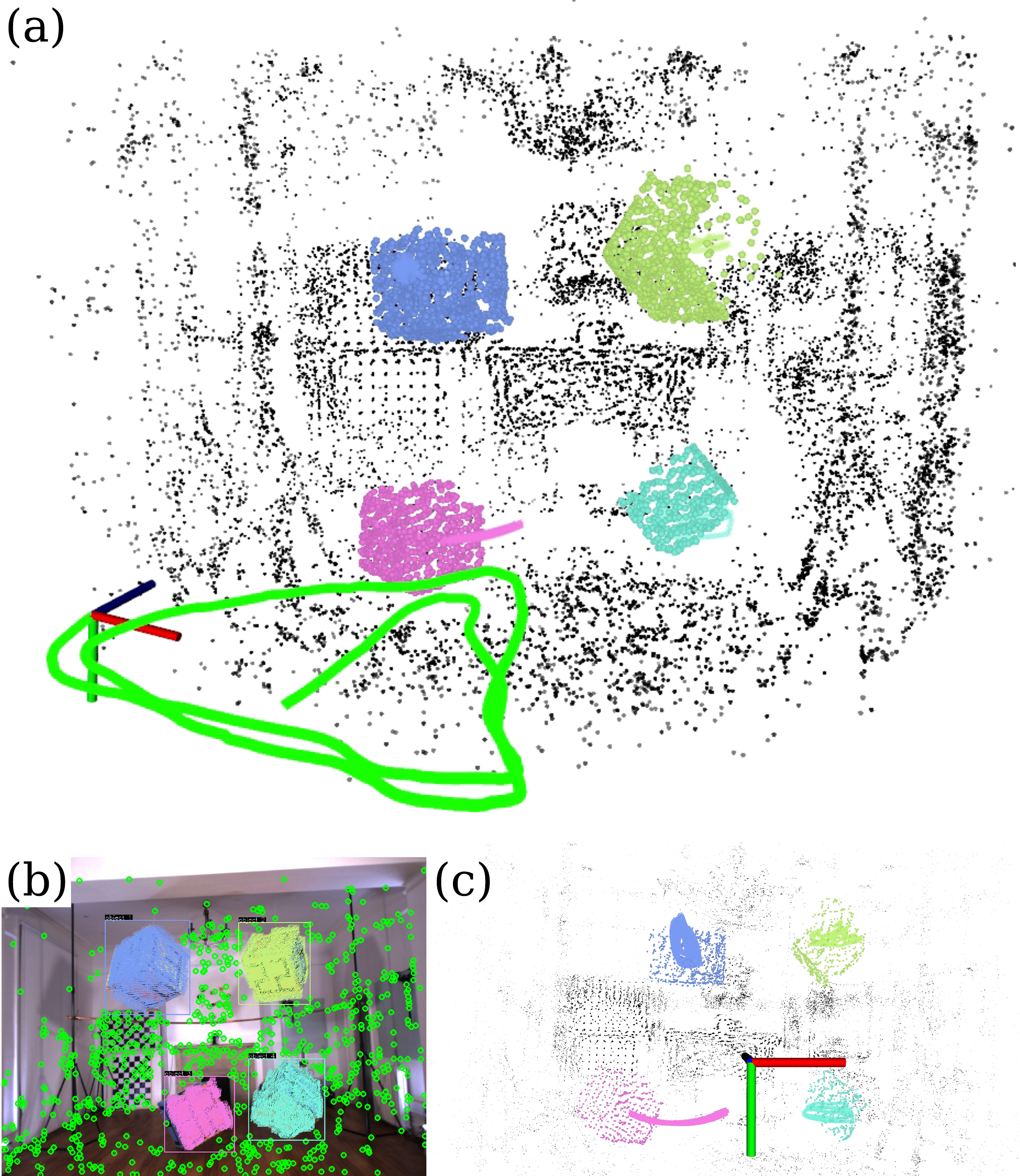}
    \caption{\footnotesize{DynoSAM is an open-source smoothing and mapping framework for Dynamic SLAM. \textbf{(a)} presents the system output, which includes camera and object trajectories, as well as the static and per-object dynamic map. \textbf{(b)} visualizes the feature-based front-end which performs multi-object tracking in addition to visual odometry. \textbf{(c)} shows the dynamic map from the camera's perspective, highlighting the estimated trajectory of each object and the tracked 3D points. }}
    \label{fig:omd_s4u_map}
    \vspace{-7mm}
\end{figure}

Traditionally, SLAM systems classify sensor measurements associated with moving objects as outliers, 
excluding such data from the estimation process~\cite{Zhao08icra,Bescos2018ral}, 
and in doing so, discard valuable information about dynamic elements. 
Incorporating these objects into the SLAM framework enables accurate modeling of dynamic environments~\cite{Newcombe2015CVPR_dynamicfusion, Wang2025icra_dynorecon}, 
which directly benefits navigation and task planning systems~\cite{Finean2021iros, Hermann2014} and enhances the overall robustness of SLAM~\cite{Henein20icra, Wang07ijrr}.

Although learning-based methods have advanced robotic capabilities in dynamic environments, 
especially in autonomous vehicle applications with access to large datasets, 
there remains a critical need for SLAM solutions in scenarios where motion models are unknown and data is scarce. 
These challenges are particularly pronounced when operating in unstructured environments~\cite{deeb2021quasistatic} or space exploration applications~\cite{setterfield2018mapping}.

As a result, there is increasing interest in extending SLAM systems to incorporate observations of dynamic entities and estimate their motions~\cite{Rosen2021annurev, zhang2020vdoslam, Liu2024ral_dynameshslam, judd2024ijrr_mvo, morris2024icra, Tian2024its_dynaquadric}. 
% In this paper we refer to such systems as \textit{Dynamic SLAM}.
In this paper we define the term \textit{Dynamic SLAM} as SLAM solutions that optimize for object motions and/or poses in addition to robot trajectory and the static structure, to distinguish from those that exclude objects to be robust in dynamic environments~\cite{Bescos2018ral, Song2022ral_dynavins, song2024dynavins++}. 
This paper focuses on studying purely Visual SLAM (vSLAM) formulations, 
though we have further discussed IMU-integrated experiments and results. 

There exists limited consensus in the community for how to best formulate the Dynamic SLAM problem. The seminal work of Henein~\etal\cite{Henein20icra} showed that 
the estimation problem can be designed around estimating for the full $\SE$ object motion. These motions are jointly optimized with the camera poses, and the static and dynamic structures, from which the per-object trajectory and velocities can be recovered. 
Their following work, VDO-SLAM~\cite{zhang2020vdoslam}, extended this formulation into a complete pipeline, but was never formally published.
The works of Judd~\etal\cite{judd2024ijrr_mvo,Judd18iros} similarly prioritize object motion, using this information to segment moving objects and track objects through occlusion.
Comparatively, many systems directly estimate for object poses~\cite{bescos2021ral, Huang2019iccv, huang2020cvpr}, typically using a constant-motion model fully constrain the estimation. 
However, downstream applications such as reconstruction~\cite{Wang2025icra_dynorecon}, navigation~\cite{Phillips2011icra_sipp} and prediction\cite{salzmann2020trajectron++} often require information on both object pose and motion.

Our prior work~\cite{morris2024icra} identified a further point of comparison between Dynamic SLAM formulations. Instead of only considering the estimation of either object pose or motion, this study termed and compared \textit{object-centric} formulations, where dynamic points and motions are represented in an object’s body frame, 
with \textit{world-centric} formulations, which represent the variables in the world-frame.
The insights from our work demonstrated that the choice of formulation plays a vital role in determining the behaviour of convergence and the resulting estimation accuracy. 
Using the world-centric motion model~\cite{Henein20icra, zhang2020vdoslam, morris2024icra} this study showed that explicit modelling of the object's rigid-body kinematics is essential for accurate results, especially when direct estimation of the object pose is required.

The inconsistencies in results and the variety of approaches, coupled with the communities growing interest in Dynamic SLAM motivated us to develop Dynamic Object Smoothing And Mapping, \textit{DynoSAM}, a framework that allows for easy and modular prototyping, implementation and evaluation of Dynamic SLAM formulations.
The suffix `-SAM' is inherited from the GTSAM~\cite{gtsam} library which our framework is built upon, thereby allowing highly customisable estimation problems to be designed and implemented.

In this work, we present a detailed theoretical foundation for Dynamic SLAM based on a world-centric model for object motion. 
Furthermore, to support downstream tasks requiring object poses, we propose a novel formulation that explicitly uses the rigid-body kinematic model for accurate, direct pose estimation.
To rigorously evaluate these formulations, 
we have integrated them into a full Dynamic vSLAM system; \figref{fig:omd_s4u_map} presents an example output of our system estimating complex dynamic object motions.
We assess the accuracy and performance of our system using a variety of indoor and outdoor datasets.
To standardise and facilitate valid comparisons between systems, we additionally propose a new error metric for evaluating object motion.
Compared with other dynamic SLAM pipelines~\cite{zhang2020vdoslam, judd2024ijrr_mvo, Huang2019iccv, bescos2021ral}, we achieve state-of-the-art accuracy in the estimation of motion and pose of objects. 
We demonstrate that our camera pose estimation is robust in highly dynamic environments and compare with AirDOS~\cite{Qiu2022icra_airdos}, DynaVINS~\cite{Song2022ral_dynavins} and ORB-SLAM3~\cite{Campos2021tro}.
Finally, we demonstrate that our formulations can directly inform downstream navigation tasks by integrating our framework with dynamic object reconstruction~\cite{Wang2025icra_dynorecon} and trajectory prediction methods.

In summary, the contribution of this paper is threefold:
\begin{itemize}
    \item \textbf{Theoretical Foundations for Object Motion Estimation.} We present a detailed theoretical formulation for representing and integrating observations of rigid bodies in motion within Dynamic SLAM. We offer a set of formulations that form the theoretical basis for practical system development, including a novel Dynamic SLAM formulation that directly parametrizes each object's pose in the scene and explicitly encodes rigid-body kinematics.
    \item \textbf{A Modular and Practical Dynamic vSLAM System.} We implement a complete Dynamic Visual SLAM system that integrates the various formulations discussed, supporting both batch and sliding-window optimizations. In addition, we provide a practical demonstration of our Dynamic SLAM pipeline with downstream tasks highly relevant for navigation, such as dynamic object reconstruction~\cite{Wang2025icra_dynorecon} and trajectory prediction. The framework is modular and extensible, allowing researchers to easily test and benchmark new formulations within a consistent evaluation setup.
    \item \textbf{Extensive Benchmarking and Novel Evaluation Metric.} We introduce an object motion evaluation metric that is agnostic to object reference frame definitions. Our framework is exhaustively evaluated on multiple public datasets, achieving state-of-the-art results. To our knowledge, this work provides one of the most comprehensive evaluations of Dynamic SLAM formulations to date.
\end{itemize}

As part of our contribution we open-source our C++ implementation of DynoSAM, 
which is integrated with the Robotic Operating System 2 (ROS2)~\cite{Thomas2014Ros2}.
We provide data logging and serialization tools for each module to facilitate evaluation and debugging, as well as an accompanying automated evaluation suite that has been used to generate all results in this paper. 
To facilitate modularity and parallelization of key components, DynoSAM's implementation is inspired by Kimera~\cite{Rosinol20icra}.

%%%%%%%%%%%%%%%%%%%%%%%%%%%%%%%%%%%%%%%%%%%%%%%%%%%%%%%%%%%%%%%%%%%%%%%%%%%%%%%%%%%%%%%%%%%%%%%%%%%%%%%%%%%%%%%

\section{Related Work}
\label{sec:related}
This section provides an overview on how SLAM systems address the challenges of dynamic environments over the years. 
We focus on state-of-the-art Dynamic SLAM systems that estimate object poses and motions, but we also include those systems that aim to improve robustness by removing or ignoring dynamic objects.

\subsection{SLAM in Dynamic Environments}
In order to handle the dynamic entities in the environment, 
conventional SLAM systems such as ORB-SLAM 3~\cite{Campos2021tro} and DynaVINS~\cite{Song2022ral_dynavins} detect any moving observations as outliers and reject them to create a global map that only contains static structures using methods such as RANSAC~\cite{fischler1981cacm}, point correlations~\cite{Dai2022pami} and robust bundle adjustment formulation~\cite{Song2022ral_dynavins, song2024dynavins++}. 
Deep learning methods have also been recently used to semantically understand the scene, 
and detect and remove dynamic objects~\cite{Bescos2018ral,hachiuma2019detectfusion,zhang2020flowfusion}. 
By taking an active approach in removing dynamic objects, these methods provide robust and accurate camera pose estimation in dynamic environments.
However, any relevant information about the motion of the objects is discarded. 
Khronos~\cite{Schmid2024rss_Khronos} operate in dynamic environments by proposing a spatio-temporal SLAM formulation that unifies short-term changes, e.g. from dynamic objects, with long-term scene changes, 
and extracts moving objects from their representation using a geometric motion detection method~\cite{Schmid2023dynablox}. 
% They extract moving objects from their representation using a geometric motion detection method~\cite{Schmid2023dynablox}.
% and are able to construct a comprehensive map in complex, changing environments. 
However, they do not utilize any estimate of object motion and rely on a dense reconstruction pipeline to detect and track objects in the scene. 

To improve the robustness of camera pose estimation in dynamic environments, many of these methods either implicitly or explicitly identify and segment out moving objects in the scene. 
While motion detection and segmentation is highly relevant for Dynamic SLAM, we highlight that our work is primarily interested in the estimation of object motions and poses, in addition to segmenting motions. 

\subsection{Dynamic SLAM}
To represent the kinematic information of these objects, 
Dynamic SLAM methods incorporate measurements of dynamic entities in the SLAM formulation in addition to the static ones.
These systems first segment dynamic observations from the static background using information such as kinematics~\cite{judd2024ijrr_mvo,Huang2019iccv} and semantics~\cite{zhang2020vdoslam, bescos2021ral, huang2020cvpr}, 
before solving for the pose or motion of these objects in addition to the camera/robot poses and the map of the environment either via joint optimization~\cite{morris2024icra, bescos2021ral} or separate estimation, such as frame-to-frame object motion tracking using sparse scene flow~\cite{Barsan2018icra}. 
% For example, Barsan~\etal\cite{Barsan2018icra} leveraged semantic and kinematic segmentation to detect dynamic objects, tracked their motions using sparse scene flow and map them separately from the static scene, 
% but they only handled frame-to-frame estimation without any joint optimization~\cite{Barsan2018icra}. 
% In this work, we focus on jointly solving object motions with the robot trajectory. 
% In recent years, the problem of SLAM within dynamic environments has been studied extensively. 
% State-of-the-art literature informs many different solutions to Dynamic SLAM. 
% Based on the estimated kinematic variables of objects, 
% there are two main categories of approaches. 
% two different solutions to represent the dynamic points, categorised by the reference frames in which these points are expressed. 

Several neural network-based methods have also been proposed in recent years~\cite{zhang2024arxiv_deflow, zhang2024arxiv_seflow, Wang2023pami, Li2023iccv} to estimate the scene flow of 3D points or voxels. 
These systems focus on estimating the motions of individual points in the form scene flow vectors. 
While they produce impressive results, 
these systems do not recover the motions or poses of dynamic objects, 
and are therefore not further discussed in this paper. 

\subsubsection{Object Pose Representation}

Within the Dynamic SLAM formulations that explicitly model the object,
the most common and intuitive representation is to directly estimate the object's pose~\cite{bescos2021ral, huang2020cvpr, Wang2007ijrr_SLAMMOT, Ballester2021icra}. 
Assuming each object is a rigid body, 
observed points on these objects are static in a body-fixed local object reference frame. 
By defining the poses of objects and in turn their local frames, 
there is an immediate advantage that each object point can be expressed in the object reference frame with a single variable in the optimization problem, 
reducing the overall number of variables in the system. 
Huang~\etal\cite{Huang2019iccv, huang2020cvpr} cluster point observations on moving objects based on temporal rigidity to identify underlying rigid bodies.
% The resulting clusters of points defines the pose of each object and the poses of the camera and each dynamic object is solved through a sliding-window optimization. 
Their method, although able to extract dynamic objects from the scene, demonstrates poor object motion accuracy. 
% which in turn defines the pose of each object so that future object poses and motions can be estimated. 
Alternatively, 
several methods model objects using simple geometric primitives such as cuboids~\cite{Yang2019tro, li2018eccv, gonzalez2022twistslam, gonzalez2023twistslam++} and ellipsoids~\cite{nicholson18cvpr} to define their poses.
DynaQuadric~\cite{Tian2024its_dynaquadric} has proposed a joint optimization framework that models dynamic objects as quadrics, explicitly defining the per-object scale and demonstrating highly accurate camera localization. 
% This unique representation allows the per-object scale to be explicitly defined, and they demonstrate highly accurate camera localization. 
However, they evaluate only their object pose errors against other quadric-based methods and neglect object motion evaluations.

When focusing on specific object types, 
BodySLAM++~\cite{Henning2023iros_bodyslam++} and DSP-SLAM~\cite{Wang20213dv_dspslam} leverage learned object shape prior models and achieve accurate object pose estimation.
Their results demonstrate the benefit of integrating learned priors into a graph-based optimization framework, 
but BodySLAM++ is restricted to only estimating human poses and DSP-SLAM provides no understanding on object dynamics. 
Assuming strong knowledge on object motion, 
TwistSLAM~\cite{gonzalez2022twistslam} uses mechanical joint constraints to restrict the degrees of freedom for object motion estimation, and estimates accurate object poses under constrained motion. 
Their results however suggest limitations in estimating complex $\SE$ object motions.
% , 
% especially for objects with more complex motion patterns.

In comparison, other methods seek to estimate the full $\SE$ motion of each object without the addition of semantic-specific constraints. 
Among these solutions, 
DynaSLAM II~\cite{bescos2021ral} takes an object-centric approach to pose and velocity estimation and reports the most accurate egomotion when compared with other approaches.
% that estimate object poses without any prior information. 
By comparison, DynaSLAM II presents poor object motion estimation, and the authors consider their use of sparse features as the main reason behind such performance.
However, our prior work~\cite{morris2024icra} shows that the formulation used by Bescos~\etal\cite{bescos2021ral} does not explicitly model the rigid-body kinematics, resulting in poor estimation accuracy even with a dense object representation. 

\subsubsection{Object Motion Representation}

An alternative formulation is to estimate object motions directly in a known reference frame, 
such as a camera frame that moves with a sliding window~~\cite{judd2024ijrr_mvo, Judd18iros, Judd19ral}, 
or a well-defined reference frame like the world frame~\cite{zhang2020vdoslam,morris2024icra, Qiu2022icra_airdos}, 
which commonly coincides with the first camera/robot pose. 
MVO~\cite{judd2024ijrr_mvo} represents the dynamic points in the camera frame, 
but models object motions in the object frame. 
They define object poses using point observations when each object is first observed, 
similar to the method of Bescos~\etal\cite{bescos2021ral}. 
% employs a sliding window in their optimisation process and reports accurate camera and object motion estimates. 
% Their formulation represents the dynamic points in the camera frame at the start of each sliding window, 
% but models object motions in the object frames that are computed using the object observations at the start of the window. 

Chirikjian~\etal\cite{Chirikjian17idetc} showed that a $\SE$ motion can be expressed in any reference frame, 
and based on this concept, 
VDO-SLAM~\cite{Henein20icra,zhang2020vdoslam, zhang20iros} proposes a factor graph formulation to represent and estimate rigid-body object motions in the world frame without the need to estimate object poses. 
AirDOS~\cite{Qiu2022icra_airdos} employs the same rigid-body motion model, 
extends it to articulated objects, e.g. humans, via a learning-based human pose estimator~\cite{Fang2017iccv_rmpe}, 
and estimates the motion of each segment of an articulated object. 
% therefore demonstrating that this approach can be extended to non-rigid bodies, e.g. humans. 
% By combining this motion model with a learning-based human pose estimator~\cite{Fang2017iccv_rmpe}, 
% they estimate for the motion of each segment of an articulated object. 
% Interestingly, t
They quantitatively demonstrate direct improvement in camera pose estimation due to the incorporation of dynamic objects.
While the study of non-rigid body motion outside the scope of this paper, 
we find such a strategy inspiring for our future works. 
The motion models employed by~\cite{zhang2020vdoslam, Qiu2022icra_airdos} are interrogated in~\cite{morris2024icra} which showed, by comparing world and object-centric formulations, that a world-centric approach better encodes rigid-body kinematic models, leading to more accurate motion estimation, more efficient convergence behaviour and often shorter computation time, despite requiring more variables.

A comprehensive review of the relevant literature highlights practical challenges, including the scarcity of open-source Dynamic SLAM libraries, systems, and datasets, which prevent rigorous comparison and evaluation. Inconsistent definitions of object reference frames further complicate pose and motion assessment. These limitations motivated the development of DynoSAM, a flexible and comprehensive framework for Dynamic SLAM with frame-agnostic error metrics for evaluation.

%%%%%%%%%%%%%%%%%%%%%%%%%%%%%%%%%%%%%%%%%%%%%%%%%%%%%%%%%%%%%%%%%%%%%%%%%%%%%%%%%%%%%%%%%%%%%%%%%%%%%%%%%%%%%%%

\section{Background}
\label{sec:background}

% This section explains the mathematical notations and background for the proposed formulations, 
% and defines the problems that the proposed formulations are trying to solve. 

\subsection{Rigid-Body Motion}
\label{sec:motion_explanation}

Understanding rigid-body motion is fundamental to Dynamic SLAM as it provides a mathematical foundation to model the motion of objects in the environment. 
Our framework builds upon this understanding to estimate the motion of objects directly using frame-to-frame observations of 3D points on rigid bodies.

A rigid body, by definition, maintains its shape and size
throughout motion; and therefore the distance between any pair of points on the object will not change.
However, while these points must move together rigidly, the apparent motion of each point depends on the location of the observer.
For instance, a point on a rotating wheel may appear stationary to an onboard observer but moving to one to a stationary observer external to the wheel. 
% For instance, a point on a rotating wheel might appear stationary to an observer positioned locally on the same wheel but will appear to be moving rapidly to a stationary observer external to the wheel. 
This example highlights the distinction between what we term as \textit{local} motion, which is perceived by a reference frame that is rigidly attached to the object, 
and the \textit{observed} motion which is perceived by an external observer.

In accordance with rigid-body kinematics, 
a reference \textit{frame} is represented by a homogeneous transformation and can be rigidly attached to a rigid body, or can be placed anywhere in the environment.
A \textit{pose} of a frame defines the $\SE$ homogeneous transform that contains the frame's position and orientation with respect to another reference frame, e.g. the observer frame.

\begin{figure}[t]
	\centering
	\includegraphics[trim={0.0cm 0.0cm 0.0cm 0cm},clip,width=1.0\columnwidth]{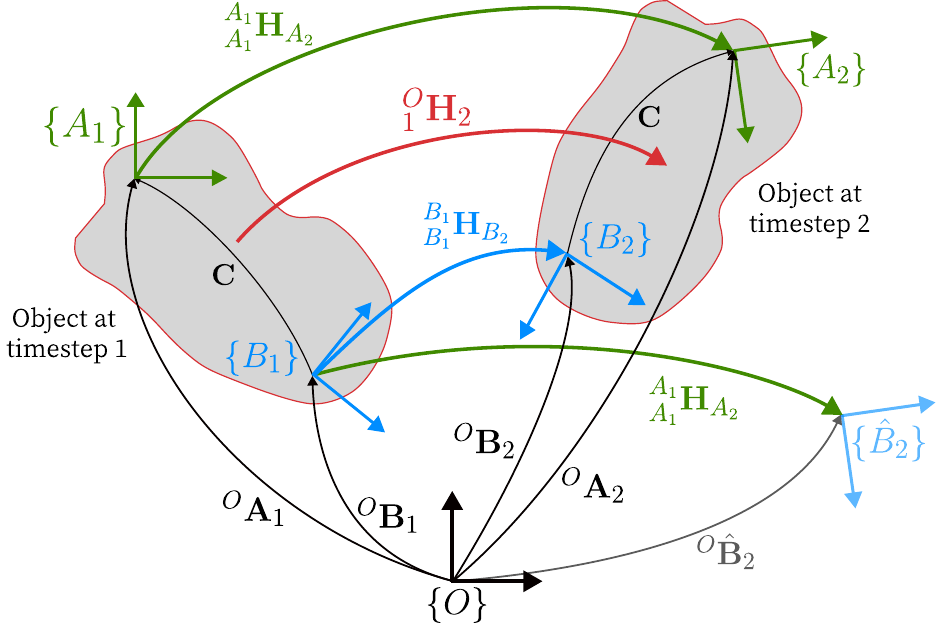}
	   \caption{\footnotesize{
       Illustrative example of two arbitrary frames, $\{A\}$ and $\{B\}$, attached to a moving rigid body.
        Solid black arrows show the poses of each frame, while green and blue arrows show the corresponding local motions of each frame.
        The red arrow denotes the observed motion $\objmotion{O}{1}{2}$ which is a unique transform that describes the rigid-body motion of all points on the object. For this reason we draw it unattached to any specific reference frame.
        $\mathbf{C}$ is a constant transform that defines the rigid-body's kinematic constraint.
       Observe that both local and observed motions can be used to correctly transform the pose of each frame in accordance the kinematic constraints. However, local motion is dependant on the chosen reference frame and incorrect use leads to the invalid propagation of $^O\mathbf{B}_1$ to ${^O\hat{\mathbf{B}}_2}$}}
    \label{fig:body_fixed_frame_example}
    \vspace{-6mm}
\end{figure}

To discuss these concepts more concretely, consider~\figref{fig:body_fixed_frame_example}, which depicts a rigid body moving freely from time-step $1$ to $2$.
Two arbitrary frames $\{A\}$ and $\{B\}$ are fixed to the object and observed by the fixed reference frame $\{O\}$. 
\mbox{$^O\mathbf{A}_1, ^O\mathbf{B}_1 \in \SE$} denote the poses of frames \mbox{$\{A_1\}$ and $\{B_1\}$ relative to $\{O\}$} prior to motion, 
while $^O\mathbf{A}_2, ^O\mathbf{B}_2 \in \SE$ denote the corresponding poses after motion, as shown by the solid black arrows in~\figref{fig:body_fixed_frame_example}.

% \Jesse{forgot to define C!}

For any moving frame, we define its \textit{motion} as a \textit{change in pose} following Chirikjian~\etal\cite{Chirikjian17idetc}.
This motion is a $\SE$ homogeneous transformation that converts one pose into another and thus describes the movement of a frame. 
Importantly, for a rigid body this motion can be represented in any arbitrary frame~\cite{Chirikjian17idetc}. 
To express this information unambiguously, we adopt the three-index notation~\cite{Chirikjian17idetc} for these motions which allows the observer frame to be explicitly identified in addition to the moving frame.
For example, $\objmotion{A_1}{A_1}{A_2}$ and $\objmotion{B_1}{B_1}{B_2}$ are the local motions for frames $\{A\}$ and $\{B\}$ respectively,
while the observed motion is $\objmotion{O}{1}{2}$.
The remainder of the section details how each local motion is depends on its frame of origin, in contrast to the observed motion which 
is independent of the moving frame and can therefore describe the rigid-body motion of \textit{any} point on the rigid-body.

\subsubsection{Local Motion}
\label{sec:local_motion}
Local motion describes the process of moving a reference frame from one pose relative to another \textit{as perceived by a frame on the moving object}. 
Local motions are shown in~\figref{fig:body_fixed_frame_example} as green and blue arrows e.g. $\objmotion{A_1}{A_1}{A_2} \in \SE$ is the transformation of pose $^O\mathbf{A}_1$ to $^O\mathbf{A}_2$ from the perspective of frame $\{A_1\}$\footnote{It is worth noting that usually in robotics $\objmotion{A}{A}{B}$ is considered, which allows for the simpler notation $\objmotion{A}{}{B}$. In this case, $\objmotion{A}{}{B}$ is the coordinate transform from $A$ to $B$ and whose matrix is the same as that of the motion $\objmotion{A}{A}{B}$.}.

These local motions form a standard kinematic chain which allows poses of frame $\{A\}$ and $\{B\}$ to be propagated such that no rigid-body kinematic constraint are violated, i.e. $\mathbf{C}$ remains constant:
\begin{equation}
\begin{aligned}
    {^O\mathbf{A}_2} &= {^O\mathbf{A}_1} \: \objmotion{A_1}{A_1}{A_2} \\
    {^O\mathbf{B}_2} &= {^O\mathbf{B}_1} \: \objmotion{B_1}{B_1}{B_2}\text{.}
    \label{equ:motion_in_local_example}
\end{aligned}
\end{equation}
Vitally, each local motion is dependent on the frame's origin and can therefore break the rigid-body constraint if applied to a pose with a different origin; this specific case is illustrated in \figref{fig:body_fixed_frame_example} where $\objmotion{A_1}{A_1}{A_2}$ is applied to $^O\mathbf{B}_1$. 
The resulting pose ${^O\hat{\mathbf{B}}_2}$ violates rigid-body constraints, as the relative transformation from ${^O\hat{\mathbf{B}}_2}$ to $^O\mathbf{A}_2$ is clearly not equal to $\mathbf{C}$.

\subsubsection{Observed Motion}
\label{sec:observed_motion}

Observed motion describes the motion of a rigid body as seen by an external observer. 
In~\figref{fig:body_fixed_frame_example}, the observed motion from time-steps $1$ to $2$ is defined with respect to $\{O\}$ and is therefore written $\objmotion{O}{1}{2}$. 
Unlike local motion in~\eqref{equ:motion_in_local_example}, this motion is invariant to the choice of body-fixed reference frame and can there be applied to any pose associated with a fixed reference on the object, whilst upholding rigid-body constraints: \mbox{$\objmotion{O}{1}{2}=\objmotion{O}{A_1}{A_2}=\objmotion{O}{B_1}{B_2}$}.
The observed motion is shown with a red arrow that, importantly, is not attached to any specific reference frame. 
This highlights that, unlike local motions, 
the observed motion does not have a geometric interpretation yet exists as a unique transform that can propagate the poses $^O\mathbf{A}_1$ and $^O\mathbf{B}_1$ to $^O\mathbf{A}_2$ and $^O\mathbf{B}_2$, as well as any other point that is rigidly attached to the object.

While these underlying principles have been previously established, 
notably in~\cite{Chirikjian17idetc} and subsequently applied to the estimation problems in~\cite{Henein20icra, zhang2020vdoslam, morris2024icra}, 
a full derivation of observed motion within the context of Dynamic SLAM has not been previously presented. 
We believe a more complete derivation is crucial for understanding the benefits of this representation and 
in this section we show how to derive $\objmotion{O}{1}{2}$ given the relations defined in~\eqref{equ:motion_in_local_example} and by exploiting the rigid-body constraint $\mathbf{C}$ between $\{A\}$ and $\{B\}$.
Each local motion can be represented with a single observed motion when expressed in a common origin frame:
\begin{subequations}
\label{equ:pose_change_to_world_example}
\begin{align}
   \objmotion{O}{1}{2} \coloneqq& {^O\mathbf{A_1}} \: \objmotion{A_1}{A_1}{A_2} \: {^O\mathbf{A_1}^{-1}} \label{equ:pose_change_a}\\ 
   =& {^O\mathbf{B_1}} \: \othmotion{B_1}{B_1}{C}{A_1} \: \objmotion{A_1}{A_1}{A_2} \: \othmotion{B_1}{B_1}{C}{A_1}^{-1} \ {^O\mathbf{B_1}^{-1}} \label{equ:pose_change_expand}\\ 
   =& {^O\mathbf{B_1}} \:\othmotion{B_1}{B_1}{C}{A_1} \: \objmotion{A_1}{A_1}{A_2} \: \othmotion{B_2}{B_2}{C}{A_2}^{-1} \ {^O\mathbf{B_1}^{-1}} \label{equ:pose_change_chain}\\ 
   =& {^O\mathbf{B_1}} \: \objmotion{B_1}{B_1}{B_2} \: {^O\mathbf{B_1}^{-1}}\text{,} \label{equ:pose_change_b}
\end{align}
\end{subequations}
where~\eqref{equ:pose_change_a} is the definition of a frame change of a pose transformation~\cite{Chirikjian17idetc}, and because $\othmotion{B}{}{C}{A} = {^O\mathbf{B}^{-1}} \: {^O\mathbf{A}}$, we arrive at~\eqref{equ:pose_change_expand}. 
Given that frames $\{A\}$ and $\{B\}$ belong to the same rigid body, their relative transformation $\othmotion{B}{}{C}{A}$, is invariant to time, 
i.e. $^{B_1}\mathbf{C}_{A_1}={^{B_2}\mathbf{C}_{A_2}}$, 
hence we can reach~\eqref{equ:pose_change_chain}, which can be re-written as~\eqref{equ:pose_change_b} following the kinematic chain visualised in~\figref{fig:body_fixed_frame_example}.

Equation~\eqref{equ:pose_change_to_world_example} shows that $\objmotion{O}{1}{2}$ remains constant regardless of the choice of $\{A\}$ and $\{B\}$ and, therefore, this representation of motion is independent of the rigid body's frame of reference.
Furthermore, using~\eqref{equ:motion_in_local_example}, we can rewrite~\eqref{equ:pose_change_to_world_example} as:
\begin{align}
   \objmotion{O}{1}{2} = {^O\mathbf{A_2}} \: {^O\mathbf{A_1}^{-1}} = {^O\mathbf{B_2}} \: {^O\mathbf{B_1}^{-1}}\text{,}
   \label{equ:generic_object_motion_defined_by_pose}
\end{align}
allowing new composition relationships to be established:
\begin{equation}
\begin{aligned}
    {^O\mathbf{A_2}} &=  \objmotion{O}{1}{2} \: {^O\mathbf{A_1}} \\
    {^O\mathbf{B_2}} &=  \objmotion{O}{1}{2} \: {^O\mathbf{B_1}}\text{.}
    \label{equ:generic_object_motion_propogation}
\end{aligned}
\end{equation}
% These equations are shown 
We highlight that the composition order of the observed motion in~\eqref{equ:generic_object_motion_propogation} is not the same as that for the local motion in~\eqref{equ:motion_in_local_example} due to their fundamental differences, 
and we direct the interested reader to~\cite{Chirikjian17idetc} for more details.

In a dynamic scene with many objects and significant occlusions, 
the distinction between local and observed motion play a crucial role. 
The camera can only observe the motions of a small set of points on each object in a busy scene and
considering only local motion makes it difficult to infer the object’s overall motion as each point experiences a unique motion.
However, if instead we expressing these points in a common external reference frame, 
\textit{the apparent motion of all points on the object can be observed directly and described using a single homogeneous transformation}, i.e. the observed motion~\cite{Chirikjian17idetc}, 
which is the foundation for our formulation of the Dynamic SLAM problem.

\subsection{Notations}
\label{sec:notations}

We define the notations for the Dynamic SLAM problem by considering a dynamic scene comprised of camera poses $\mathcal{X}$ and object poses $\mathcal{L}$:
\begin{equation*}
   \mathcal{X} = \{ \campose{\worldf}{k} \in \SE \}_{{k \in \mathcal{K}}}, \quad \mathcal{L} = \{ \objpose{\worldf}{k}^j  \in \SE \}{\substack{{j \in \mathcal{J}_k} \\ {k \in \mathcal{K}_j}}}
\end{equation*}
where $\{\worldf\}$ is the fixed world frame, $\mathcal{K}$ is the set of all time-steps and $\mathcal{J}$ is the set of all object indices. 
% We define the notations for the Dynamic SLAM problem by considering a scene at a discrete time-step $k$ with a set of objects and their associated poses with respect to a fixed world frame $\{\worldf\}$ denoted as ${^\worldf \mathcal{L}_k}$:
% \begin{equation*}
%     {^\worldf \mathcal{L}_k} = \{ \objpose{\worldf}{k}^j \in \SE,  j \in \mathcal{J}, k\in\mathcal{K}\}
%     % {^\worldf \mathcal{L}_k} = \{ \objpose{\worldf}{k}^j  \in \SE \}{\substack{{j \in \mathcal{J}_k} \\ {k \in \mathcal{K}_j}}}
% \end{equation*}
Throughout this work we liberally use the super-script $j$ or sub-script $k$ to specify the subset denoting to a particular object or at a specific time-step, e.g. $\mathcal{J}_k \in \mathcal{J}$ referring to all the observed objects at time-step $k$. 
The camera $\campose{\worldf}{k}$ and object $\objpose{\worldf}{k}$ poses define the location of frames $\{ \camf_k\} $ and $\{\objf_k\}$ with respect to $\{\worldf\}$.  
% Each object pose $\objpose{\worldf}{k}^j$ is associated with a body-fixed reference frame $\{\objf_k^j\}$. 
% Similarly, the observing camera pose is denoted as $\campose{\worldf}{k} \in \SE$ and is associated with a camera frame $\{\camf_k\}$. The set of all camera poses is denoted as:
% \begin{equation*}
%     ^\worldf \mathcal{X} =  \{ \campose{\worldf}{k} \in \SE,  k \in \mathcal{K}\}\text{.}
% \end{equation*}
% The trajectory of object $j$ can be defined using the $n+1$ consecutive frames that the object has been observed in:
% \begin{equation*}
%     {^\worldf \mathcal{L}^j} = \{ \objpose{\worldf}{k}^j \}_{k=s...s+n}\text{,}
% \end{equation*}
% where $s$ is the first time-step at which this object is observed. 
% Using this notation we can write the first pose of the object trajectory as $\objpose{\worldf}{s}^j$.

We denote $\mpoint{}{}^{i} = \left[\tilde{\mathbf{m}}^i, 1\right]^\top$ as the homogeneous coordinates of a 3D point $\tilde{\mathbf{m}}^i\in\mathbb{R}^3$, 
where $i$ is a unique tracklet index and indicates correspondences across frames. 
Any point in the world frame can be transformed into the camera frame:
\begin{equation*}
    \mpoint{\camf_k}{k}^{i} = \campose{\worldf}{k}^{-1} \: \mpoint{\worldf}{k}^{i}\text{.}
\end{equation*}
We use $\mpoint{}{} \in \mathcal{M}$ to refer to both static and dynamic points.
% A static point is specified as $\mpoint{\worldf}{}^{i}$, where the time-step $k$ is omitted, indicating that the variable is time-independent.
In the case of a static point, $\mpoint{\worldf}{k} \doteq \mpoint{\worldf}{}$ as we omit the time-step $k$ to indicate the variable is time invariant. 
To minimize notation clutter, we liberally omit indices $i$ and $j$ when there is no ambiguity.

% We define a motion as any \textit{change in pose}, following~\cite{Chirikjian17idetc}, 
% which describes the conversion of one pose into another. 
% Both poses and motions are represented using $\SE$ homogeneous transformations. 
% However, while a pose defines a body's position and orientation relative to a reference frame, 
% a motion maps one pose onto another. 
% Our work leverages this concept and employs the three-index notation~\cite{Chirikjian17idetc} for homogeneous transformations which allows both the frame of reference and the frame in motion to be identified. 
% For example, $^O_A\mathbf{H}_B \in \SE$ is the transformation which describes the motion from frame $\{A\}$ to frame $\{B\}$ from the perspective of frame $\{O\}$.

We denote motion of the camera as $\cammotion{}{}{}$ and the motion of any object $j$ as $\objmotion{}{}{}$. From now onwards we use $\objmotion{}{}{}$ to only refer to the motion of objects, rather denoting general motion as in \secref{sec:motion_explanation}.
Between time-steps $k-1$ and $k$ the local motions of the camera and object are $\cammotion{\camf_{k-1}}{\camf_{k-1}}{\camf_{k}}$ and $\objmotion{\objf_{k-1}}{\objf_{k-1}}{\objf_{k}}$ respectively. 
In the case where our notation is already explicit in denoting which pose the transformation is referring to, 
i.e., camera pose with $\cammotion{}{}{}$ or object pose with $\objmotion{}{}{}$, 
we simplify the notation by only using time-steps:
\begin{equation*}
      \cammotion{\camf_{k-1}}{k-1}{k} \doteq \cammotion{\camf_{k-1}}{\camf_{k-1}}{\camf_{k}}, \quad 
   \objmotion{\objf_{k-1}}{k-1}{k} \doteq \objmotion{\objf_{k-1}}{\objf_{k-1}}{\objf_{k}}
\label{equ:relative_camera_object_motions} 
\end{equation*}
These transformations describe local motions and, following~\eqref{equ:motion_in_local_example}, 
can be used to propagate the camera and object poses from $k-1$ to $k$:
% and can be obtained directly by decomposing camera or object poses at $k-1$ and $k$:
% \begin{align}
% \label{equ:motion_cam_in_L}
%     \cammotion{\camf_{k-1}}{k-1}{k} &= \campose{\worldf}{k-1}^{-1}\: \campose{\worldf}{k} \\
% \label{equ:motion_in_L_definition}
%     \objmotion{\objf_{k-1}}{k-1}{k} &= \objpose{\worldf}{k-1}^{-1}\: \objpose{\worldf}{k}
% \end{align}
\begin{align}
    \label{equ:motion_cam_in_L}
    \campose{\worldf}{k} &= \campose{\worldf}{k-1} \: \cammotion{\camf_{k-1}}{k-1}{k}\\
     \label{equ:motion_in_L_definition}
    \objpose{\worldf}{k} &= \objpose{\worldf}{k-1} \: \objmotion{\objf_{k-1}}{k-1}{k}
\end{align}

For the purposes of object motion estimation, 
our framework uses an observed motion representation, as discussed in~\secref{sec:observed_motion}. 
In the context of SLAM this reference frame can be any observable frame, 
e.g. the camera frame at the start of a sliding-window. 
In this work, we represent our object motions in the world frame $\{\worldf\}$, 
following the works of~\cite{Henein20icra, zhang2020vdoslam,morris2024icra}. 
Such a representation is convenient for downstream applications like mapping and planning, 
as it allows all state variables in our formulations to be defined with respect to a common coordinate system. 
For any object $j$ this motion is denoted as $\objmotion{\worldf}{k-1}{k}$. 
Due to its representation in the world frame we refer specifically to it as the \textit{absolute} motion.
We define the set of all object motions as:
% The set of motions at time-step $k$ for objects $j \in \mathcal{J}_k$ is:
\begin{equation*}
   % \othmotion{\worldf}{k-1}{\mathcal{H}}{k} = \{ \objmotion{\worldf}{k-1}{k}^j,  j \in \mathcal{J}_k, k\in\mathcal{K}\}
   \othmotion{\worldf}{k-1}{\mathcal{H}}{k} = \{ \objmotion{\worldf}{k-1}{k}^j\}{\substack{{j \in \mathcal{J}_k} \\ {k \in \mathcal{K}_j}}}
\end{equation*}

Using~\eqref{equ:generic_object_motion_defined_by_pose} we define our absolute motion in terms of a change in object pose:
\begin{equation}
    \objmotion{\worldf}{k-1}{k} = \objpose{\worldf}{k} \: \objpose{\worldf}{k-1}^{-1}
\label{equ:object_motion_from_LL_definition}
\end{equation}
%  \Jesse{this sentence: put into the notation section!}
% \Jesse{start with the motion (name and colour) and then say what it does}
% \Jesse{Fix}

% We choose $\{\worldf\}$ as our external reference frame because it allows all state variables in our formulations to be represented using a common coordinate system. To specify this motion representation (as opposed to observed motions in a different frame, or observed motions in general) we will 
% Due to its representation in the world frame we refer to this specific motion as an \textit{absolute} motion. We choose this 
% We choose to use $\{\worldf\}$ because it allows all state variables in our formulations to be represented using a common coordinate system.
Following the composition rules defined in~\eqref{equ:generic_object_motion_propogation}, $\objmotion{\worldf}{k-1}{k}$ will move $\objpose{\worldf}{k-1}$ to $\objpose{\worldf}{k}$ when applied on the left-hand side:
\begin{equation}
    \objpose{\worldf}{k} = \objmotion{\worldf}{k-1}{k} \: \objpose{\worldf}{k-1}\text{.}
\label{equ:object_pose_propogation}
\end{equation}
Vitally, this propagation also holds for any point $i$ on a rigid-body:
\begin{equation}
    \mpoint{\worldf}{k}^{i} = \objmotion{\worldf}{k-1}{k} \: \mpoint{\worldf}{k-1}^{i}
\label{equ:point_motion_in_world}\text{.}
\end{equation}
Thus, the transform $\objmotion{\worldf}{k-1}{k}$ describes the motion of \textit{any and all} points on an object from $k-1$ to $k$.
Since we can directly observe 3D object points, \eqref{equ:point_motion_in_world} allows us to easily derive cost functions in terms of all tracked points on the object.
This equation underpins all our formulations and is fundamental to our Dynamic SLAM framework.
A more detailed derivation of~\eqref{equ:point_motion_in_world} is included in Appendix~\ref{sec:app_rigid_body_on_point} which highlights how this equation implicitly encodes the rigid-body kinematic constraints discussed in~\secref{sec:motion_explanation}.

% For all notation we often omit indices $i$ and $j$ when there is no ambiguity.

% Similar to notations for points $\mpoint{}{}$ and object poses $\objpose{}{}$ in this paper, 
% we omit the index $j$ for motions when there is no ambiguity. 

\section{Foundations for Dynamic SLAM}
\label{sec:recipe_dynamic_slam}

This section details the graph-based Dynamic SLAM formulations included within our framework. 
We demonstrate how the concept of observed motion, explained in~\secref{sec:background}, is applied to directly estimate both object motions (\secref{sec:recipe_motion}) and poses (\secref{sec:recipe_pose}). 
These formulations constitute the back-end of a visual Dynamic SLAM pipeline. 
As a result, we assume dynamic objects can be tracked and 3D point measurements of static and dynamic features can be provided by a visual SLAM front-end. 
While we outline our front-end (\secref{sec:slam_frontend}) and include it in our open-source implementation, the formulations discussed in our framework are agnostic to any specific front-end implementation.

For each approach, we define residual functions $\mathbf{r}$ associated with covariance matrices $\Sigma$. 
These are used to construct factors $\phi$ in the form:
\begin{equation}
    \phi(\cdot) \propto \text{exp} \Bigl\{ -\frac{1}{2} \factor{\mathbf{r}}{} \Bigr\}
\label{equ:factor_form}
\end{equation}
Under the standard assumption of zero-mean Gaussian noise, we take the negative log of~\eqref{equ:factor_form} and drop the scaling-factor allowing us to collect all residuals and construct a nonlinear least-squares problem:
\begin{equation}
     \mathbf{\theta}^{\text{MAP}} =\underset{\mathcal{\theta}}{\mathrm{argmin}} \sum \factor{\mathbf{r}}{}
\end{equation}
For completeness, our framework also includes the object-centric formulations explored in~\cite{morris2024icra}. 
However, due to their inferior performance, this work does not discuss them further. 

\subsection{Measurements}
\label{sec:visual_measurements}
We define the following notation for visual features. 
Direct measurements are denoted as $\ztwod \in \mathbb{R}^2$ and represent 2D pixel measurements on the image plane. 
We denote 3D point measurements in the local sensor frame $\{\camf\}$ as $\zthreed \in \mathbb{R}^3$. 
In the context of stereo/RGBD SLAM, we assume the depth $d$ of each pixel measurement $\ztwod$ is available for every time-step $k$, and 3D measurements are constructed accordingly:
\begin{equation}
     \left[\zthreed, 1\right]^\top = \mpoint{\camf_k}{k} = \pi^{-1}(\ztwod, d)\text{.}
    \label{equ:3d_measurement_using_projection}
\end{equation}
where $\pi^{-1}(\cdot)$ is a sensor's back-projection function, e.g. pinhole camera model.
We further specify the notation for sets of measurements corresponding to static and dynamic entities.
$\mathcal{S}_{\text{2D}, k}$ and $\mathcal{S}_{\text{3D}, k}$ denote the set of static 2D keypoints $\ztwod$ and their corresponding 3D landmark measurements $\zthreed$ at time $k$, respectively. 
Similarly, $\mathcal{D}_{\text{2D}, k}$ and $\mathcal{D}_{\text{3D}, k}$ are the sets of dynamic measurements. 
The set of all static measurements (2D and 3D) is $\mathcal{S}_k$.
The set of dynamic measurements on a particular object is specified using the super-script $j$, e.g. $\mathcal{D}_{\text{3D}, k}^j$ is the sub-set of 3D measurements on object $j$ at time $k$. 

\subsection{World-Centric Motion Estimator} 
\label{sec:recipe_motion}
\begin{figure}[t]
	\centering
	\includegraphics[width=0.89\columnwidth]{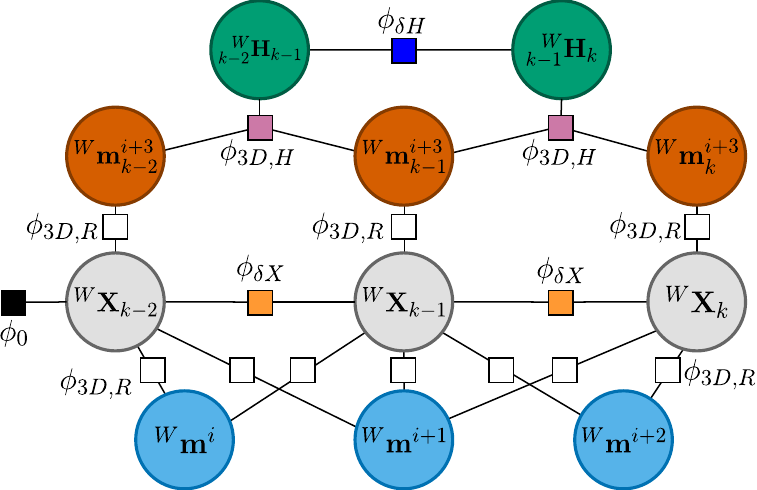}
	\caption{\footnotesize{World-centric motion formulation factor-graph. We show three static landmarks (blue) and one landmark on a dynamic object (orange), tracked over three frames. Camera poses are shown in grey and object motion in green. The \textit{point measurement factor} ($\phi_{3D, R}$) is shown in white, the \textit{ternary motion factor} ($\phi_{3D, H}$) in magenta and the \textit{object motion smoothing factor} ($\phi_{\delta H}$) in dark blue. The prior factor on the first camera pose is shown in black.} }
    \label{fig:motion_estimator_fg}
    \vspace{-6mm}
\end{figure}

The formulation for the World-Centric Motion Estimator (WCME) directly estimates object motions and their rigid-structure, as well as the static scene and camera pose:
\begin{equation}
\mathbf{\theta} \doteq [ \campose{\worldf}{k}, \  \othmotion{\worldf}{k-1}{\mathcal{H}}{k}, \  {^\worldf\mathcal{M}_k}], \: k\in \mathcal{K}\text{.}
\label{equ:world_centric_motion_states}
\end{equation}
This formulation was first proposed in~\cite{zhang2020vdoslam} and evaluated in~\cite{morris2024icra} which established it as a highly accurate method for object motion estimation. 
\figref{fig:motion_estimator_fg} shows the corresponding factor graph for this formulation and includes three static landmarks (blue) and one landmark (orange) on a dynamic object that are tracked over three frames. 

Given a 3D observation of a point (static or dynamic), 
we constrain the point to the world frame using a \textit{point measurement factor}:
\begin{equation}
     \mathbf{r}_{\text{3D}, R} = \mathbf{z}_{\text{3D}, k} - \campose{\worldf}{k}^{-1} \:  \mpoint{\worldf}{k}\text{.}
     \label{equ:proejction_3d_residual}
\end{equation}
For every static point, 
$\mpoint{\worldf}{}$ is initialized as $\mpoint{\worldf}{} = \campose{\worldf}{k} \: \mathbf{z}_{\text{3D}, k}$ when the point is first observed. 
We use the same method to initialize each dynamic point at every frame $k$.

The relative transformation between consecutive camera poses is modeled using the \textit{between factor}:
\begin{equation}
     \mathbf{r}_{\delta X}=  \left[\log\left( \cammotion{\camf_{k-1}}{k-1}{k}^{-1} \: \campose{\worldf}{k-1}^{-1} \: \campose{\worldf}{k}  \right)\right]^{\vee}\text{,}
     \label{equ:camera_between_factor}
\end{equation}
where the odometry $\cammotion{\camf_{k-1}}{k-1}{k}$ can be estimated by the front-end of a visual SLAM system, 
such as the proposed one described in~\secref{sec:slam_frontend}, 
or from alternative sources such as an Inertial Measurement Unit (IMU) if available. 
The operation $\left[\log\left(\cdot\right)\right]^{\vee}$ maps an $\SE$ transformation to $\mathbb{R}^6$. 

From \eqref{equ:point_motion_in_world} the \textit{ternary object motion factor} models the motion of any point $i$ on a rigid body $j$ as in~\cite{zhang2020vdoslam, morris2024icra}:
\begin{equation}
    \mathbf{r}_{\text{3D}, H} =  \mpoint{\worldf}{k}^{i} - \objmotion{\worldf}{k-1}{k} \: \mpoint{\worldf}{k-1}^{i}\text{,}
    \label{equ:world_landmark_motion_tenary_factor}
\end{equation}
This cost function relates a pair of tracked points on a rigid-body object with the object motion.

Finally, for each object $j$ an \emph{object smoothing factor} is introduced between consecutive motions:
\begin{equation}
\label{equ:world_motion_smoothing_factor}
    \mathbf{r}_{\delta H}=  \left[\log\left(\objmotion{\worldf}{k-2}{k-1}^{-1} \: \objmotion{\worldf}{k-1}{k}\right)\right]^{\vee}\text{.}
\end{equation}
This enforces a constant motion model represented in $\{ \worldf \}$ and prevents abrupt, drastic and unrealistic changes in object motions between consecutive frames. This constraint holds since any constant change in the body frame also leads to a constant change in the reference frame pose change, as shown in~\cite{Henein20icra};
% then the reference frame pose change is also constant; 
an extended version of this proof is included in Appendix~\ref{sec:app_motion_model}.

The nonlinear least-squares problem is constructed using the combination of these factors: 
% The combination of these factors leads to the construction of the nonlinear least-squares problem:
\begin{equation}
\begin{aligned}
     \mathbf{\theta}^{\text{MAP}} = \underset{\bm{\theta}}{\mathrm{argmin}} \biggl(\factor{\mathbf{r}_0}{0} + \: \factor{\mathbf{r}_{\delta X}}{\delta X} \\
    + \sum_{m \in \mathcal{S}_{\text{3D},k} } \rho_h \factor{\mathbf{r}_{\text{3D},R}}{{\text{3D},R}} % static points
    +  \sum_{m \in \mathcal{D}_{\text{3D},k}}\rho_h \factor{\mathbf{r}_{\text{3D},R}}{{\text{3D},R}} \\%dynamic points
    + \sum_{j \in \mathcal{J}_k} \: \sum_{m \in \mathcal{D}_{\text{3D},k}^j} \rho_h \factor{\mathbf{r}_{\text{3D}, H}}{{\text{3D}, H}}% motion factor
    + \sum_{j \in \mathcal{J}_k} \factor{\mathbf{r}_{\delta H}}{{\delta H}}\biggr)\text{,} % smoothing factor
\end{aligned}
\end{equation}
where $\mathbf{r}_0$ is prior residual on the first camera pose state variable and $\rho_h$ is any robust cost function. In our implementation, we use the Huber~\cite{Huber92bs} function.
While not being directly incorporated in this estimation, 
the trajectory of each object $ {^\worldf \mathcal{L}}$ can be recovered by recursively propagating the previous pose using the estimated motion following~\eqref{equ:object_pose_propogation}: 
\begin{equation}
    \objpose{\worldf}{k} = \objmotion{\worldf}{k-1}{k}\dots\objmotion{\worldf}{s}{s+1} \: \objpose{\worldf}{s}\text{,}
    \label{equ:rpe_expand}
\end{equation}
where $s$ is the first time-step at which this object is observed. 
This method requires an arbitrary first pose $\objpose{\worldf}{s}$ to be defined, which anchors the resulting trajectory.
Since the estimated motion is an absolute motion,
we can define this pose anywhere relative to $\{\worldf\}$.
Practically, we construct $\objpose{\worldf}{s}$ upon the initial observation of an object using the centroid of all object point observations as its position and identity as its orientation. 
In addition, any method that provides object pose, such as learned techniques, can easily be adopted.
For experiment evaluation, we use the ground truth object pose to set $\objpose{\worldf}{s}$, 
aligning the origins of estimated object trajectories and that of the ground truth to facilitates proper evaluation as discussed in~\secref{sec:metrics}.

\subsection{World-Centric Pose Estimator}
\label{sec:recipe_pose}
% While estimating $\objmotion{\worldf}{k-1}{k}$ provides a direct representation of object motions, 
% object pose $\objpose{\worldf}{k}$ is also highly relevant as it can enable prior information about the object, 
% i.e. size, shape and other semantically driven properties, 
% to be directly integrated into the estimation.

\begin{figure}[t]
	\centering
	\includegraphics[width=0.89\columnwidth]{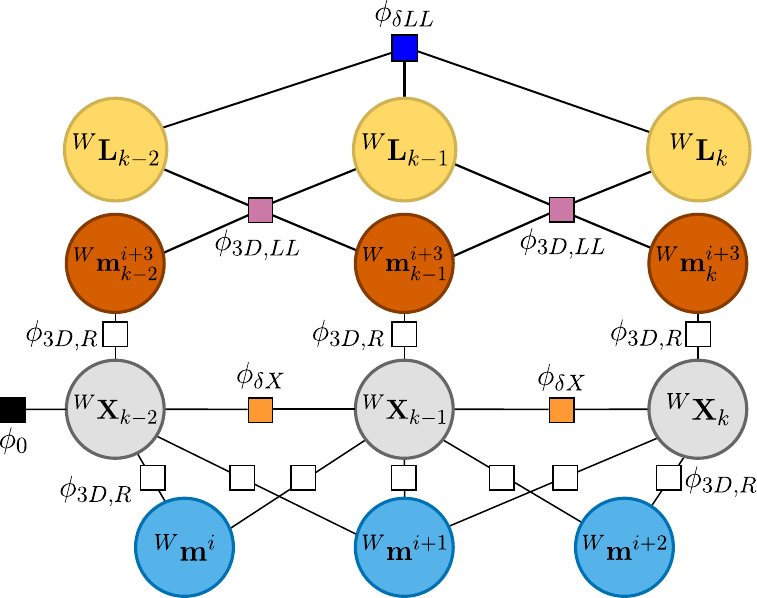}
	\caption{\footnotesize{World-centric pose formulation factor graph. As with the previous formulation (shown in~\figref{fig:motion_estimator_fg}), three static landmarks (blue) and one dynamic landmark (orange) are shown. The object pose, $\objpose{}{}$, is shown in yellow with the modified motion ($\phi_{3D, LL})$ and smoothing factor ($\phi_{\delta LL})$ in magenta and dark blue respectively.}}
    \label{fig:pose_estimator_fg}
    \vspace{-8mm}
\end{figure}

Up to this point, we have used the rigid-body motion $\objmotion{\worldf}{}{}$ to directly parametrize the object state. 
This perspective, used in both our work and prior studies, has enabled effective estimation of object motion and recovery of object pose. 
However, while $\objmotion{\worldf}{k-1}{k}$ provides valuable insight regarding object dynamics, directly estimating the object pose $\objpose{\worldf}{k}$ is highly relevant for downstream tasks.
% can offer a more practical and comprehensive understanding of the scene in many cases.
% Literature advises that object pose enables more direct reasoning about the semantic and geometric properties of objects, facilitating downstream tasks such as scene understanding, reconstruction~\cite{Wang2025icra_dynorecon} and object-centric planning~\cite{Phillips2011icra_sipp} and prediction~\cite{salzmann2020trajectron++}. 
% \Jesse{are there more citations for this...}
We therefore propose the World-Centric Pose Estimator (WCPE), a novel formulation that parametrizes the object state in terms of its pose $\objpose{\worldf}{}$. Unlike other estimators that use a motion prior~\cite{judd2024ijrr_mvo, bescos2021ral} between poses, our formulation uses the rigid-body motion directly as a constraint between poses, based on the measurement of object points.  
This approach therefore demonstrates the importance of using the absolute motion representation as the foundation for new Dynamic SLAM formulations.

WCPE estimates the states:
\begin{equation}
\mathbf{\theta} \doteq [ \campose{\worldf}{k},   ^{\worldf}\mathcal{L}_{k}, {^\worldf\mathcal{M}_k}] \: k\in \mathcal{K}
\label{equ:world_centric_pose_states}
\end{equation}
The core of this approach is as follows: equation~\eqref{equ:object_motion_from_LL_definition} shows how a motion $\objmotion{\worldf}{k-1}{k}$ explicitly models a pair of consecutive object poses.
Using this idea, we reformulate our factors to make every $\objpose{\worldf}{}$ a variable of the system.
The corresponding factor graph is shown in \figref{fig:pose_estimator_fg}.
Substituting~\eqref{equ:object_motion_from_LL_definition} into \eqref{equ:world_landmark_motion_tenary_factor} forms a new \textit{quaternary object motion factor}:
\begin{equation}
    \mathbf{r}_{\text{3D}, LL} =  \mpoint{\worldf}{k}^{i} - \objpose{\worldf}{k} \: \objpose{\worldf}{k-1}^{-1} \: \mpoint{\worldf}{k-1}^{i}
\label{equ:world_landmark_motion_pose_factor}.
\end{equation}
We similarly define the smoothing factor in-terms of object pose by substituting~\eqref{equ:object_motion_from_LL_definition} into \eqref{equ:world_motion_smoothing_factor}. This forms the \textit{object pose smoothing factor}:
\begin{equation}
    \mathbf{r}_{\delta LL} = \left[\log\left( {\left( \objpose{\worldf}{k-1} \: \objpose{\worldf}{k-2}^{-1} \right)^{-1}} \: {\left( \objpose{\worldf}{k} \: \objpose{\worldf}{k-1}^{-1} \right)} \right)\right]^{\vee}
\label{eq:world_LL_smoothing_factor}
\end{equation}
The nonlinear least-squared problem is constructed as:
\begin{align}
\begin{split}
    \mathbf{\theta}^{\text{MAP}} =\underset{\bm{\mathbf{\theta} }}{\mathrm{argmin}} \biggl(\factor{\mathbf{r}_0}{0}
    + \factor{\mathbf{r}_{\delta X}}{\delta X} \\
    + \sum_{s \in \mathcal{S}_{\text{3D},k} } \rho_h \factor{\mathbf{r}_{\text{3D},R}}{{\text{3D},R}} % static points
    +  \sum_{d \in \mathcal{D}_{\text{3D},k} }\rho_h\factor{\mathbf{r}_{\text{3D},R}}{{\text{3D},R}} \\%dynamic points
    + \sum_{j \in \mathcal{J}_k} \: \sum_{m \in \mathcal{D}_{\text{3D},k}^j}\rho_h \factor{\mathbf{r}_{\text{3D}, LL}}{{\text{3D}, LL}} % motion factor
    + \sum_{j \in \mathcal{J}_k} \factor{\mathbf{r}_{\delta LL}}{{\delta LL}} \biggr) % smoothing factor
\end{split}
\end{align}

We highlight that our formulation represents both object motions and their points in the world frame. 
This differs from existing methods that also estimate for object pose but use an object-centric representation for points~\cite{bescos2021ral, morris2024icra}. 
This small difference is vital to ensure the rigid-body motion model remains explicit and therefore the potential problems of object-centric formulations outlined by Morris~\etal\cite{morris2024icra} are avoided.
% differs from other methods which also estimate for object pose (ie. ~\cite{bescos2021ral, morris2024icra}) in that our appraoch 
% modles both object motions and their points in the world frame. 
% % We highlight that our formulation differs from other methods which also estimate for object pose (ie. ~\cite{bescos2021ral, morris2024icra}) in that our appraoch 
% % modles both object motions and their points in the world frame. 
% continues to model dynamic points, as well as their motions, directly in the world frame rather than using an object-centric representation.
% This maintains an explicit rigid-body motion model and therefore avoids the potential problems of object-centric formulations, as outlined in~\cite{morris2024icra}.
Furthermore, by continuing to base our formulation on the absolute motion representation, 
we can define $\{\objf\}$ anywhere on the object, 
as the absolute motion representation is independent of the object's reference frame and pose (\secref{sec:motion_explanation}). 
% Furthermore, by simply re-parameterizing our cost functions in terms of pose we can ensure that our formulation directly embeds the observed motion model, which allows us to define $\{\objf\}$ anywhere on the object. 
% As outlined in~\secref{sec:motion_explanation}, the absolute motion representation is independent of the object's reference frame and pose. 
This flexibility enables the integration of various initialization methods, e.g. SLAM front-ends or observation centroids, without compromising rigid-body kinematic constraints.

\section{System}
\label{sec:dynosam}

\begin{figure*}[t]
	\centering
    \includegraphics[width=\textwidth]{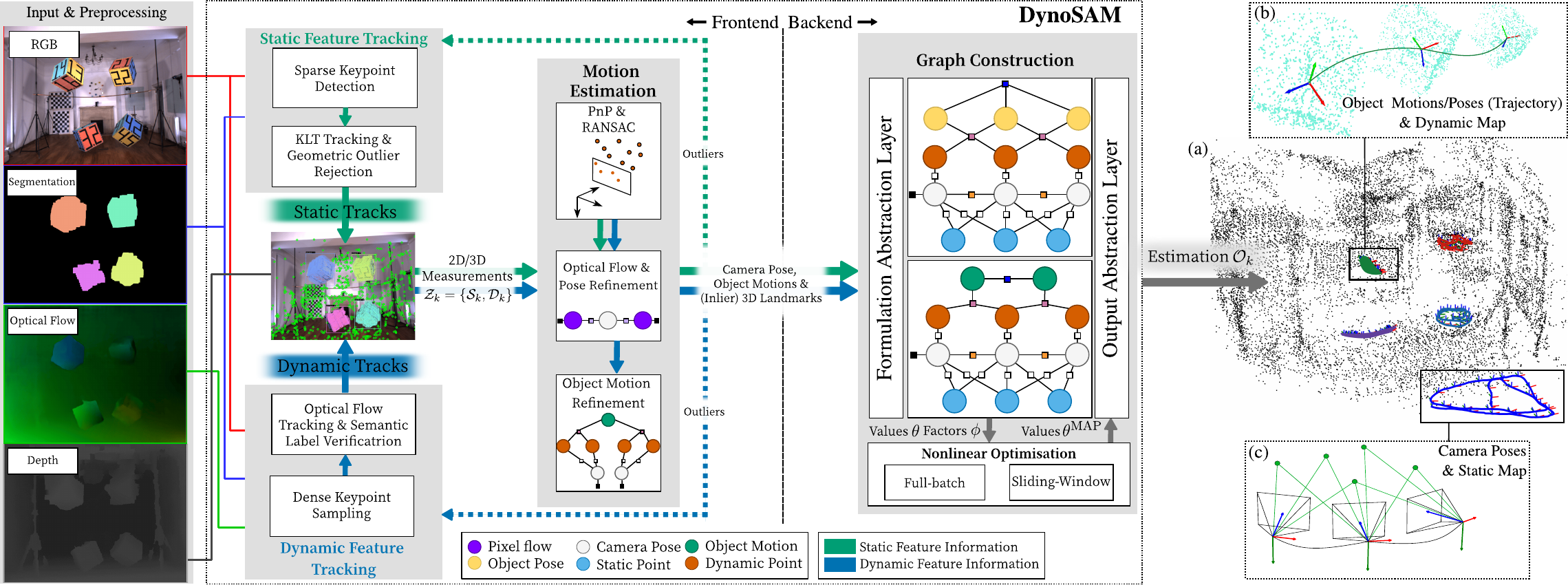}
	\caption{\footnotesize{DynoSAM system diagram. Our pipeline receives RGB, Segmentation, Optical Flow and Dense Depth images (on the left) as input and outputs static/dynamic map points, object and robot trajectories, shown in  \textbf{(a)}. A complete front-end is included for feature and object tracking as well as a back-end where different Dynamic SLAM formulations are implemented. The per-object map and trajectory is highlighted in  \textbf{(b)} and the camera trajectory in \textbf{(c)}. }}
    \label{fig:dynosam_system_diagram}
    \vspace{-6mm}
\end{figure*}
Our work is primarily focused on the estimation of object motions and poses and it is highly relevant to provide an open-source Dynamic SLAM system to facilitate evaluation and wide-spread use. 
Based on the formulations discussed in~\secref{sec:recipe_dynamic_slam}, this section presents the DynoSAM pipeline, a factor-graph-based Dynamic Visual SLAM system.
% As shown in the system diagram (\figref{fig:dynosam_system_diagram}), our pipeline takes stereo/RGBD, instance segmentation masks and dense optical flow as input,
% and produces globally consistent trajectories of the camera and objects as well as the static structure and temporal map of each object. 

Our pipeline is broken into the typical frontend-backend structure.
As shown in~\figref{fig:dynosam_system_diagram}, our pipeline takes stereo/RGBD, instance segmentation masks and dense optical flow as input,
and produces globally consistent trajectories of the camera and objects as well as the static structure and temporal map of each object (\figref{fig:dynosam_system_diagram} (a-c)). 
The front-end (\secref{sec:slam_frontend}) processes this input data to generate static and dynamic feature tracks and provides initial estimates for the back-end.
It is responsible for object-level data association across frames, 
ensuring that features are consistently tracked on the same object to enable robust estimation.
The robust handling of noisy depth and optical flow data, as well as segmentation inaccuracies, have been well studied in~\etal\cite{Henein20icra,zhang20iros}, which serve as inspiration for our front-end implementation.
Further evaluations, while relevant for real-world deployment, is beyond the scope of this particular work, since our primary focus is the understanding and evaluation of the Dynamic SLAM back-end.
The back-end (\secref{sec:backend}) implements our presented formulations in~\secref{sec:recipe_dynamic_slam} and defines the Dynamic SLAM estimation problem. 
It fuses static and dynamic measurements via a GTSAM-based~\cite{gtsam} factor-graph optimization to produce a globally consistent estimate of the dynamic scene. 
Our implementation is flexible to accommodate any custom formulation for Dynamic SLAM and options for full-batch smoothing where the system is optimized over all measurements, 
as well as a sliding-window estimation which bounds the size of the optimization problem are available. 
We emphasize that while our primary contributions encompass the back-end of a Dynamic SLAM pipeline, 
we highlight that our open-source front-end also includes implementation improvements, 
such as multi-threaded object tracking for improved efficiency, and a range of user options.

We implement both the world-centric motion and pose formulations explained in~\secref{sec:recipe_dynamic_slam} as part of our system,
and exhaustively test and analyze them on a wide variety of datasets, 
the result of which is presented in~\secref{sec:expe}.

\subsection{Image Preprocessing}
\label{sec:image_preprocessing}

The front-end requires a set of four images $\mathcal{I}_k$ as input per time-step. 
Each $\mathcal{I}_k$ consists of a RGB image \rgbi, an instance segmentation \semantici, per-pixel optical-flow \flowi, and per-pixel depth \depthi, 
as shown in \figref{fig:dynosam_system_diagram}. 
We undistort and align \rgbi with the depth image \depthi that is obtained from an RGBD or stereo camera system. 
Both depth and optical-flow are expected to be dense and can be obtained from classical~\cite{farneback2003denseflow} or learned methods~\cite{teed2020raft, huang2022flowformer}. 
Finally, \semantici is used in the process of masking out dynamic objects from the static background. 
Our system requires all background pixels to be labeled as $0$ and object pixels in the scene are labeled ${1\dots n_o}$. 
% Our current implementation uses RAFT~\cite{teed2020raft} for dense optical-flow and YOLOv8~\cite{JocherUltralyticsYOLO2023} for semantic instance segmentation.
This processing is performed online as an integral part of our complete SLAM pipeline; currently RAFT~\cite{teed2020raft} and YOLOv8~\cite{JocherUltralyticsYOLO2023} are used to compute dense optical-flow and semantic instance segmentation respectively.  

\subsection{Front-end}
\label{sec:slam_frontend}

% \Yiduo{Phrase it as an online system with preprocessing running frame to frame.}
% The front-end of DynoSAM produces the following states at every time-step:
% \begin{equation}
% \mathbf{\theta}_k = [ \campose{\worldf}{k},  \othmotion{\worldf}{k-1}{\mathcal{H}}{k}, \mathcal{Z}_{k} ]\text{,}
% \label{equ:frontend_estimated_states}
% \end{equation}
% which are then provided to the back-end.
% Our front-end is shown in \figref{fig:dynosam_system_diagram} and contains a Feature Detection \& Tracking (\secref{sec:frontend_feature_tracking}) module and a multi-stage motion estimation module responsible for camera pose (\secref{sec:frontend_camera_pose_estimation}) and object motion estimation (\secref{sec:frontend_object_motion_estimation}). Within this module the Joint Optical-Flow (\secref{sec:joint_optical_flow_refinement}) and Object Motion Refinement (\secref{sec:object_motion_refinement}) components are used to refine the camera pose and object motion estimates.

%  The front-end of DynoSAM produces the following states at every time-step:
% \begin{equation}
% \mathbf{\theta}_k = [ \campose{\worldf}{k},  \othmotion{\worldf}{k-1}{\mathcal{H}}{k}, \mathcal{Z}_{k} ]\text{,}
% \label{equ:frontend_estimated_states}
% \end{equation}
% which are then provided to the back-end.
% DynoSAM's Feature Detection \& Tracking (\secref{sec:frontend_feature_tracking}) module tracks landmarks across frames.

The DynoSAM front-end performs per-frame visual tracking, producing an initial estimation of camera pose $\campose{\worldf}{k}$ and per-object motions $\othmotion{\worldf}{k-1}{\mathcal{H}}{k}$ as well as a set of visual measurements as defined in~\secref{sec:visual_measurements}: 
%supplying the following measurements and initial values to the back-end at each time step:
%\begin{equation}
%\mathbf{\theta}_k = [ \campose{\worldf}{k},  \othmotion{\worldf}{k-1}{\mathcal{H}}{k}, \mathcal{Z}_{k} ] \: k\in \mathcal{K} \text{,}
%\label{equ:frontend_estimated_states}
%\end{equation}
%where $\mathcal{Z}_{k}$ is the set of all visual measurements:
\begin{equation*}
    \mathcal{Z}_{k} = \{ \mathcal{S}_{k}, \mathcal{D}_{k} \}\text{.}
\end{equation*}
The reminder of this section introduces each front-end module. 
The Feature Detection \& Tracking module (\secref{sec:frontend_feature_tracking}) generates $\mathcal{Z}_{k}$ by tracking static and dynamic features from $\mathcal{I}_k$. 
These measurements support camera pose estimation (\secref{sec:frontend_camera_pose_estimation}) and multi-object motion estimation (\secref{sec:frontend_object_motion_estimation}).
Joint Optical-Flow (\secref{sec:joint_optical_flow_refinement}) and Object Motion Refinement (\secref{sec:object_motion_refinement}) components are used to further refine camera pose and object motion estimates.

\subsubsection{Feature Detection \& Tracking}
\label{sec:frontend_feature_tracking}

The feature detection and tracking module matches corresponding static and per-object features between consecutive frames.
\rgbi and \flowi are used for feature detection and tracking while \semantici is used to determine whether a keypoint belongs to the static background or a dynamic object. 
This module ensures consistent tracking of features on the same object while maintaining sufficient spread and density of features across frames to preserve the accuracy of both visual odometry and object motion estimation.

After detection and tracking, 
all features are initially marked as inliers and may be marked as outliers during the motion estimation module as visualized in~\figref{fig:dynosam_system_diagram}. 
Once both static and dynamic keypoints are tracked,
we use the input depth map \depthi to directly obtain 3D measurements $\mathcal{D}_{\text{3D}, k}$ and $\mathcal{S}_{\text{3D}, k}$ from 2D pixel measurements, forming the local map shown in~\figref{fig:dynosam_system_diagram}. 
We provide a detailed explanation below.

\textbf{Static Feature Tracking}. 
At each time-step, 
a \textit{sparse} set of detected static keypoints $\mathcal{S}_{\text{2D}, k}$ are tracked across consecutive frames using optical flow. 
Our framework offers options to select different feature extraction algorithms depending on the use case; for our experiments
we used Shi-Tomasi corners~\cite{shi1994good} for indoor environments and ORB~\cite{rublee2011ORB} features for outdoor environments. 
By default, our front-end uses the Lucas-Kanade tracker~\cite{lucas1981iterative_LK} to generate feature correspondences. 
The input optical flow \flowi can also be used when it is reliable. 
Once tracked, our front-end performs geometric verification to discard poor correspondences.
% We can also use the flow image, if it is relaiable. 

To improve robust tracking, we retain only relevant features by applying
Adaptive Non-Maximal Suppression (ANMS)~\cite{bailo2018efficient_anms} to the detected keypoints.
This procedure retains only the most informative and reliable features, while simultaneously promoting a spatially uniform distribution of features across the image. 
ANMS achieves this by culling uninformative features whilst ensuring that a minimum number of features remain ($800$ in our experiments).
Our front-end will detect new keypoints on the static structure if the number of inlier tracks falls below this minimum threshold.
This adaptive mechanism ensures that a consistently high number of well-distributed features are tracked across successive frames, thereby enhancing tracking robustness and accuracy.

\textbf{Dynamic Feature Tracking}. 
We track a \textit{dense} set of features $\mathcal{D}_{\text{2D}, k}$ for each object.
Dense tracking is important to achieve good coverage over the entire observable object for robust motion estimation~\cite{zhang20iros}. 
For each dynamic measurement we maintain an associated object-level label $j$, 
such that features can be associated to the same object between frames. 

The instance segmentation mask \semantici is used to retrieve the object label $j$ per pixel. 
While this mask can be generated from any standard instance segmentation network,
many of these networks do not guarantee that the instance labels will be temporally consistent. 
Before tracking dynamic keypoints, object masks are tracked using~\cite{zhang2022bytetrack}. 
We then remap the original instance labels to a temporally consistent label $j$ per object. 
%If the user has their own method of object tracking, they may specify to not use this algorithm,  at which point the original pixel values in in the semantic instance mask will considered as the temporally consistent object label.
If the user has their own instance segmentation and tracking method, they can choose to bypass this algorithm. 
In that case, the user only needs to ensure that each pixel represents the unique tracking label $j$.

Although the instance segmentation mask \semantici is able to separate objects from the background, it cannot distinguish between static and moving objects. 
To address this, we use the method of Zhang~\etal\cite{zhang20iros} to identify moving objects based on scene flow, 
allowing us to focus on tracking features only on the dynamic objects. %This method was shown to produce very good results~\cite{zhang20iros}. 
This module can be easily modified to integrate learning-based methods to identify moving objects.

Once the moving objects in the image are identified and tracked, 
dense features are extracted by sampling keypoints uniformly within each tracked object mask. 
Optical flow \flowi is then used directly to find the correspondences between frames. 
Our tracking algorithm ensures that, where possible, a consistent and significant number of features ($800$ in our experiments) on each object are tracked. 
Similarly to static points,
only inliers from the previous frame are used.

\subsubsection{Initial Camera Pose Estimation}
\label{sec:frontend_camera_pose_estimation}

At each time-step, 
static measurements $\mathcal{S}_{k}$ are used to estimate the initial camera pose. 
As shown in the motion estimation module in~\figref{fig:dynosam_system_diagram}, 
the first step is the PnP algorithm~\cite{ke17cvpr} which estimates $\campose{\worldf}{k}$ by minimizing the re-projection error:
\begin{equation}
    \mathbf{r}_{\text{2D}, R} = \mathbf{z}_{\text{2D}, k}^i - \pi(\campose{\worldf}{k}^{-1} \: \mpoint{\worldf}{}^i)
    \label{equ:proejction_2d_residual}
\end{equation}
between the tracked static keypoints in the current frame at time-step $k$ and the local map constructed from the previous frame. RANSAC verification and outlier rejection is performed using the implementation as provided by OpenGV~\cite{kneip2014opengv} to obtain robust initial estimation of the camera pose. This initial camera pose is further refined through joint optical flow and pose optimization method detailed in~\secref{sec:joint_optical_flow_refinement}.

\subsubsection{Initial Object Motion Estimation}
\label{sec:frontend_object_motion_estimation}
%The module estimates the initial motion of each object $\objmotion{\worldf}{k-1}{k} \in \mathcal{H}_k$ at each time-step 
%following the three-step process in the motion estimation module outlined in~\figref{fig:dynosam_system_diagram}. 
The initial object motion $\objmotion{\worldf}{k-1}{k} \in \mathcal{H}_k$ is estimated following the method of Zhang~\etal\cite{zhang2020vdoslam}:
\begin{align}
    \mathbf{r} &= \mathbf{z}_{\text{2D}, k}^i- \pi(\campose{\worldf}{k}^{-1} \: \objmotion{\worldf}{k-1}{k} \: \mpoint{\worldf}{k-1}^i) \\
      &= \mathbf{z}_{\text{2D}, k}^i - \pi(\othmotion{\worldf}{k-1}{G}{k} \: \mpoint{\worldf}{k-1}^i)\text{,}
    \label{equ:object_motion_pnp_G}
\end{align}
where $\othmotion{\worldf}{k-1}{G}{k} = \campose{\worldf}{k}^{-1} \: \objmotion{\worldf}{k-1}{k}$. 
Since~\eqref{equ:object_motion_pnp_G} is in the same form as~\eqref{equ:proejction_2d_residual}, 
we can solve for $\othmotion{\worldf}{k-1}{G}{k}$ in the same fashion, 
and the object motion:
\begin{equation*}
    \objmotion{\worldf}{k-1}{k} = \campose{\worldf}{k} \: \othmotion{\worldf}{k-1}{G}{k}
\end{equation*}
can be directly recovered.  
%, as the camera pose is solved as discussed in~\secref{sec:frontend_camera_pose_estimation}. 
As before, 
RANSAC is used to detect outliers and update the inlier tracks in the current frame. 
% \Jesse{Make stronger comment that successful tracking is really important}. 
Building on~\cite{zhang2020vdoslam}, we enhance the initial object motion estimation and refine the inlier tracks through two additional steps, detailed in~\secref{sec:joint_optical_flow_refinement} and~\secref{sec:object_motion_refinement}, respectively.
The method proposed in ~\secref{sec:joint_optical_flow_refinement} was used in~\cite{zhang2020vdoslam} to improve the estimation and tracking. With the addition of ~\secref{sec:object_motion_refinement} our method augments this process to further improve the estimation and identify outliers.
At each stage of the refinement process, the set of inlier tracks is updated, significantly improving the robustness of the estimation.
%Building upon~\cite{zhang2020vdoslam}, we improve the initial object motion estimation and refine the inlier tracks by applying two additional steps, explained in~\secref{sec:joint_optical_flow_refinement} and ~\secref{sec:object_motion_refinement}, respectively.
%After each step in the refinement process, the set of inlier tracks are updated which helps to improve the robustness of the estimation.
% The set of inlier tracklets per object are continually refined, 
% as is the motion estimation, 
% at each step in the process. 
The three-steps outlined assumes that the motion of each object is independent, allowing DynoSAM to parallelize the object motion estimation process by handling each object instance in a separate thread.

\subsubsection{Joint Optical-Flow Refinement}
\label{sec:joint_optical_flow_refinement}

Following the work of~\cite{zhang2020vdoslam, zhang20iros}, 
the optical-flow is jointly refined with the camera pose and object motions. 
This step ensures robust and accurate tracking of both static and dynamic features and helps to account for any errors in the initial optical-flow calculation.
The measured flow and the associated $\SE$ transformation are jointly refined by reformulating the re-projection errors from~\eqref{equ:object_motion_pnp_G} and~\eqref{equ:proejction_2d_residual} in terms of the measured flow:
\begin{equation}
\begin{aligned}
    \mathbf{r}_{f, H} &= \mathbf{z}^i_{\text{2D}, k-1} + \mathbf{f}^i_{k-1, k} - \pi(\othmotion{\worldf}{k-1}{G}{k} \: \mpoint{\worldf}{k-1}^i) \\
    \mathbf{r}_{f, X} &= \mathbf{z}^i_{\text{2D}, k-1} + \mathbf{f}^i_{k-1, k} - \pi(\campose{\worldf}{k}^{-1} \: \mpoint{\worldf}{}^i)
\label{equ:frontend_pose3_flow_projection_factor}
\end{aligned}    
\end{equation}
\\
where $f_{k-1, k}$ defines the optical flow between two keypoints:
\begin{equation*}
f_{k-1, k} = \mathbf{z}_{\text{2D}, k} - \mathbf{z}_{\text{2D}, k-1}\text{.}
\end{equation*}
Given 3D-2D point correspondences,
the resulting non-linear least squares problem is formulated using a factor graph:
\begin{equation}
     \label{equ:frontend_flow_refinement}
    \{\theta, \mathbf{f}_{k-1, k}\}  = \underset{\{\theta, \mathbf{f}_{k-1, k}\}}{\mathrm{argmin}}
    \left(\sum_{i}^{N} \rho_h \factor{\mathbf{r}_{f, \theta}}{{f}} + \sum_{i}^{N} \factor{\mathbf{r}_0(f)}{{0}}\right)\text{,}
\end{equation}
where $\theta$ represents either $\campose{\worldf}{k}$ camera pose 
or $\othmotion{\worldf}{k-1}{G}{k}$ in the case of object motion estimation. Depending on $\theta$, $\mathbf{r}_{f, \theta}$ is either $\mathbf{r}_{f, X}$ or $\mathbf{r}_{f, H}$, as in~\eqref{equ:frontend_pose3_flow_projection_factor}.
The covariance matrix $\Sigma_0 \in \mathbb{R}^{{2 \times 2}}$ associated with the flow prior $\mathbf{r}_0(f)$ is diagonal and associated with the measured optical-flow. 
As each object moves independently, 
we construct and solve this optimization problem in parallel per object. 
% Through this joint optimization, we refine temporal point associations, improving feature tracking and pose estimation. 
During optimization, we enhance robustness by eliminating additional outliers, specifically those exhibiting large re-projection errors.

% of the module. 

\subsubsection{Object Motion Refinement}
\label{sec:object_motion_refinement}
\begin{figure}[t]
	\centering
	\includegraphics[trim={0cm 0cm 0cm 0cm},clip,width=0.89\columnwidth]{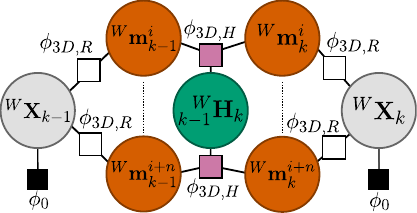}
	\caption{\footnotesize{Factor graph representing the motion only refinement graph. For each object $j \in \mathcal{J}_k$, the motion is refined by minimizing the 3D motion residual using measurements of points tracked between $k-1$ and $k$. In our example we show the dynamic points in orange, the motion in green, the 3D motion residual in purple and the re-projection factors in white. We only show two sets of points $i$ and $i+n$, while in reality there would be $n$ motion factors. }}
    \label{fig:motion_frontend_refinement_graph}
    \vspace{-6mm}
\end{figure}

The object motion estimates can be further refined by employing the rigid-body motion model defined in~\eqref{equ:world_landmark_motion_tenary_factor} to formulate an additional nonlinear least-square optimization problem that directly estimates the object motion $\objmotion{\worldf}{k-1}{k}$. 
In contrast to the preceding step, which uses a 2D pixel error, this residual function is based on 3D point errors. Empirically, we have observed that this 3D-based approach significantly enhances estimation accuracy and removes additional outliers not previously identified. 
Using the motion estimate from the previous step as an initial estimate, the \textit{motion only refinement graph} shown in~\figref{fig:motion_frontend_refinement_graph} is build per object and defines the nonlinear least-squares problem:
\begin{subequations}
\begin{align}
     \label{equ:frontend_motion_refinement_pose_prior}
    \mathbf{\theta_k}^{\text{MAP}} =& \underset{\bm{\objmotion{}{}{}}}{\mathrm{argmin}} \biggl( \factor{\mathbf{r}_0 (\campose{\worldf}{k-1})}{0}  + \factor{\mathbf{r}_0 (\campose{\worldf}{k})}{0} \\
    \label{equ:frontend_motion_refinement_2d_projection}
    &+ \sum_{{\mathcal{D}_{k-1}^j} \cup {\mathcal{D}_{k}^j}} \factor{\mathbf{r}_{\text{2D}, R}}{{\text{2D},R}} \\
     \label{equ:frontend_motion_refinement_3d_H}
    &+ \sum_{{\mathcal{D}_{k-1}^j} \cup {\mathcal{D}_{k}^j}} \factor{\mathbf{r}_{\text{3D}, H}}{{\text{3D},H}} \biggr)\text{,}
\end{align}
\end{subequations}
The Levenberg–Marquardt solver is used to obtain the final object motion estimate $\objmotion{\worldf}{k-1}{k}$.
Based on~\eqref{equ:proejction_2d_residual}, \eqref{equ:frontend_motion_refinement_2d_projection} forms projection factors with an associated covariance matrix $\Sigma_{2D, R} \in \mathbb{R}^{2 \times 2}$, connecting dynamic points and the observing camera poses at time-step $k-1$ and $k$. 
\eqref{equ:frontend_motion_refinement_3d_H} describes the 3D motion residual of~\eqref{equ:world_landmark_motion_tenary_factor} and connects the common motion with each tracked landmark. $\Sigma_{3D, H} \in \mathbb{R}^{3 \times 3}$ is the associated covariance matrix.
Equation~\eqref{equ:frontend_motion_refinement_pose_prior} represents prior factors on the camera poses. 
Since this refinement step focuses exclusively on improving object motion estimation, and not camera pose,
a strong prior on the camera poses are used during optimization. This is achieved by assigning the covariance matrix $\Sigma_0 \in \mathbb{R}^{6 \times 6}$ with very small values ($\sigma=0.0001$ in our experiments). 

% The back-end component is responsible for constructing new factors and variables based on inputs received from the front-end. 
% These factors and variables are subsequently optimized, 
% and the resulting optimized states are post-processed into the output $\mathcal{O}_k$. 

\begin{figure*}[t]
	\centering
    \includegraphics[width=\textwidth]{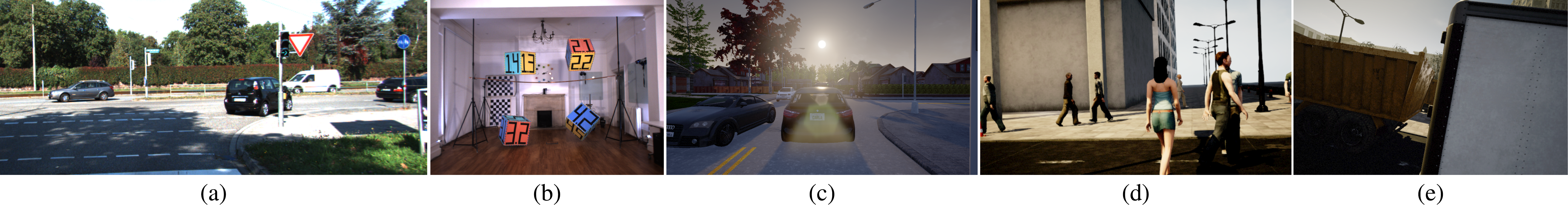}
	\caption{\footnotesize{ Example scenarios from datasets used: \textbf{(a)}: KITTI tracking, \textbf{(b)}: OMD, \textbf{(c)}: Outdoor Cluster, \textbf{(d)}: TartanAir (Shibuya), \textbf{(d)}: VIODE. }}
    \label{fig:example_datasets}
    \vspace{-6mm}
\end{figure*}

\subsection{Back-end}
\label{sec:backend}
% The DynoSAM back-end is responsible for joint estimating the camera pose and static map points, as well as object points and their poses or motions. 
The back-end fuses static and dynamic measurements to jointly estimate the trajectory of the camera and each object as well as a global map of static and dynamic points. 
% Compared to the initial estimates from the front-end, 
% this fusion  over a greater time horizon. 
Our back-end allows different Dynamic SLAM formulations to be selected and used for estimation. 
Full-batch and sliding window methods are provided for optimisation.
%The state of each dynamic object is dependent upon the formulation of the non-linear optimization problem used to solve the Dynamic SLAM system. 

As previously discussed, formulations can represent objects using different state variables (poses or motions as shown in this work) and in different frames of reference (locally as in DynaSLAM II~\cite{bescos2021ral} or in the camera frame as in MVO~\cite{judd2024ijrr_mvo}).
Therefore, to support modularity, evaluation and use in downstream applications DynoSAM defines a common interface between the formulation the system output.
A pose-estimation processing step is implemented that converts all estimated states into a common output:
\begin{equation}
\mathcal{O}_k = [ \campose{\worldf}{k},   \othmotion{\worldf}{k-1}{\mathcal{H}}{k}, {^{\worldf}\mathcal{L}_{k}} , {^\worldf\mathcal{M}_k}]\: k\in \mathcal{K}
\end{equation}
If the formulation used does not directly estimate a required output element, it must be computed directly from the available information. 
This behavior is enforced in DynoSAM's back-end implementation via clear layers of abstraction.

% DynoSAM's back-end implementation offers abstraction layers that clearly enforce this behavior, enabling streamlined evaluation, visualization, and integration with downstream tasks.
% \Jesse{Is this worth a UML or class diagram?}

% The back-end provides functionality for full batch or sliding window estimation, and is independent to the formulation selected. 

\subsection{Implementation}
The camera pose and motion estimation algorithms described in the previous sections are implemented in C++. 
GTSAM 4.2~\cite{gtsam} is used for all non-linear optimization problems which are solved using Levenberg-Marquardt. 
For all experiments, unless otherwise specified (\secref{sec:sliding_window_opt}), full-batch optimization is used. 

% The variables and factors used to construct the detailed factor graphs are detailed in \tabref{tab:fg_implementation}. The \verb|PriorFactor| and \verb|PoseToPointFactor| are built into the GTSAM library while the other factors are custom for our framework. The \verb|LandmarkMotionTernaryFactor| was first introduced in~\cite{zhang2020vdoslam}, while the \verb|LandmarkMotionPoseFactor| and \verb|LandmarkPoseSmoothingFactor| are novel for this work. 
% The Levenberg-Marquardt optimizer is used to solve the resulting nonlinear least-squares problems. 
% For all experiments except~\secref{sec:sliding_window_opt}, full-batch optimization is used. 

At each time-step, the back-end component constructs a system by incorporating new factors and variables based on inputs from the front-end, 
which consists of initial estimates for the camera pose $\campose{\worldf}{k}$, per-object motions $\othmotion{\worldf}{k-1}{\mathcal{H}}{k}$ and a set of visual measurements $\mathcal{Z}_k$. 
After optimisation, the estimates are processed to generate the output $\mathcal{O}_k$. 

For both formulations discussed in~\secref{sec:recipe_dynamic_slam}, 
the system defines new variables at each time-step for $\objmotion{\worldf}{k-1}{k}$ or $\objpose{\worldf}{k}$ which represent the state of each object, 
and initializes from the front-end where applicable. 
% Each dynamic point measurement in $\mathcal{D}_{3D, k}$ contributes to constraining object states based on the observed camera pose(s) and its corresponding dynamic map point.

% Compared to classical static SLAM, 
% this combination of factors and state variables results in a significant increase in graph size and connectivity complexity as well as associated bookkeeping, 
% making Dynamic SLAM a far more complex problem. 
% DynoSAM efficiently manages and constructs the Dynamic SLAM problem by decoupling the measurement bookkeeping from the implementation of individual Dynamic SLAM formulations through abstraction layers, as mentioned in the previous section.

% Compared to a typical static SLAM system, the graph format of the input data is significantly more complex, and may contain ternary~\eqref{equ:world_landmark_motion_tenary_factor} and quaternary edges~\eqref{equ:world_landmark_motion_pose_factor}.

The combination of factors and state variables results in a Dynamic SLAM problem that, compared to a typical static SLAM system, is significantly more complex, and may contain ternary~\eqref{equ:world_landmark_motion_tenary_factor} and quaternary edges~\eqref{equ:world_landmark_motion_pose_factor}.
Despite the significant increase in graph size and associated bookkeeping DynoSAM efficiently manages the construction of each graph. 
DynoSAM achieves this by decoupling the measurement bookkeeping from the implementation of individual Dynamic SLAM formulations through abstraction layers, as mentioned in the previous section.
This design choice dramatically simplifies the implementation of new formulations, allowing researchers to easily experiment with different approaches.
Further technical details regarding graph construction and the abstractions used are available in our documented open-source software. 
% \footnote{\url{https://github.com/ACFR-RPG/DynOSAM}}.

Finally, we note that loop closure has not been incorporated into the system and is beyond the scope of this work.

\section{Experiments}
\label{sec:expe}

% %%%%%%%%%%%%%%%%%%%%%%%%%%%%%%%%%%%%%%%%%%%%%%%%%%%%%%%%%%%%%%%%%%%%%%%%%%%%%%%%%%%%%%%%%%%%%%%%%%%%%%%%%%%%%%%

The primary focus of this work is the estimation object motion and pose. 
Hence, we evaluate both DynoSAM's WCME (\secref{sec:recipe_motion}) and WCPE (\secref{sec:recipe_pose}) formulations on a range of indoor and outdoor datasets featuring diverse dynamic objects and contain ground truth for both camera pose and object motion/pose.
% Our framework has been rigorously tested on indoor and outdoor datasets featuring diverse dynamic objects.
% We present results of camera trajectories and object motions using full-batch optimization, 
% as well as the performance using sliding window optimization. 
To facilitate a wide range of comparison we evaluate DynoSAM on the KITTI tracking~\cite{Geiger13ijrr}, OMD~\cite{Judd19ral} and Outdoor Cluster~\cite{Huang2019iccv}\footnote{The datasets are provided in addition to the paper and can be found at \url{https://huangjh-pub.github.io/page/clusterslam-dataset/}. Note that the ground truth rotation is actually in $[qw, qx, qy, qz]$ form instead of the $[qx, qy, qz, qw]$ format stated on the website}.
We use VDO-SLAM~\cite{zhang2020vdoslam}, MVO~\cite{judd2024ijrr_mvo} and ClusterSLAM~\cite{Huang2019iccv} as state-of-the-art Dynamic SLAM systems to assess our object motion/pose estimation. 
Furthermore, we propose a new evaluation metric for object motion and discuss how it better facilitates comparison across systems.
% We compare the accuracy of our object motion/pose estimation against other state-of-the-art Dynamic SLAM systems VDO-SLAM~\cite{zhang2020vdoslam}, MVO~\cite{judd2024ijrr_mvo} and ClusterSLAM~\cite{Huang2019iccv} as these systems provide comparable results. 

In addition to object motion/pose, we evaluate our camera pose estimation by comparing against DynaSLAM II~\cite{bescos2021ral}, VDO-SLAM~\cite{zhang2020vdoslam}, MVO~\cite{judd2024ijrr_mvo}, ClusterSLAM~\cite{Huang2019iccv}, AirDOS~\cite{Qiu2022icra_airdos}, ORB-SLAM3~\cite{Campos2021tro} and DynaVINS~\cite{Song2022ral_dynavins}.
Datasets that do not contain ground truth object information, such as TartanAir Shibuya~\cite{Qiu2022icra_airdos} and VIODE~\cite{minoda2021viode}, are also included to validate the accuracy and robustness of our camera pose estimation in highly dynamic and challenging environments.
% The main scope of this paper is to evaluation of motion estimation
% \Jesse{the main scope of the paper is.... is to propose main estimation}
% We 
% \Jesse{this and metrics and contributions}
% Since not all systems evaluate or estimate for the objects in the scene, we compare our object motion/pose estimation with VDO-SLAM~\cite{zhang2020vdoslam}, MVO~\cite{judd2024ijrr_mvo} and ClusterSLAM~\cite{Huang2019iccv} as these systems provide comparable results. 
% In particular, we compare the accuracy of DynoSAM's WCME (\secref{sec:recipe_motion}) and WCPE (\secref{sec:recipe_pose}) against MVO~\cite{judd2024ijrr_mvo}, VDO-SLAM~\cite{zhang2020vdoslam}, ClusterSLAM~\cite{Huang2019iccv} as these systems provide comparable object pose and/or motion evaluations. 
% DynaSLAM II~\cite{bescos2021ral} is additionally included when comparing camera pose estimation.
% \Jesse{wer have new evaluation matrics and...}
Our evaluation is conducted using the metrics outlined in~\secref{sec:metrics} and provides a comprehensive assessment of these systems. 
All evaluation metrics are included in our open-source framework and are based on the implementation of \textit{evo}~\cite{grupp2017evo}.

% Of the systems we compare against, only VDO-SLAM~\cite{zhang2020vdoslam} provides an open-source implementation of their pipeline, which we have used to directly generate the results reported in this paper.
% In the case of MVO, their pipeline's raw output data was provided by the authors, 
% allowing all metrics to be used during evaluation. 
% ClusterSLAM~\cite{Huang2019iccv} uniquely reports accumulated average camera and object pose errors over all sequences, 
% instead of individual results per sequence. 
% We therefore report an accumulated average error to facilitate comparison with their system and use the values reported in~\cite{Huang2019iccv}. 
% We use the camera pose errors of DynaSLAM II as reported in~\cite{bescos2021ral}. 
% DynaSLAM II only reports individual object errors and provide evaluations for on average only one or two objects per sequence. 
% In addition, 
% DynaSLAM II does not provide an open-source implementation, 
% and there lacks a clear association of object IDs between their results and ours; 
% hence we have to unfortunately exclude DynaSLAM II object errors from our comparisons. 
% However, we note that DynaSLAM II's reported object poses have significantly higher errors, especially in rotation, than our system's estimations.

\subsection{Datasets and Third Party Results}
\label{sec:datasets}

\subsubsection{Sequences with Camera \& Object Ground Truth}
% We use the KITTI tracking dataset~\cite{Geiger13ijrr}, Outdoor Cluster~\cite{Huang2019iccv} and OMD~\cite{Judd19ral} to asses object motion and pose estimation in scenarions that contain diverse and complex motions from multiple objects, partial and full occlusions, as well as semi-rigid objects such as cyclists. 
% Importantly, these dataets contain ground truth object pose/motion as well as camera pose.
We use the KITTI tracking dataset~\cite{Geiger13ijrr}, Outdoor Cluster~\cite{Huang2019iccv} and OMD~\cite{Judd19ral} (\figref{fig:example_datasets}~a-c) to evaluate object motion and pose estimation in scenes with multiple objects, complex motion, occlusion, and semi-rigid entities such as cyclists. 
% All datasets provide ground truth for both object and camera trajectories.
We use a processed version of the KITTI tracking dataset containing ground truth information per object per frame. Outdoor Cluster is simulated and provides ground truth. 
OMD uses a Vicon system to capture all ground truth trajectories~\cite{Judd19ral}.

KITTI and Outdoor Cluster represent large-scale outdoor driving scenarios with numerous moving objects.
However, most of these objects experience relatively constant motion. 
% In contrast, 
% To assess DynoSAM's ability to estimate complex dynamics, 
We therefore additionally include \textit{swinging\_4\_unconstrained} (S4U), 
the OMD sequence with the most challenging dynamics, 
which features $4$ swinging cubes with unconstrained and unpredictable motions. 
The hand-held camera motion further increases the difficulty. 
% The authors of MVO~\cite{judd2024ijrr_mvo} kindly provided the output of MVO tested against S4U for our new motion error metric evaluation. 
% Although OMD includes other sequences, S4U presents the most challenging dynamics and facilitates comparison with MVO. 
While MVO uses another OMD sequence to test their object re-association ability after occlusion, 
this functionality is not relevant for motion evaluation and is therefore not assessed in this work.

\subsubsection{Challenging Sequences with Only Camera Ground Truth}

While not our primary focus, evaluating camera pose estimation is crucial to demonstrate DynoSAM’s accuracy and robustness in challenging conditions, including dynamic objects, occlusions, and low light. 
We test on the simulated TartanAir Shibuya~\cite{Qiu2022icra_airdos} and VIODE~\cite{minoda2021viode} datasets. 
TartanAir features over $30$ moving humans with complex motions that frequently occlude the view (\figref{fig:example_datasets}~(d)). 
VIODE poses further challenges with full occlusion scenarios as shown in~\figref{fig:example_datasets}~(e), low-light, and includes UAV-like motion with extreme translation and rotation. This dataset additionally contains IMU.

% While additional evaluation of camera pose estimation is not our main focus, it remains highly relevant to validate the accuracy and robustness of DynoSAM in challenging sequences with highly dynamic objects, full occlusion and low-light conditions. 
% We therefore evaluate DynoSAM on the simulated TartanAir Shibuya~\cite{Qiu2022icra_airdos} and VIODE~\cite{minoda2021viode}. 
% Each TartanAir Shibuya sequence contains more than $30$ moving humans with complex and non-rigid motions that regularly cover the majority of the view. This can be seen in~\figref{fig:example_datasets}~(d).
% % Moving objects often cover a large field of view, creating scenarios where the majority of the image is dynamic. 
% VIODE is particularly challenging for pure visual SLAM due to sequences in low light as well as total occlusion situations, whereby most or all of the view is occluded by dynamic objects, as shown in~\figref{fig:example_datasets}~(e).
% Uniquely, the camera motion in VIODE emulates a UAV with extreme translation and rotation. 
% VIODE, however, additional contains IMU.
% and is tailored for Visual-Inertia Navigation Systems (VINS)~\cite{Song2022ral_dynavins}.
% \Yiduo{VIODE however also contains IMU.}
% \subsubsection{System Comparisons}

\subsubsection{Obtaining Third Party Results For Object Motions}

Obtaining consistent object motion/pose evaluation from third-party systems is challenging due to limited open-source options and different metrics and formats.
Of the Dynamic SLAM systems we compare against, only VDO-SLAM~\cite{zhang2020vdoslam} is open-source, and we used it to generate results directly. To ensure fairness we run VDO-SLAM on the same processed data as DynoSAM.
VDO-SLAM shares the underlying motion formulation with WCME but differs in its use of g2o~\cite{Kummerle11icra} for optimizations, as well as in its front-end design and tracking strategy.
MVO~\cite{judd2024ijrr_mvo} reports object motions for a few sequences, and the authors kindly provided results on S4U and KITTI $00$ for our evaluation. 
% , the results of which the authors have kindly provided. 
% On the Outdoor Cluster sequences, 
ClusterSLAM~\cite{Huang2019iccv} uniquely reports average camera and object pose errors accumulated over all Outdoor Cluster sequences.
We therefore report an accumulated average error on this dataset and compare against the values reported by Huang~\etal\cite{Huang2019iccv}. 
DynaSLAM II~\cite{bescos2021ral} provides per-object pose errors for a few selected objects per sequence. 
On the KITTI dataset, without source code, we cannot align object IDs and thus unfortunately compare only camera pose for these sequences reported in their paper.
We however note that DynoSAM improves upon DynaSLAM II’s reported object pose estimation by $91\%$ in rotation and $36\%$ in translation.

\subsection{Metrics}
\label{sec:metrics}
% \Jesse{actually highlight we report average object results per frame, however, our framework outputs per object errors - allowing more detailed comparison if necessary}

This section details the metrics used in our DynoSAM framework to evaluate the accuracy of pose and motion estimations for camera and objects. 
For any error $\mathbf{E}_k \in \SE$ at time-step $k$,
we compute and report the root-mean-squared error (RMSE) of the translation and rotation component separately throughout a sequence: 
\begin{equation}
\begin{aligned}
    \label{equ:rmse}
    \mathbf{E}_t &= \text{RMSE}(trans\left(\mathbf{E_k}\right))\\
    \mathbf{E}_r &= \text{RMSE}(rot\left(\mathbf{E_k}\right))\\
    \text{RMSE}(e) &= \sqrt{ \frac{1}{n(\mathcal{K})}\sum_{k\in\mathcal{K}} \lVert e_k \rVert^2}\text{,}
\end{aligned}
\end{equation}
where $e_k$ is the scalar error of either the $L_2$ norm of the translational component, denoted as $trans\left(\mathbf{E_k}\right)$, 
or the angle of the rotational component, denoted as $rot\left(\mathbf{E_k}\right)$, of $\mathbf{E}_k \in \SE$. 
% The scalar translational error is the $L_2$ norm of the translation component $trans(\mathbf{E}_k)$, 
% and the rotational error is the angle of the rotation component $rot(\mathbf{E}_k)$

\subsubsection{Pose Evaluation}
Camera and object pose estimation accuracy is evaluated using Absolute Trajectory Error (ATE) and Relative Pose Error (RPE) as defined by Sturm~\etal\cite{sturm2012iros}.
Given a ground truth transformation ${\mathbf{M}_\text{\textcolor{red}{gt}, k}\in \SE}$ at $k$ and a corresponding reference estimate $\mathbf{M}_\text{k}\in \SE$, the ATE is defined as:
\begin{equation}
    \mathbf{ATE} = \text{RMSE}\big(trans\left( \mathbf{M}_{\text{\textcolor{red}{gt}}, k}^{-1} \: \mathbf{M}_{k} \right)\big)\text{.}
\end{equation}
The ground truth reference frame is emphasized in red to clearly distinguish it from the estimated reference frame.

ATE is commonly used to assess the global consistency in trajectory estimation~\cite{mur2017orb, bescos2021ral, huang2020cvpr} and we use ATE to evaluate the accuracy in camera pose. 
We do not evaluate ATE for objects since it greatly depends on the ATE of the camera and how the object frame is defined for each system.

\begin{table*}[t]
\footnotesize
\centering
\setlength{\tabcolsep}{5.3pt}
\caption{\footnotesize{Quantitative evaluations of camera trajectory (ATE and RPE) against other Dynamic SLAM systems. Entries marked with `-' indicate that results are not available for that sequence. Best results are marked in bold and second best with underscore.}}
\label{tab:camera_experiments}
\begin{tabular}{cc|ccccccccc|>{}c>{}c>{}c>{}c>{}c|c}
% \begin{tabular}{cc|ccccccccc|ccccc|c}
\toprule
 & & \multicolumn{9}{c|}{KITTI} & \multicolumn{5}{c|}{Outdoor Cluster} & OMD \\
 & & $00$ & $01$ & $02$ & $03$ & $04$ & $05$ & $06$ & $18$ & $20$ & L1 & L2 & S1 & S2 & avg. & S4U \\
\midrule
\midrule
\parbox[t]{2mm}{\multirow{6}{*}{\rotatebox[origin=c]{90}{ATE(\si{\meter})}}}
% \multirow{5}{*}{ATE(\si{\meter})} 
 & DynaSLAM II &  $\underline{1.29}$ & $2.31$ & $0.91$ & $\mathbf{0.69}$ & $\mathbf{1.42}$ & $\mathbf{1.34}$ &  $\mathbf{0.19}$ &  $\mathbf{1.09}$ &  $\underline{1.36}$ & - & - & - & - & - & $0.21$ \\
 & VDO-SLAM & $3.37$ & $6.74$ & $2.47$ & $2.12$ & $4.53$ & $3.8$ & $0.45$ & $9.94$ & $7.82$ & - & - & - & - & - & $0.19$ \\
 & ClusterSLAM & - & - & - & - & - & - & - & - & - & - & - & - & - & $0.53$ & - \\
 & MVO & $1.53$ & - & - & - & - & - & - & - & - & - & - & - & - & - & $\mathbf{0.05}$ \\
 & WCPE (ours) & $\mathbf{0.82}$ & $\mathbf{2.00}$ & $\mathbf{0.73}$ & $\underline{0.82}$ & $\underline{2.01}$  & $\underline{1.58}$ & $\underline{0.31}$ & $\underline{1.84}$ & $\mathbf{1.26}$  & $\mathbf{0.61}$ & $\mathbf{0.52}$ & $\mathbf{0.09}$ & $\mathbf{0.13}$ & $\mathbf{0.34}$ & $\underline{0.11}$ \\
 & WCME (ours) & $\mathbf{0.82}$ & $\mathbf{2.00}$ & $\mathbf{0.73}$ & $\underline{0.82}$ & $\underline{2.01}$  & $\underline{1.58}$ & $\underline{0.31}$ & $\underline{1.84}$ & $\mathbf{1.26}$  & $\mathbf{0.61}$ & $\mathbf{0.52}$ & $\mathbf{0.09}$ & $\mathbf{0.13}$ & $\mathbf{0.34}$ & $\underline{0.11}$ \\
\midrule
\midrule
\parbox[t]{2mm}{\multirow{6}{*}{\rotatebox[origin=c]{90}{$\text{RPE}_r$(\si{\degree})}}}
% \multirow{5}{*}{$\text{RPE}_r$(\si{\degree})}
 & DynaSLAM II & $0.06$ & $\underline{0.04}$ & $\mathbf{0.02}$ & $0.06$ & $\mathbf{0.06}$ & $\mathbf{0.03}$  & $\mathbf{0.04}$  & $\mathbf{0.02}$ & $\mathbf{0.04}$ & - & - & - & - & - & - \\
 & VDO-SLAM & $0.08$ & $0.05$ & $\underline{0.03}$ & $\mathbf{0.03}$ & $\mathbf{0.06}$ & $\mathbf{0.03}$ & $0.1$ & $\underline{0.03}$ & $\mathbf{0.04}$ & - & - & - & - & - & $0.77$ \\
 & ClusterSLAM & - & - & - & - & - & - & - & - & - & - & - & - & - & $1.15$ & - \\
 & MVO & $0.19$ & - & - & - & - & - & - & - & - & - & - & - & - & - & $\underline{0.76}$ \\
 & WCPE (ours)& $\underline{0.05}$ & $\mathbf{0.03}$ & $\mathbf{0.02}$ & $\underline{0.05}$ & $\mathbf{0.06}$ & $0.06$ & $\underline{0.05}$ & $0.04$ & $\mathbf{0.04}$ & $\mathbf{0.02}$ & $\mathbf{0.02}$ & $\mathbf{0.01}$ & $\mathbf{0.02}$ & $\mathbf{0.02}$ & $\mathbf{0.69}$ \\
 & WCME (ours) & $\mathbf{0.04}$ & $\mathbf{0.03}$ & $\mathbf{0.02}$ & $\underline{0.05}$ & $\mathbf{0.06}$ & $\underline{0.05}$ & $\underline{0.05}$ & $0.04$ & $\mathbf{0.04}$ & $\mathbf{0.02}$ & $\mathbf{0.02}$ & $\mathbf{0.01}$ & $\mathbf{0.02}$ & $\mathbf{0.02}$ & $\mathbf{0.69}$ \\
\midrule
\parbox[t]{2mm}{\multirow{6}{*}{\rotatebox[origin=c]{90}{$\text{RPE}_t$(\si{\meter})}}}
% \multirow{5}{*}{$\text{RPE}_t$(\si{\meter})}
 & DynaSLAM II & $\mathbf{0.04}$ & $\underline{0.05}$ & $\underline{0.04}$ & $\mathbf{0.04}$ & $\underline{0.07}$ & $\underline{0.06}$ & $\underline{0.02}$ & $\underline{0.05}$ & $0.07$  & - & - & - & - & - & - \\
 & VDO-SLAM & $0.09$ & $0.15$ & $0.05$ & $0.09$ & $0.14$ & $0.11$ & $0.04$ & $0.09$ & $0.30$ & - & - & - & - & - & $0.12$ \\
 & ClusterSLAM & - & - & - & - & - & - & - & - & - & - & - & - & - & $1.10$ & - \\
 & MVO & $\underline{0.07}$ & - & - & - & - & - & - & - & - & - & - & - & - & - & $ \mathbf{0.004}$ \\
 & WCPE (ours) & $\mathbf{0.04}$ & $\mathbf{0.04}$ & $\mathbf{0.03}$ & $\underline{0.05}$ & $\underline{0.07}$ & $\mathbf{0.05}$ & $\mathbf{0.01}$ & $\mathbf{0.04}$ & $\underline{0.04}$ & $\mathbf{0.04}$ & $\mathbf{0.02}$ & $\mathbf{0.01}$ & $\mathbf{0.02}$ & $\mathbf{0.02}$ & $\underline{0.006}$ \\
 & WCME (ours) & $\mathbf{0.04}$ & $\mathbf{0.04}$ & $\mathbf{0.03}$ & $\underline{0.05}$ & $\mathbf{0.06}$ & $\mathbf{0.05}$ & $\mathbf{0.01}$ & $\mathbf{0.04}$ & $\mathbf{0.02}$ & $\mathbf{0.04}$ & $\mathbf{0.01}$ & $\mathbf{0.01}$ & $\mathbf{0.02}$ & $\mathbf{0.02}$ & $\underline{0.006}$ \\
\bottomrule
\end{tabular}
\vspace{-4mm}
\end{table*}

% \begin{table*}[t]
% \footnotesize
% \centering
% \setlength{\tabcolsep}{4pt}
% \caption{ATE evaluation of camera pose on TartanAir (Shibuya)~\cite{Qiu2022icra_airdos} compared against AirDOS~\cite{Qiu2022icra_airdos} (with mask) and VDO-SLAM~\cite{zhang2020vdoslam}}.
% \label{tab:camera_experiments_tas}
% \begin{tabular}{cc|ccccccc}
% \toprule
%  & & \multicolumn{7}{c}{TartanAir Shibuya} \\
%  & & \multicolumn{2}{c}{Standing Human} & \multicolumn{3}{c}{Road Crossing (Easy)} &  \multicolumn{2}{c}{Road Crossing (Hard)}\\
%  & & I & II & III & IV & V & VI & VII  \\
% \midrule
% \midrule
% \parbox[t]{2mm}{\multirow{6}{*}{\rotatebox[origin=c]{90}{\qquad ATE(\si{\meter})}}}
% % \multirow{5}{*}{ATE(\si{\meter})} 
% & AirDOS & $0.0606$ & $\mathbf{0.0193}$ & $0.0951$ & $0.0331$ & $\mathbf{0.0206}$ & $0.2230$ & $0.5625$ \\
% & VDO-SLAM & $0.0994$ & $0.6129 $ & $0.3813$ & $0.3879$ & $0.2175$ & $0.2400$ & $0.6628$ \\
% & WCME (ours) & $\mathbf{0.0243}$ & $0.0386$ & $\mathbf{0.023}$ & $\mathbf{0.0227}$ & $0.0383$ & $\underline{0.0335}$ & $\underline{0.1845}$ \\
% & WCPE (ours) & $\underline{0.0266}$ & $\underline{0.0331}$ & $\underline{0.0234}$ & $\underline{0.0212}$ & $\underline{0.0304}$ & $\underline{0.0393}$ & $\mathbf{0.1762}$ \\
% \bottomrule
% \end{tabular}
% % \vspace{-4mm}
% \end{table*}

We use the RPE metric for camera and object pose evaluation.
Unlike in some classical visual SLAM systems~\cite{Geiger13ijrr, mur2017orb} where RPE evaluates the accumulated camera pose errors over distance traveled, 
this RPE metric quantifies the difference in relative poses between consecutive frames and is employed by state-of-the-art Dynamic SLAM systems~\cite{Huang2019iccv, bescos2021ral, huang2020cvpr, judd2024ijrr_mvo}:
\begin{equation}
    \mathbf{RPE}_{k} = \big( \mathbf{M}_{\text{\textcolor{red}{gt}}, k-1}^{-1} \: \mathbf{M}_{\text{\textcolor{red}{gt}}, k} \big)^{-1} \: \big( \mathbf{M}_{k-1}^{-1} \: \mathbf{M}_{k} \big) \text{.} 
\label{equ:rpe}
\end{equation}
We report translation and rotation error separately using~\eqref{equ:rmse}.
% \begin{equation}
% \begin{aligned}
%     \mathbf{RPE}_t &= \text{RMSE}\big(trans\left(\mathbf{RPE}_k\right)\big)\\
%     \mathbf{RPE}_r &= \text{RMSE}\big(rot\left(\mathbf{RPE}_k\right)\big)\text{,}
% \end{aligned}
% \end{equation}
For camera pose, we set $\mathbf{M} = \campose{\worldf}{k}$ to compute ATE and RPE. 
We assess object RPE by setting $\mathbf{M} = \objpose{\worldf}{k}$.
% the pose of an object changes based on where its local reference frame is defined, and may vary significantly for each system.

\subsubsection{Motion Evaluation}
\label{sec:motion_metrics}

The primary scope of our work focuses not only on poses but, more importantly, on object motion estimation. 
% We estimate object motion in the world frame, $\objmotion{\worldf}{k-1}{k}$, because $\objmotion{\worldf}{k-1}{k}$ is invariant to the arbitrary object frame definition. 
% However, the magnitude of the motion error, is influenced by the object position relative to the world frame, 
% as indicated by Appendix~\ref{sec:app_motion_model}. 
% Even a small discrepancy in the actual motion can have a large error when transformed into the world frame. 
% Hence we do not evaluate $\objmotion{\worldf}{k-1}{k}$ directly. 
% \Jesse{in experiments related to samne discissions (RPE vs ME) the WCPE vs WCME }\
For proper motion evaluation, we must again consider how the motion changes when represented in different frames. 
As discussed in~\secref{sec:motion_explanation}, the local motion of an object depends directly on where the body frame is defined on the object. 
Since the object frame may be arbitrarily defined, there is no guarantee that the frame $\{\objf\}$ for any object $j$ will align between different systems and with the ground truth. 
Setting $\mathbf{M} = \objpose{\worldf}{k}$ in~\eqref{equ:rpe} and evaluating the object RPE, as many systems do, highlights the potential discrepancy:
\begin{equation*}
    \mathbf{RPE}_{k}(\objpose{}{}) = \objmotion{\objf_{\text{\textcolor{red}{gt}}, k-1}}{k-1}{\text{\textcolor{red}{gt}}, k}^{-1} \: \objmotion{\objf_{k-1}}{k-1}{k}\text{,}
\end{equation*}
as this error compares the ground truth motion in the \textit{ground truth} object frame with the estimated motion that is represented in the \textit{estimated} object frame.
Therefore, depending on the definition of $\{L\}$ relative to $\{L_{\textcolor{red}{gt}}\}$ the reported error may be highly inconsistent. 

An obvious solution is to instead directly evaluate the observed motion $\objmotion{\worldf}{}{}$. This representation is agnostic to the placement of the object frame on the rigid-body and is shared between all systems, including ground truth. 
However, expressing the motion error in the world frame has the unintended effect of `scaling' the error's magnitude with the objects position relative to $\{\worldf\}$, as shown in Appendix~\ref{sec:app_motion_model}. 
Consequently, even a small discrepancy in the actual motion can have a large error when the object is far from the origin. 
% \Jesse{appendix?}

To resolve both of these issues, we propose a new metric, Motion Error $\mathbf{ME}$, that evaluates all motions
in the \textit{ground truth object frame} as a local object frame that is common to all systems.
This metric can be easily calculated from the observed motion representation using~\eqref{equ:observed_motion_to_gt_body_motion} or if the transform between the estimated and ground truth object frame is well defined. 
For each object $j$ at $k$, ME is defined as: 
\begin{equation}
     \mathbf{ME}_k =\objmotion{\objf_{\text{\textcolor{red}{gt}}, k-1}}{k-1}{\text{\textcolor{red}{gt}}, k}^{-1}\: \objmotion{\objf_{\text{\textcolor{red}{gt}}, k-1}}{k-1}{k}
\label{equ:rme_metric}
\end{equation}
where
\begin{equation}
    \objmotion{\objpose{\worldf}{\text{\textcolor{red}{gt}}, k-1}}{k-1}{k} = \objpose{\worldf}{\text{\textcolor{red}{gt}}, k-1}^{-1} \: \objmotion{\worldf}{k-1}{k} \: \objpose{\worldf}{\text{\textcolor{red}{gt}}, k-1}\text{.}
\label{equ:observed_motion_to_gt_body_motion}
\end{equation}
Conceptually, ME compares the ground truth motion in the ground truth frame with the estimated motion also expressed in the ground truth object frame. 
Equation~\eqref{equ:observed_motion_to_gt_body_motion} demonstrates how the estimated motion $\objmotion{\worldf}{k-1}{k}$ can be expressed in the ground truth object frame using only ground truth object poses.

While both RPE and ME evaluate the accuracy in relative transformations along an objects trajectory, ME is agnostic to how each system defines the object frame.
By accounting for any differences in definitions, this metric facilitates valid comparisons between systems.  
As most existing systems primarily evaluate RPE for objects, we report both RPE and ME metrics. 
However, we primarily focus on analysing ME as we believe this is a better indicator of accuracy. 

As before, we report the RMSE for the translation and rotation error throughout the sequence for all objects, following~\eqref{equ:rmse}.
We report average motion error for each sequence. However, DynoSAM also outputs per-object errors for more detailed comparison if necessary. 
Only objects observed for at least $3$ consecutive frames are included.

\subsubsection{Frame Alignment}
To ensure our framework performs fair and accurate evaluation, 
it is vital that we correctly align the world, the camera and object frames with their respective ground truth frames.
During experiments we align the ground truth origin to the estimated trajectory using Umeyama’s method~\cite{umeyama1991least}.
For camera pose evaluation, the estimated and ground truth states share a common and well defined reference frame, 
making alignment and comparison simple. 
To facilitate a valid evaluation of RPE, 
we define the start of each object trajectory using the first ground truth object pose, i.e.~\eqref{equ:rpe_expand} to ensure a common object frame definition. 
To guarantee that the estimated objects also share the same world frame, 
we calculate $\objmotion{\worldf}{k-1}{\text{\textcolor{red}{gt}}, k}$ and $\objpose{\worldf}{\text{\textcolor{red}{gt}}, k}$ directly from the dataset using the newly aligned ground truth odometry. 
% This ensures that both the world and object frames are defined correctly for each sequence, facilitating a valid comparison.

\subsection{Camera Pose Error}
\tabref{tab:camera_experiments}, \tabref{tab:camera_experiments_tas} and \tabref{tab:camera_experiments_viode} show comparisons of camera pose estimation using ATE and RPE.
% The accuracy of DynoSAM's camera pose estimation is evaluated using ATE and RPE as shown in~\tabref{tab:camera_experiments}, \tabref{tab:camera_experiments_tas} and \tabref{tab:camera_experiments_viode}.
Our results demonstrate that in RPE, we predominantly achieve better results than all other systems.
In cases where we perform worse, the difference is marginal with the maximum error difference being \SI{0.02}{\degree} in rotation and \SI{0.02}{\meter} in translation. 

ATE is a measure of global consistency and accumulated drift. 
Across all datasets we demonstrate comparable and in some cases better results than other systems. 
Notably on the TartanAir (Shibuya) dataset, we perform better than VDO-SLAM in all cases and AirDOS in most sequences, as shown in~\tabref{tab:camera_experiments_tas}, 
demonstrating our robustness in environments with a large number of (non-rigid) dynamic objects.
DynaSLAM II produces marginally better results over KITTI Sequence $03$--$18$ in ATE.
We believe this is due to our more conservative motion segmentation algorithm which results in relatively fewer static tracks, particularly in scenes with many objects. This introduces greater rotation bias into the visual odometry, notably on longer sequences, thus increasing the ATE.

The VIODE dataset is challenging due to the camera's complex motions and scenarios of full occlusion by dynamic objects. 
% In these conditions many instance segmentation algorithms will fail and the work of DynaVINS~\cite{Song2022ral_dynavins} demonstrate that IMU integration is vital as no static features are available. 
Song~\etal have demonstrated with their Visual-Inertial Odometry (VIO) system, DynaVINS~\cite{Song2022ral_dynavins}, that, in these situations, tight IMU integration is vital \textcolor{red}{for} robust ego-motion estimation as no static features are available. 
We therefore tightly integrate the IMU into our backend using the preintegration method of Forster~\etal\cite{Forster16tro} for the VIODE experiments. 
This highlights the flexibility of our framework to easily integrate inertial data. 
\tabref{tab:camera_experiments_viode} shows DynoSAM's superior RPE on the majority of sequences.
% Our results re-emphasize the need for IMU in these challenging sequences as, without ego-motion information from IMU, our estimated trajectory does diverge upon full occlusion. 
% Without the additional ego-motion information from IMU our estimated trajectory diverges upon full occlusion, as indicated by $^\dagger$ in~\tabref{tab:camera_experiments_viode}. These results re-emphasize the need for the IMU in these challenging sequences~\cite{Song2022ral_dynavins}.
Compared with ORB-SLAM3~\cite{Campos2021tro} (running in IMU mode), we demonstrate superior robustness as DynoSAM completes all sequences while ORB-SLAM3 fails on three, as indicated by $^\ast$.
% We believe that our semantic segmentation helps 
% We believe that our  robustness is because of our semantic segmentation prior.
While we outperform DynaVINS~\cite{Song2022ral_dynavins} in RPE, DynaVINS produces the best ATE, 
likely due to its comprehensive mechanism that rejects dynamic object features using IMU-informed pose priors.
Consequently, DynaVINS is more robust during the transition phase of total occlusion. 
These situations occur in all VIODE sequences and biases DynoSAM's eventual ATE, but do not affect our RPE accuracy on average. 
However, we believe that DynaVINS's VIO mechanism is complimentary to our system and can be integrated in the future to improve overall accuracy and robustness.
Running DynoSAM without the IMU results in the estimation diverging, as indicated by $^\dagger$ in~\tabref{tab:camera_experiments_viode}. We omit the numerical results due to space limitations and because it is clear from the DynaVINS paper that the IMU is needed in these challenging conditions~\cite{Song2022ral_dynavins}.

Finally, both proposed DynoSAM formulations demonstrate almost identical performance. This is expected as they share a common formulation for visual odometry. 

\begin{table}[t]
\footnotesize
\centering
\setlength{\tabcolsep}{3pt}
\caption{\footnotesize{ATE evaluation of camera pose on TartanAir (Shibuya)~\cite{Qiu2022icra_airdos} compared against AirDOS~\cite{Qiu2022icra_airdos} (with mask) and VDO-SLAM~\cite{zhang2020vdoslam}.}}.
\label{tab:camera_experiments_tas}
\begin{tabular}{cc|ccccccc}
\toprule
 & & \multicolumn{7}{c}{TartanAir Shibuya} \\
 & & \multicolumn{2}{c}{\makecell{Standing \\Human}} & \multicolumn{3}{c}{\makecell{Road Crossing\\(Easy)}} &  \multicolumn{2}{c}{\makecell{Road Crossing\\(Hard)}}\\
 & & I & II & III & IV & V & VI & VII  \\
\midrule
\midrule
\parbox[t]{2mm}{\multirow{6}{*}{\rotatebox[origin=c]{90}{\qquad ATE (\si{\meter})}}}
% \multirow{5}{*}{ATE(\si{\meter})} 
& AirDOS & $0.06$ & $\mathbf{0.02}$ & $0.10$ & $0.03$ & $\mathbf{0.02}$ & $0.22$ & $0.56$ \\
& VDO-SLAM & $0.10$ & $0.61$ & $0.38$ & $0.39$ & $0.22$ & $0.24$ & $0.66$ \\
& WCPE (ours) & $\underline{0.03}$ & $\underline{0.03}$ & $\mathbf{0.02}$ & $\underline{0.02}$ & $\underline{0.03}$ & $\underline{0.04}$ & $\mathbf{0.18}$ \\
& WCME (ours) & $\mathbf{0.02}$ & $0.04$ & $\mathbf{0.02}$ & $\mathbf{0.02}$ & $0.04$ & $\mathbf{0.03}$ & $\underline{0.19}$ \\
\bottomrule
\end{tabular}
% \vspace{-4mm}
\end{table}

Overall we demonstrate our camera pose estimation is accurate and robust in various highly dynamic environments that include rigid and non-rigid objects, full occlusion and complex camera motion. 
We outperform most systems on RPE and produce comparable if not better ATE results. 
The IMU integration on the challenging VIODE dataset further demonstrates the flexibility of our framework and the robustness of our algorithm under these challenging conditions.

\subsection{Object Motion and Pose Errors}
\label{sec:object_motion_errors}

\tabref{tab:object_experiments} report object motion and pose errors across all sequences. ~\tabref{tab:avg_percent_improvement} summarizes the average percentage improvement of our approach over existing systems to highlight our framework's consistent accuracy. 
% We summarize the average percentage improvement of our approach over existing systems in~\tabref{tab:avg_percent_improvement} to highlight our framework's consistent accuracy.
As discussed previously, we have found comprehensive comparison challenging due to the limited number of open-source systems and sequences that each system reports on. 
We report both RPE and ME evaluations, but we focus on ME since we believe this metric is more accurate for evaluation between systems (\secref{sec:motion_metrics}).
However, existing systems do not report ME and many are closed-source.
We use `-' to indicate sequences where results could not be obtained. 

% Overall our results demonstrate state-of-the-art estimation in both ME and RPE metrics. 

ClusterSLAM~\cite{Huang2019iccv} only reports RPE as an accumulated average.
 \tabref{tab:avg_percent_improvement} shows that our framework is upwards of   \textcolor{red}{$6$} times more accurate in both rotation and translation, demonstrating our superior performance.
% Given ClusterSLAM only reports accumulated error, it is difficult to perform further comparisons. 
% However, the significant difference in accuracy between the two systems strongly indicates DynoSAM's superior performance. 

\begin{table}[t]
\footnotesize
\centering
\setlength{\tabcolsep}{3pt}

\begingroup % To limit the scope of \arraystretch
\renewcommand{\arraystretch}{1.1} % Adjust this value as needed

\caption{\footnotesize{{ATE and RPE evaluation of camera pose on VIODE~\cite{minoda2021viode} compared against DynaVINS~\cite{Song2022ral_dynavins} and ORB-SLAM3~\cite{Campos2021tro}. Both systems are run in stereo-inertial mode. DynoSAM is run with full IMU fusion in the backend and sequences marked with $^\dagger$ indicate failure cases (i.e. trajectory divergence) without the use of the IMU. $^\ast$ indicates total failure cases. Mid/High labels indicate sequence difficulty and number of dynamic objects.}}}
\label{tab:camera_experiments_viode}
\begin{tabular}{cc|cccccc}
\toprule
 & & \multicolumn{6}{c}{VIODE} \\
 & & \multicolumn{2}{c}{City Day} & \multicolumn{2}{c}{City Night} &  \multicolumn{2}{c}{Parking Lot}\\
 & & Mid & High$^{\dagger}$ & Mid & High$^{\dagger}$ & Mid$^{\dagger}$ & High$^{\dagger}$  \\
\midrule
\midrule
\parbox[t]{2mm}{\multirow{4}{*}{\rotatebox[origin=c]{90}{ATE (\si{\meter})}}}
% \multirow{5}{*}{ATE(\si{\meter})} 
& DynaVins & $\mathbf{0.104}$ & $\mathbf{0.150}$ & $\mathbf{0.194}$ & $\mathbf{0.147}$ & $\mathbf{0.056}$ & $\mathbf{0.065}$ \\
& ORB-SLAM3 & $\underline{0.217} $ & $\ast$ & $1.693 $ & $3.006$ & $\ast$ & $\ast$ \\
& WCPE (ours) & $2.515$ & $\underline{2.128}$ & $\underline{1.360}$ & $\underline{2.560}$ & $\underline{1.377}$ & $\underline{0.764}$ \\
& WCME (ours) & $2.515$ & $\underline{2.128}$ & $\underline{1.360}$ & $\underline{2.560}$ & $\underline{1.377}$ & $\underline{0.764}$ \\
\midrule
\parbox[t]{2mm}{\multirow{3}{*}{\rotatebox[origin=c]{90}{\qquad $\text{RPE}_t$(\si{\meter})}}}
& DynaVins & $0.024$ & $0.027$ & $0.019$ &  $\mathbf{0.023}$ & $0.019$ & $0.015$ \\
& WCPE (ours)   & $\mathbf{0.008}$ & $\mathbf{0.014}$ & $\mathbf{0.015}$ & $0.020$ & $\mathbf{0.006}$ & $\mathbf{0.005}$ \\
& WCME (ours)  & $\mathbf{0.008}$ & $\mathbf{0.014}$ & $\mathbf{0.015}$ & $0.020$ & $\mathbf{0.006}$ & $\mathbf{0.005}$ \\
\midrule
\parbox[t]{2mm}{\multirow{3}{*}{\rotatebox[origin=c]{90}{\qquad $\text{RPE}_r$(\si{\degree})}}}
& DynaVins &  $0.087$ & $\mathbf{0.09}$ & $0.102$ & $\mathbf{0.096}$ & $0.126$ & $0.111$\\
& WCPE (ours) & $\mathbf{0.049}$ & $0.105$ & $\mathbf{0.070}$ & $0.190$ & $\mathbf{0.036}$ & $\mathbf{0.040}$ \\
& WCME (ours) & $\mathbf{0.049}$ & $0.105$ & $\mathbf{0.070}$ & $0.190$ & $\mathbf{0.036}$ & $\mathbf{0.040}$ \\
\bottomrule
\end{tabular}

\endgroup
\vspace{-4mm}
\end{table}

\begin{table*}[t]
\footnotesize
\centering
\setlength{\tabcolsep}{5.3pt}
\caption{\footnotesize{Quantitative evaluations of average object motions (ME) and poses (RPE) for all experiments. Entries marked with `-' indicate that results are not available for that sequence. Best results are marked in bold and second best with an underscore.}}
\label{tab:object_experiments}
\begin{tabular}{cc|ccccccccc|>{}c>{}c>{}c>{}c>{}c|c}
% \begin{tabular}{cc|ccccccccc|ccccc|c}
\toprule
 & & \multicolumn{9}{c|}{KITTI} & \multicolumn{5}{c|}{Outdoor Cluster} & OMD \\
 & & $00$ & $01$ & $02$ & $03$ & $04$ & $05$ & $06$ & $18$ & $20$ & L1 & L2 & S1 & S2 & avg. & S4U \\
\midrule
\midrule
\parbox[t]{2mm}{\multirow{5}{*}{\rotatebox[origin=c]{90}{$\text{ME}_r$(\si{\degree})}}}
% \multirow{5}{*}{$\text{RME}_r$(\si{\degree})} 
 & VDO-SLAM & $1.38$ & $2.15$ & $1.68$ & $0.39$ & $2.8$ & $\mathbf{0.48}$ & $\underline{2.8}$ & $\mathbf{0.36}$ & $0.47$ & - & - & - & - & - & $\underline{0.96}$ \\
 & ClusterSLAM & - & - & - & - & - & - & - & - & - & - & - & - & - & - & - \\
 & MVO & $3.36$ & - & - & - & - & - & - & - & - & - & - & - & - & - & $1.1$ \\
 & WCPE (ours) & $\mathbf{1.23}$ &$\underline{0.91}$ & $\mathbf{0.95}$ & $\underline{0.27}$ & $\mathbf{0.76}$ & $0.56$ & $\underline{2.8}$  & $1.15$ & $\underline{0.39}$ & $\underline{0.86}$ & $\underline{0.74}$ & $\mathbf{0.68}$ & $\underline{2.40}$ & $\underline{1.17}$ & $1.6$ \\
 & WCME (ours) & $\underline{1.29}$ & $\mathbf{0.86}$ & $\underline{1.06}$ & $\mathbf{0.26}$ & $\underline{1.01}$ & $\underline{0.49}$ & $\mathbf{0.39}$ & $\underline{0.6}$ & $\mathbf{0.33}$ & $\mathbf{0.82}$ & $\mathbf{0.70}$ & $\underline{0.69}$ & $\mathbf{2.36}$ & $\mathbf{1.14}$  & $\mathbf{0.71}$ \\
\midrule
\parbox[t]{2mm}{\multirow{5}{*}{\rotatebox[origin=c]{90}{$\text{ME}_t$(\si{\meter})}}}
% \multirow{5}{*}{$\text{RME}_t$(\si{\meter})} 
 & VDO-SLAM & $\underline{0.11}$ & $\underline{0.35}$ & $\underline{0.43}$ & $\mathbf{0.15}$ & $0.38$ & $0.19$ & $\mathbf{0.11}$ & $\mathbf{0.16}$ & $0.57$ & - & - & - & - & - & $\mathbf{0.02}$ \\
 & ClusterSLAM & - & - & - & - & - & - & - & - & - & - & - & - & - & - & - \\
 & MVO & $0.27$ & - & - & - & - & - & - & - & - & - & - & - & - & - & $\underline{0.03}$ \\
 & WCPE (ours) & $\mathbf{0.09}$ & $0.40$ & $0.73$ & $\mathbf{0.15}$& $\underline{0.1}$ & $\underline{0.14}$ & $\underline{0.22}$ & $0.31$ & $\underline{0.39}$  & $\mathbf{0.07}$ & $\underline{0.08}$ & $\mathbf{0.03}$ & $\mathbf{0.13}$ & $\mathbf{0.08}$ & $0.07$ \\
 & WCME (ours) & $0.15$ & $\mathbf{0.34}$ & $\mathbf{0.4}$ & $\mathbf{0.15}$ & $\mathbf{0.09}$ & $\mathbf{0.13}$ & $\mathbf{0.11}$ & $\underline{0.20}$ & $\mathbf{0.05}$& $\underline{0.08}$ & $\mathbf{0.06}$ & $\underline{0.04}$ & $\underline{0.15}$ & $\mathbf{0.08}$ & $\mathbf{0.02}$ \\
\midrule
\midrule
\parbox[t]{2mm}{\multirow{5}{*}{\rotatebox[origin=c]{90}{$\text{RPE}_r$(\si{\degree})}}}
% \multirow{5}{*}{$\text{RPE}_r$(\si{\degree})}
 & VDO-SLAM & $1.40$ & $1.25$ & $1.34$ & $0.32$ & $\underline{1.04}$ & $0.64$ & $\mathbf{1.49}$ & $\mathbf{0.38}$ & $0.47$ & - & - & - & - & - & $3.3$ \\
 & ClusterSLAM & - & - & - & - & - & - & - & - & - & - & - & - & - & $10.3$ & - \\
 & MVO & $2.7$ & - & - & - & - & - & - & - & - & - & - & - & - & - & $\mathbf{3.1}$ \\
 & WCPE (ours) & $\mathbf{1.27}$ & $\underline{0.89}$ & $\mathbf{1.02}$ & $\mathbf{0.27}$ & $\mathbf{0.74}$ & $\mathbf{0.6}$ & $3.00$ & $1.38$ & $\underline{0.37}$& $\underline{1.7}$ & $\underline{2.34}$ & $\mathbf{0.61}$ & $\underline{2.07}$ & $\underline{1.68}$ & $4.1$ \\
 & WCME (ours) & $\underline{1.38}$ & $\mathbf{0.80}$ & $\underline{1.06}$ & $\mathbf{0.27}$ & $\underline{1.04}$ & $\underline{0.62}$ & $\underline{2.74}$ & $\underline{1.16}$ & $\mathbf{0.33}$ & $\mathbf{1.3}$ & $\mathbf{2.28}$ & $\underline{0.67}$ & $\mathbf{2.00}$ & $\mathbf{1.56}$ & $\underline{3.2}$ \\
\midrule
\parbox[t]{2mm}{\multirow{5}{*}{\rotatebox[origin=c]{90}{$\text{RPE}_t$(\si{\meter})}}}
% \multirow{5}{*}{$\text{RPE}_t$(\si{\meter})}
 & VDO-SLAM & $\underline{0.28}$ & $\underline{0.34}$ & $\mathbf{0.30}$ & $\underline{0.20}$ & $\underline{0.94}$ & $\underline{0.17}$ & $\mathbf{0.46}$ & $\mathbf{0.13}$ & $\mathbf{0.09}$ & - & - & - & - & - & $0.06$ \\
 & ClusterSLAM & - & - & - & - & - & - & - & - & - & - & - & - & - & $8.65$ & - \\
 & MVO & $\underline{0.28}$ & - & - & - & - & - & - & - & - & - & - & - & - & - & $\underline{0.05}$ \\
 & WCPE (ours) & $0.43$ & $1.18$ & $2.0$ & $0.67$ & $1.28$ & $2.6$ & $1.84$  & $2.17$ & $\mathbf{0.09}$ & $\mathbf{1.84}$ & $\mathbf{0.74}$ & $\mathbf{0.99}$ & $\underline{1.63}$ & $\underline{1.33}$ & $0.06$ \\
 & WCME (ours) & $\mathbf{0.27}$ & $\mathbf{0.32}$ & $\underline{0.79}$ & $\mathbf{0.19}$ & $\mathbf{0.92}$ & $\mathbf{0.16}$ & $\underline{0.48}$ & $\underline{0.2}$ & $\underline{0.12}$ & $\underline{1.9}$ & $\underline{0.72}$ & $\underline{0.94}$ & $\mathbf{1.50}$ & $\mathbf{1.27}$ & $\mathbf{0.04}$ \\
\bottomrule
\end{tabular}
\end{table*}

\begin{table}[t]
\footnotesize
\centering
\setlength{\tabcolsep}{3pt}
\caption{\footnotesize{Average percentage improvement of DynoSAM Motion estimator compared to state-of-the-art Dynamic SLAM systems. Green cell blocks indicate improvement over existing systems while red blocks indicate relatively worse accuracy.}}
\begin{tabular}{c|c|c|c}
\toprule
 & 
VDO-SLAM~\cite{zhang2020vdoslam} &
ClusterSLAM~\cite{Huang2019iccv} &
MVO~\cite{judd2024ijrr_mvo} \\
\midrule
\midrule

% %seq E(r)     E(t)$$
% \multirow{1}{*}{\makebox[33pt]{$\text{AME}_r$(\si{\degree})}} & \cellcolor{nicebluishgreen}$+$\SI{31}{\percent} & $-$ & \cellcolor{nicebluishgreen}$+$\SI{48}{\percent} \\
% \multirow{1}{*}{\makebox[30pt]{$\text{AME}_t$(m)}} & \cellcolor{nicevermillion}$-$\SI{35}{\percent} & $-$ & \cellcolor{nicebluishgreen}$+$\SI{27}{\percent} \\
% \midrule
\multirow{1}{*}{\makebox[30pt]{$\text{ME}_{r}$(\si{\degree})}} & \cellcolor{nicebluishgreen}$+28\%$ & $-$ & \cellcolor{nicebluishgreen}$+49\%$ \\
\multirow{1}{*}{\makebox[30pt]{$\text{ME}_{t}$(m)}} & \cellcolor{nicebluishgreen}$+9\%$ & $-$ & \cellcolor{nicebluishgreen}$+39\%$ \\
\midrule
\multirow{1}{*}{\makebox[30pt]{$\text{RPE}_r$(\si{\degree})}} & \cellcolor{nicevermillion}$-9\%$ & \cellcolor{nicebluishgreen}{$+660\%$} & 
\cellcolor{nicevermillion}$-9\%$ \\
\multirow{1}{*}{\makebox[30pt]{$\text{RPE}_t$(m)}} & \cellcolor{nicebluishgreen}$+4\%$ & \cellcolor{nicebluishgreen}{$+681\%$} & \cellcolor{nicebluishgreen}$+12\%$ \\

\bottomrule
\end{tabular}
\label{tab:avg_percent_improvement}
\vspace{-4mm}
\end{table}

\begin{table}[t]
\footnotesize
\centering
\setlength{\tabcolsep}{3pt}
\caption{\footnotesize{{Per-object ME and ATE errors on the OMD (S4U) using the WCME. Object ID's correspond with the following object's in the dataset - 1: top left, 2: top right, 3: bottom left, 4: bottom right.}}}
\label{tab:mvo_errors}
\begin{tabular}{cc|cccc}
\toprule
 \multicolumn{2}{c|}{} & $1$ & $2$ & $3$ & $4$ \\
\midrule
\midrule
\multirow{3}{*}{$\text{ME}_r$(\si{\degree})}
% \multirow{5}{*}{$\text{RME}_r$(\si{\degree})} 
 & VDO-SLAM & $1.256$ & $\underline{0.770}$ & $\underline{0.907}$ & $0.927$  \\
 & MVO & $\mathbf{0.542}$ & $0.843$ & $1.648$ & $\underline{0.854}$  \\
 & DynoSAM (ours) & $\underline{1.138}$ & $\mathbf{0.544}$ & $\mathbf{0.443}$ & $\mathbf{0.474}$ \\
\midrule
\multirow{3}{*}{$\text{ME}_t$(\si{\meter})}
 & VDO-SLAM & $0.0243$ & $\underline{0.0234}$ & $\underline{0.0148}$ & $\underline{0.0293}$  \\
 & MVO & $\mathbf{0.0169}$ & $0.0269$ & $0.0232$ & $0.0309$ \\
 & DynoSAM (ours) & $\underline{0.0214}$ & $\mathbf{0.0233}$ & $\mathbf{0.0086}$ & $\mathbf{0.0291}$ \\
\midrule
\midrule
\multirow{2}{*}{$\text{ATE}$(\si{\meter})}
% \multirow{5}{*}{$\text{RPE}_r$(\si{\degree})}
 & DynaSLAM II & $0.41$ & $0.37$ & $1.09$ & $0.28$ \\
 & DynoSAM (ours) & $\mathbf{0.09}$ & $\mathbf{0.21}$ & $\mathbf{0.08}$ & $\mathbf{0.15}$ \\
\bottomrule
\end{tabular}
\vspace{-4mm}
\end{table}

\begin{figure*}
    \includegraphics[trim={0.0cm 0.0cm 0.0cm 0cm},clip,width=\textwidth]{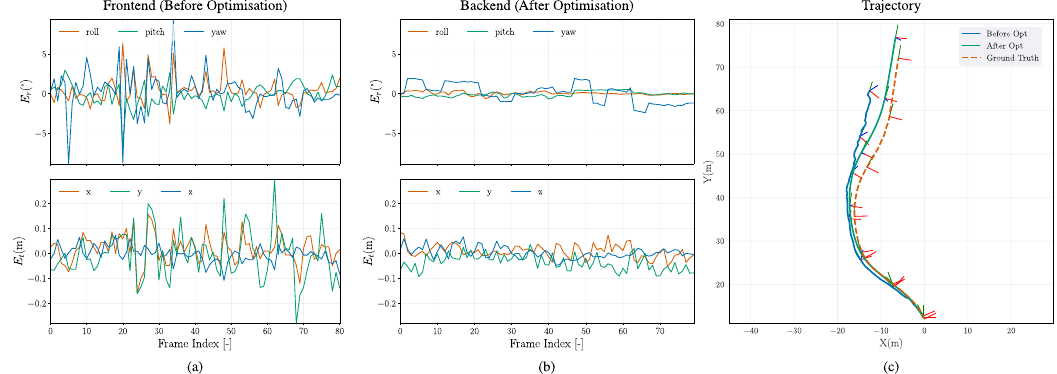}
 % \caption{\small{Translation and rotation ME for Object $2$ (cyclist) on KITTI $00$ using the motion estimator. Results before and after optimisation (\textbf{a} and \textbf{b} respectively) are shown  per frame.
 % % The improvement in per-frame error after optimisation is evident in both cases and the reduction in error scale should be noted.
 % }}
    \caption{\footnotesize{Example of per-frame ME before (\textbf{a}) and after (\textbf{b}) optimization using the WCME. We show results for object $2$, a non-rigid body (cyclist) on KITTI $00$. Translation and rotation errors are shown individually; along the top and bottom rows respectively. The resulting trajectories are compared to the ground truth in (\textbf{c})}. These results demonstrate our systems capacity to handle non-rigid bodies in the case where the object experiences a predominant motion.}
    \label{fig:motion_errors_fontend_backend_comparison}
    \vspace{-6mm}
\end{figure*}

\tabref{tab:avg_percent_improvement} demonstrates that DynoSAM outperforms MVO~\cite{judd2024ijrr_mvo} across all metrics, except in $\text{RPE}_r$ on the OMD where MVO is more accurate by a small margin. 
Our framework is particularly accurate in translation, 
with a $39\%$ improvement in ME$_t$ and $12\%$ improvement in RPE$_t$.
However, reporting the ME individually for each object in the S4U sequence in~\tabref{tab:mvo_errors} shows that that DynoSAM outperforms MVO on most objects, thereby reinforcing our accuracy even on this highly dynamic dataset. 
We additionally outperform VDO-SLAM on all objects.
Compared to DynaSLAM II we show an average $445\%$ improvement in object ATE.

% While in general we outperform VDO-SLAM~\cite{zhang2020vdoslam} on the KITTI dataset, there are some situations that require more investigation such as KITTI $05$, $06$ and $18$. While KITTI $05$ the error difference is marginal, the other sequences have more signifcant difference. While VDO-SLAM shares a similar motion formulation to WCME, different solvers (ie g2o~\cite{Kummerle11icra} vs GTSAM~\cite{gtsam}) are used in both the frontend and backend and the mechanism for managing object tracking is different across systems. By observation this results in different lengths of tracking, and therefore trajectories, between the two systems. As we report the average over all objects, the more objects in each sequence the more likely these differences are to affect the results. We believe this is a key reason for the difference in results on these sequences. In future we could further explore tracking quality by including the MOTP metric~\cite{bescos2021ral}, however, this is outside the main scope of this paper.

\begin{figure}[b]
    \vspace{-4mm}
    \centering
    \includegraphics[trim={0.0cm 0.0cm 0.0cm 0cm},clip,width=\linewidth]{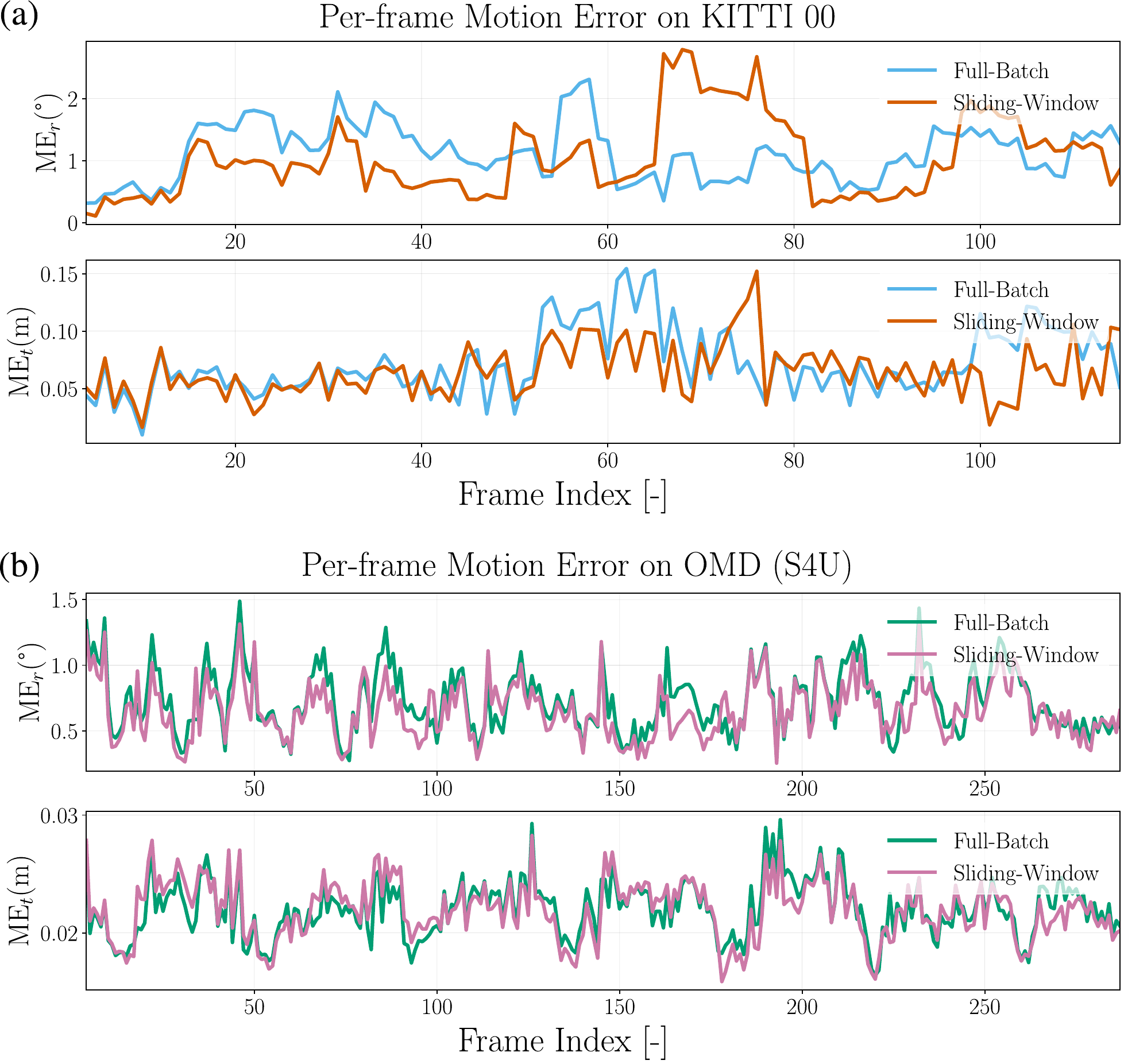}
   \caption{\footnotesize{Object motion errors comparing full-batch vs sliding window. We take the $\mathbf{ME}$ (rotation and translation) for each object \textit{per frame} and report the average for KITTI $00$ \textbf{(a)} and OMD (S4U) \textbf{(b)}. The WCME is used for both sets of results. }}
    \label{fig:batch_vs_sliding_window_ame}
\end{figure}

Compared to VDO-SLAM~\cite{zhang2020vdoslam}, our framework in general exhibits superior accuracy in object motion and pose estimation.
Such improvement is particularly pronounced on the OMD, where DynoSAM is $34\%$ more accurate in rotation and $21\%$ in translation errors, as measured by the ME metric. 
However, there remains some situations such as KITTI $05$, $06$ and $18$ where VDO-SLAM performs better.
While the difference on KITTI $05$ is marginal, $06$ and $18$ exhibit larger performance differences that merit further investigation.
Although these two systems share a similar WCME formulation and are expected to yield comparable results, they differ in the solvers they use and in their underlying approaches to object tracking.
Since we report performance averaged across all objects, such variations can significantly affect the final results, especially in sequences with many objects (KITTI $18$). 
We believe this is a key reason for the discrepancies in the corresponding VDO-SLAM and DynoSAM performances. 
In the future, we will explore tracking quality by including the MOTP metric~\cite{bescos2021ral}; however, this is outside the main scope of this paper.

% Finally we consider a comparison between WCME and WCPE. 
% Overall, WCME is \SI{40}{\percent} more accurate in $\text{ME}_t$ and \SI{135}{\percent} in $\text{ME}_r$ compared to the WCPE.
% Furthermore, WCME estimates the object trajectory more accurately, as indicated by the lower RPE. 
% However, across most sequences our presented formulations provides comparable results with marginal differences, especially given that the travel distance in each sequence is on the scale of hundreds of meters to several kilometers, 
% while the differences in motion estimations are on the scale of a few centimeters and degrees. 
% We believe that the slightly worse performance of the WCPE is due to an inherent sensitivity to the initalisation of object pose, a issue not present in the WCME. 
% This often the case since we initalise the object pose at each time-stamp with the centroid of the observed object points. 
% These observations are often minimal and represent only a partial view of the object and therefore
% do not guarnatee initalisation is consistent with previous poses and/or motions. This may be corrected by utilising pose-prediction algorithms for inital values and to act as measurements.
% While we did not necessarily expect this formulation to outperform the WCME in the context of motion estimation there exists many downstream applications that require the direst estimation of object pose. Therefore we believe that it remains highly relevant to introduce and examine the performance of an object pose estimator despite the slight degredation in performance.

Finally we compare between WCME and WCPE. 
% Overall, WCME is $79\%$ more accurate in translational motion ($\text{ME}_t$) and $80\%$ more accurate in rotational motion ($\text{ME}_r$).
Across all sequences the average error difference is only \SI{0.06}{\meter} in translation and \SI{0.11}{\degree} in rotation, demonstrating comparable performance. Out of the $14$ sequences tested there are only three sequences (KITTI $03$, $06$ and $20$) where the difference in error is greater than one standard-deviation, calculated independently for translation and rotation across all sequences. 
% The marginal difference in performance is particularly notable given that the average travel distance in each KITTI and Outdoor Cluster sequence is on the scale of hundreds of meters to several kilometers.
In these instances, WCPE’s slightly degraded performance is likely due to its sensitivity to object pose initialization.

In WCPE, object poses are initialized at each time-step using the centroid of observed object points. 
Since these observations often represent only a partial view of the object, the resulting initialization may not be consistent with previous estimates, introducing additional noise. 
This limitation can be mitigated in future work by using pose prediction methods to provide better priors or serve as complementary measurements.
While WCPE is not necessarily expected to outperform WCME in motion estimation specifically, it remains highly relevant, 
as direct pose estimation is attractive for many downstream applications such as navigation~\cite{salzmann2020trajectron++} and reconstruction~\cite{Wang2025icra_dynorecon}.
Therefore, we believe that it remains relevant to introduce WCPE and examine its performance despite the slight degradation in motion accuracy.

Finally, \figref{fig:motion_errors_fontend_backend_comparison} compares the per-frame ME before and after optimization on object $2$ (cyclist) from KITTI $00$, highlighting our accurate estimation across the entire trajectory.
These results further demonstrate our systems capability to deal with only partially rigid objects, i.e. a cyclist, despite its motion breaking the explicit rigid-body assumption. 
We hypothesize that as long as the object experiences a single predominant motion, and sufficient object points can be tracked, our method will recover the overall trajectory, as shown in~\figref{fig:motion_errors_fontend_backend_comparison}~(c).

\subsection{Sliding Window Optimization}
\label{sec:sliding_window_opt}

We have so far presented results from a full-batch solution that incorporates all measurements into a single optimization. 
This solution, while accurate, is computationally intensive and unsuitable for online use. 
Moreover, we hypotheise that object motions should exhibit increasing independence over extended time horizons, even for objects with strong motion priors, such as vehicles on highways, thus making optimization over the global trajectory potentially unnecessary. 

Therefore, we implemented a sliding-window approach that solves smaller, more efficient batch problems every $w$ frames, reusing estimates from the previous window to initialize overlapping variables. As shown in~\tabref{tab:sliding_window_details_table} the sliding-window method achieves marginally better per-frame error. 
Furthermore, 
\figref{fig:batch_vs_sliding_window_ame} compares the per-frame error of each approach.
The average ME for each object on the KITTI $00$ and OMD sequences is reported.
Note that full-batch demonstrates slightly more consistent accuracy, as seen in~\figref{fig:batch_vs_sliding_window_ame}~(a) between frames $70$ and $85$. The larger error occurs at window overlap and is likely a result of poor front-end tracking at that frame. Our preliminary results indicate that incorporating motion data over a receding time horizon is beneficial.

\subsection{Computation Time}
\begin{table}[t]
\footnotesize
\centering
\setlength{\tabcolsep}{4.8pt}
\caption{\footnotesize{Average object  $\text{ME}$ comparison of full-batch vs. sliding window approach on the KITTI and OMD datasets.}}
\label{tab:sliding_window_details_table}
\begin{tabular}{c|cc|cc}
\toprule
 & 
 \multicolumn{2}{c|}{Full-Batch} & 
  \multicolumn{2}{c}{\makebox[15pt]{Sliding Window}} \\
 % \multicolumn{2}{c}{\makebox[0pt]{Motion-only estimator}} \\
\multirow{1}{*}{\makebox[0pt]{}} & $\text{ME}_{r}$(\si{\degree}) & $\text{ME}_{t}$(m) & $\text{ME}_{r}$(\si{\degree}) &$\text{ME}_{t}$(m) \\
\midrule
\midrule
% %seq E(r)     E(t)
\multirow{1}{*}{\makebox[35pt]{KITTI 00}} &  $1.11$ & $0.072$ & $\mathbf{1.039}$ & $\mathbf{0.065}$ \\
\multirow{1}{*}{\makebox[33pt]{OMD (S4U)}} &  $0.729$ & $0.022$ & $\mathbf{0.659}$ & $\mathbf{0.021}$ \\

\bottomrule
\end{tabular}

\vspace{-4mm}
\end{table}

\begin{figure}[t]
    \centering
  \includegraphics[width=0.9\columnwidth]{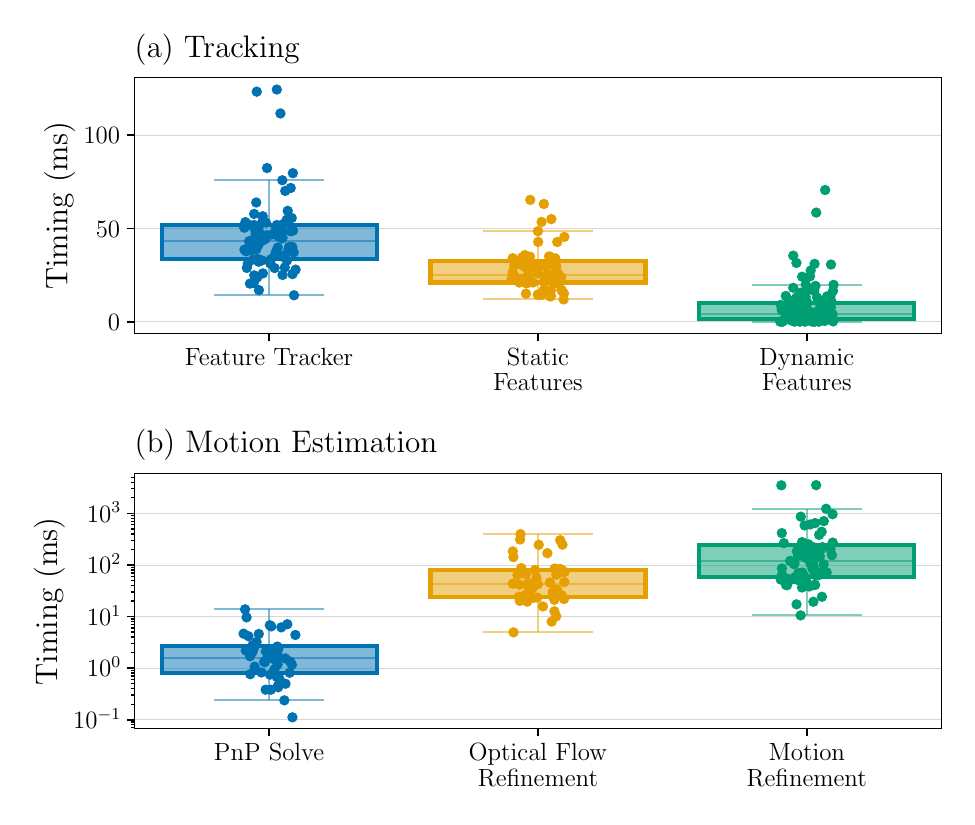}
   \caption{\footnotesize{Runtime breakdown of DynoSAM's front-end. \textbf{(a)} shows the timing for feature tracking and \textbf{(b)} shows the timing per each estimation module discussed in Sections \ref{sec:frontend_camera_pose_estimation}, \ref{sec:frontend_object_motion_estimation} (PnP Solve), \ref{sec:joint_optical_flow_refinement} (Joint Optical Flow) and \ref{sec:object_motion_refinement} (3D Motion Refinement). \textbf{(b)} uses log-scale for the timing axes.}}
    \label{fig:frontend_timing}
    \vspace{-4mm}
\end{figure}

\figref{fig:frontend_timing} presents the runtime breakdown for each module in the front-end. The feature tracking module, which includes feature extraction, matching, and geometric verification, requires less than $\SI{50}{\ms}$. The motion estimation module, responsible for estimating the camera pose and all object motions per frame, averages $\SI{100}{\ms}$ to produce all estimates. However, the majority of processing time is spent on the object motion refinement component, which takes approximately \SI{\sim250}{\ms} per object to solve the nonlinear optimization problem. The runtime of this component depends heavily on the number of points involved in the optimization, leading to significant variation in processing times.

In our experiments, the time required for full-batch optimization ranges from \SI{80}{\s} to \SI{700}{\s}, depending on the size of the constructed factor graph. 
% These values align with those reported in~\cite{morris2024icra}. 
While the full-batch optimization may not meet real-time requirements, it is important to note that practical applications typically rely on much smaller sliding window optimizations in the back-end. 
This approach significantly reduces computational demands and improves efficiency for online operation. 
In our experiments, using a window size of $20$, the average sliding-window optimization is \SI{\sim16}{\s}.
Future work will focus on enhancing the efficiency of the back-end, including techniques such as conditional variable elimination~\cite{carlone2014eliminating}, to better support real-time applications.

\section{Downstream Tasks}
The proposed Dynamic SLAM system accurately estimates the motions of dynamic objects, as validated in~\secref{sec:expe}. 
By eliminating the need for prior knowledge of motion models or object categories, our framework enables a range of downstream applications. 
This section explores how accurate motion estimation, produced by our pipeline, can facilitate dynamic object reconstruction and trajectory prediction, both of which are crucial for navigation systems.

\subsection{Dynamic Object Reconstruction}

Many reconstruction systems assume the environment to be static and rigid~\cite{Niessner2013tog, Kahler2015tvcg_infinitam}, 
as with static SLAM systems. 
Leveraging DynoSAM for object segmentation and motion estimation, 
we show that the output of DynoSAM can be directly used by a reconstruction system, such as DynORecon~\cite{Wang2025icra_dynorecon}, to incrementally map dynamic rigid-body objects. 

\begin{figure}[b]
    \vspace{-4mm}
    \centering
    \includegraphics[width=1.0\linewidth]{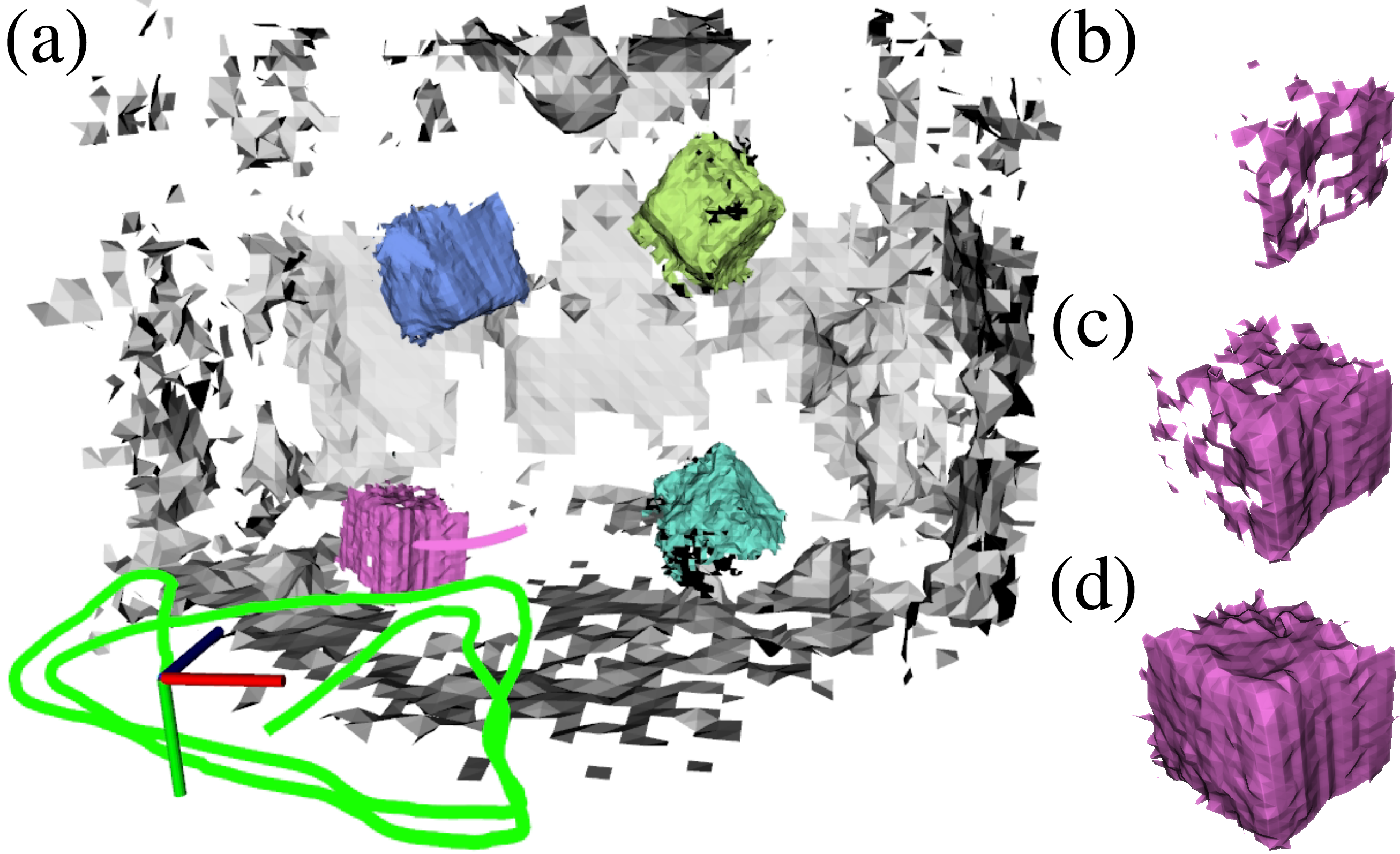}
    \caption{\footnotesize{Our downstream application, DynORecon~\cite{Wang2025icra_dynorecon}, incrementally builds up surface mesh reconstructions of both dynamic objects and static background in the OMD (S4U) experiment~\cite{Judd19ral}. \textbf{(a)}: dynamic object meshes and their trajectories (uniquely colored) as well as the static background (grey), in addition to camera pose and trajectory (green); \textbf{(b-d)}: the incrementally constructed mesh of Object $3$.}}
    \label{fig:dynorecon}
\end{figure}

The accurate motion estimation from DynoSAM facilitates an easy and efficient integration of new object measurements into each existing object reconstruction while maintaining object rigidity. 
Based on~\eqref{equ:point_motion_in_world}, 
applying $\objmotion{\worldf}{k-1}{k}$ as estimated by DynoSAM to all points on the object guarantees a consistent motion from time-step $k-1$ to $k$. 
Equation~\eqref{equ:pose_change_to_world_example} and~\eqref{equ:generic_object_motion_propogation} further show that the same motion can consistently transform any reference frame fixed to the object. 
Similar to the treatment of object pose in~\secref{sec:recipe_motion} and~\secref{sec:recipe_pose}, 
we can define an arbitrary reference frame that is rigidly attached to the object body without any prior knowledge of its pose, 
and use $\objmotion{\worldf}{k-1}{k}$ to move this body-fixed frame with respect to the global world frame.
This allows us to express all object points in the body-fixed frame, $\mpoint{\objf}{}$, where they are static and time-invariant with respect to their local frame. 
This representation allows new observations to be integrated into each object reconstruction while remaining consistent with previous measurements~\cite{Wang2025icra_dynorecon}.

\figref{fig:dynorecon} presents an example of DynORecon incrementally constructing all $4$ dynamic objects (free-floating cubes) in addition to the static background using DynoSAM's estimations in the OMD experiment~\cite{Judd19ral}. 
As shown in~\figref{fig:dynorecon} (b-d), accurate motion estimations from the upstream Dynamic SLAM system enables a consistent incremental surface reconstruction as the object undergoes complex movement. 
Building up a correct representation of dynamic objects, as more of them are observed, provides a more comprehensive understanding of moving obstacles in the scene, 
and is therefore beneficial to other robotic applications such as planning and navigation. 

\subsection{Object Trajectory Prediction}
\label{sec:object_trajectory_prediction}

\begin{figure}[t]
\centering
\includegraphics[width=\columnwidth]{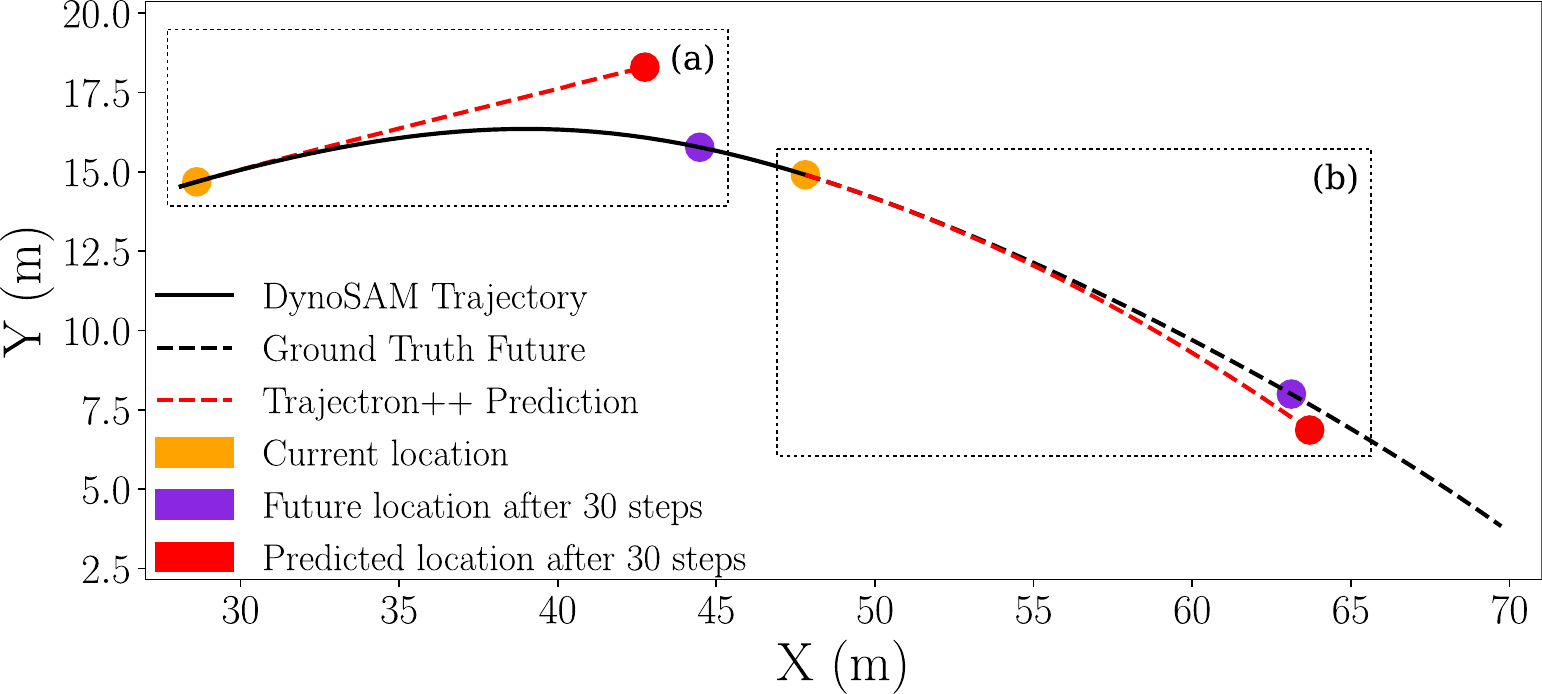}
\caption{\footnotesize{Predicted object trajectories based on the output estimation of DynoSAM. We show inferred predictions compared to the future ground truth trajectory of Object $2$ on KITTI $00$. Two snapshots are shown along the object's trajectory where \textbf{(a)} illustrates an initial prediction based on only $1$ previous state and \textbf{(b)} shows prediction based on $4$ prior states.}}
\label{fig:trajectronpp_predictions}
\vspace{-6mm}
\end{figure}

Trajectory prediction plays a key role in path planning and control in dynamic environments. 
Safer and more efficient navigation can be achieved by incorporating the anticipated future movements of dynamic objects into navigation algorithms~\cite{Hermann2014, Phillips2011icra_sipp, Finean2021icaps, Finean2023ras}.
 
% To test DynoSAM's capabilities for trajectory prediction, 
% To test DynoSAM's usability for a downstream task of trajectory prediction, 
We therefore integrated our DynoSAM pipeline with Trajectron++~\cite{salzmann2020trajectron++}, 
a state-of-the-art trajectory prediction algorithm that uses a graph-structured recurrent model to forecast trajectories. 
% Trajectron++ generates real-time 2D predictions as is typical for vehicle prediction problems and operates on state information including position, orientation and velocity for each object in the scene for both training and inference.
Trajectron++ uses current and historical state information, including pose and velocity for each object in the scene to generate real-time 2D predictions. 
% This state information is used as input during training and inference.
% uses 2D data as an input and 
% Additionally, it provides a Gaussian distribution for every time-step, capturing the uncertainty in predicted paths.
% Trajectron++ requires 2D state information including position, orientation, velocity, acceleration and category (e.g. vehicle or pedestrian) for each object in the scene.
% Trajectron++ requires 2D state information including position, orientation and velocity for each object in the scene.
% DynoSAM provides the object trajectory data for training and testing. 
% To provide this information for training and testing, we utilize the object trajectory data produced by DynoSAM.
Object poses are taken directly from DynoSAM's pose estimates
$\objpose{\worldf}{k}$, while velocities are calculated from the estimated motion $\objmotion{\worldf}{k-1}{k}$. 
This processed DynoSAM output is used to train and test \mbox{Trajectron++} on the KITTI tracking dataset (\secref{sec:datasets}). 
% Our resulting testing dataset included  $88$ input instances from KITTI $00$, while used $406$ training instances collected from the remaining seven KITTI sequences.
% To illustrate the effectiveness of our methods in a simplified setting, we limit our dataset to only include objects categorized as vehicles.
We trained the base Trajectron++ model from scratch within the adaptive prediction framework~\cite{ivanovic2023AdaptivePrediction}.

\figref{fig:trajectronpp_predictions} presents a snapshot of Object $2$'s (cyclist) predicted trajectory at frames $67$ and $103$ (a-b respectively). 
On average, our predictions closely track the ground truth trajectory without exhibiting significant overshooting. 
These results, while preliminary, demonstrate that estimates from our Dynamic SLAM framework are accurate enough for trajectory predictions tasks, 
and our prediction results are of sufficiently high accuracy to be leveraged for planning modules~\cite{Finean2021iros, Finean2023ras, king2022scate}, 
or as informative motion priors within our Dynamic SLAM framework.
% While the achieved predictions could be further improved by incorporating map generated by the DynORecon~\cite{Wang2025icra_dynorecon} module into Trajectron++, our predictions are already of high accuracy and could be leveraged to improve the motion planning module~\cite{Finean2021iros, Finean2023ras, king2022scate}, 
% or as informative motion priors within our Dynamic SLAM framework. 
Furthermore, our future work will investigate trajectory forecasting algorithms that generate 3D predictions fully leveraging the $\SE$ output of DynoSAM.

\section{Limitations and Future Work}

Our proposed Dynamic SLAM method achieves high accuracy in both motion and object trajectory estimation, highlighting the crucial role of the back-end in generating precise and smooth object trajectories. While the front-end already operates close to real time, the back-end is currently solved in a full-batch manner and will be improved in future work to support incremental inference.

Currently, our system uses a simplistic approach that relies on a constant motion constraint to model changes in object motion.
As discussed in~\cite{judd2024ijrr_mvo}, this is akin to a constant velocity model, i.e. zero acceleration, 
which produces locally plausible motion but cannot capture motions of accelerating objects as accurately. 
Moving forward, we will exploring more sophisticated motion models, 
potentially through learning, 
which better capture an objects the short-term dynamics.

In addition, our system is currently designed to estimate the motion of rigid-bodies. While we have shown examples demonstrating its robustness in semi-rigid cases, a more comprehensive approach to non-rigid cases is highly relevant.

% \Jesse{no non-rigid bodies here? Maybe this is a better future work than the trajectory prediction stuff?}
% \Yiduo{Yes, definitely a better and more relevant future work.}

% The initial trajectory prediction results shown in~\secref{sec:object_trajectory_prediction} are promising; 
% therefore we plan to further integrate these predictions into our system. 
% Additionally, we aim to investigate how understanding the dynamics of each object can provide valuable feedback to enhance the SLAM pipeline and improve visual odometry.

Finally, we plan to leverage advanced transformer models such as SAM2~\cite{ravi2024sam2, yang2024samurai} and FlowFormer~\cite{huang2022flowformer} to improve scene understanding and explore new directions in Dynamic SLAM.
We have designed DynoSAM with an emphasis on modularity and extensibility to facilitate such further research. 

\section{Conclusion}
\label{sec:conclusion}
We have introduced DynoSAM, a cutting-edge, open-source framework for Dynamic SLAM that represents a significant advancement in the robotics field. By outlining the key theoretical concepts and formulations underpinning our approach, we provide a robust foundation for tackling the challenges of estimation in dynamic environments. DynoSAM offers a well-structured platform for implementing, testing, and evaluating Dynamic SLAM solutions, empowering researchers to develop and benchmark innovative methodologies with greater ease and precision. 
Importantly, our implementation is designed for flexibility, with clearly define interfaces between modules, facilitating integration with existing and new methods. This contribution paves the way for more reliable and adaptable robotic systems in dynamic and complex settings.

This paper thoroughly examines state-of-the-art methods for the Dynamic SLAM problem and introduces a novel formulation tailored for real-world applications. 
We highlight the importance of framing the problem in terms of observed motion, which enables accurate estimation and recovery of object trajectories. Additionally, we evaluate all the discussed formulations and demonstrate that our framework outperforms existing systems in both motion estimation and visual odometry, setting a new benchmark for Dynamic SLAM solutions.

The paper also demonstrates DynoSAM's effectiveness in downstream tasks such as motion prediction and 3D reconstruction. These capabilities collectively form the foundation for future dynamic object-aware navigation systems. With its user-friendly infrastructure and comprehensive evaluation suite, we aim for DynoSAM to serve as a robust platform for advancing research in Dynamic SLAM.

%%%%%%%%%%%%%%%%%%%%%%%%%%%%%%%%%%%%%%%%%%%%%%%%%%%%%%%%%%%%%%%%%%%%%%%%%%%%%%%%%%%%%%%%%%%%%%%%%%%%%%%%%%%%%%%

\section*{Acknowledgements}
\label{sec:acknowledge}
We would like to thank Kevin Judd and Jonathan Gammell for providing the results of MVO and their discussions. We additionally thank Tim Bailey for pointing out the scaling behaviour when evaluating world frame motion error.
% This research is funded with the support of ARIA Research and the Australian Government via the Department of Industry, Science, and Resources CRC-P program (CRCPXI000007).

% \section{Appendix}
\appendix
\label{sec:appendix}

\subsection{Rigid-body motion on points in the world frame}
\label{sec:app_rigid_body_on_point}

As discussed in the work of Zhang~\etal\cite{zhang2020vdoslam}, 
for any point on a rigid-body object measured in the world frame $\mpoint{\worldf}{k}$, 
there is the following equation: 
\begin{equation}
\begin{aligned}
    \mpoint{\worldf}{k} &= \objpose{\worldf}{k} \: \mpoint{\objf_k}{k} \\
    &= \objpose{\worldf}{k-1} \: \objmotion{\objf_{k-1}}{k-1}{k} \: \mpoint{\objf_k}{k}
\end{aligned}
\end{equation}
since $\objpose{\worldf}{k} = \objpose{\worldf}{k-1} \: \objmotion{\objf_{k-1}}{k-1}{k}$.

Based on the rigid-body assumption of this object, 
$\mpoint{\objf_k}{k}$ is time-invariant, 
and therefore $\mpoint{\objf}{} = \mpoint{\objf_k}{k} = \mpoint{\objf_{k-1}}{k-1}$: 
\begin{equation}
\begin{aligned}
    \mpoint{\worldf}{k} &= \objpose{\worldf}{k-1} \: \objmotion{\objf_{k-1}}{k-1}{k} \: \mpoint{\objf_{k-1}}{k-1} \\
    &= \objpose{\worldf}{k-1} \: \objmotion{\objf_{k-1}}{k-1}{k} \: \objpose{\worldf}{k-1}^{-1} \: \mpoint{\worldf}{k-1} \\
    &= \objmotion{\worldf}{k-1}{k} \: \mpoint{\worldf}{k-1}
\end{aligned}
\end{equation}
where $\objmotion{\worldf}{k-1}{k} := \objpose{\worldf}{k-1} \: \objmotion{\objf_{k-1}}{k-1}{k} \: \objpose{\worldf}{k-1}^{-1}$, 
an operation referred to as \textit{a frame change of a pose transformation} by Chirikjian~\etal\cite{Chirikjian17idetc}. 

% \subsubsection{Jacobians}
% \Jesse{also todo: Jacobians...}

\subsection{Constant motion model in different frames}
\label{sec:app_motion_model}
% \Jesse{is this saying what we want to say. Yes, but honestly very unclear...}
As discussed by Henein~\etal\cite{Henein20icra}, we can show that if the body-fixed frame pose change is constant then the absolute reference frame change is constant too:
\begin{equation}
\begin{aligned}
    \objmotion{\objpose{}{k-1}}{k-1}{k} &= \mathbf{C} =  \objmotion{\objpose{}{k}}{k}{k+1} \\
     \objmotion{\worldf}{k-1}{k}  &= \objpose{\worldf}{k-1} \: \mathbf{C} \: \objpose{\worldf}{k-1}^{-1} \\
      \objmotion{\worldf}{k}{k+1}  &= \objpose{\worldf}{k} \: \mathbf{C} \: \objpose{\worldf}{k}^{-1} \\
     \text{given that,} \quad
     \objpose{\worldf}{k} &= \objpose{\worldf}{k-1} \: \mathbf{C} \\
     \objmotion{\worldf}{k}{k+1} &= \objpose{\worldf}{k-1} \: \mathbf{C} \: \mathbf{C} \: \mathbf{C}^{-1} \: \objpose{\worldf}{k-1}^{-1} \\
     &= \objpose{\worldf}{k-1} \: \mathbf{C} \: \objpose{\worldf}{k-1}^{-1} \\
     &=  \objmotion{\worldf}{k-1}{k}
 \label{equ:constant_motion_in_world_proof}
\end{aligned}
\end{equation}
% This equations shows that when the change in local motion is constant, 
% the corresponding change in absolute motion is also constant. 

However, when the change in local motion is not constant, 
the change in absolute motion scales with distance the object is from $\{\worldf\}$.
This relationship is highly relevant for the object smoothing factor in~\eqref{equ:world_motion_smoothing_factor} which enforces a constant motion constraint in the $\{ \worldf \}$ frame.
Let the change in local motion be defined as:
\begin{equation}
    ^L\mathbf{\Delta} =  \objmotion{\objpose{}{k}}{k}{k+1}^{-1} \:  \objmotion{\objpose{}{k+1}}{k+1}{k+2} \in \SE
\end{equation}
and the change in absolute motion:
\begin{equation}
    ^W\mathbf{\Delta} = \objmotion{\worldf}{k}{k+1}^{-1} \: \objmotion{\worldf}{k+1}{k+2} \in \SE
\label{equ:absolute_motion_change}
\end{equation}
We can then say:
\begin{equation}
\begin{aligned}
     \objmotion{\worldf}{k}{k+1} &= \objpose{\worldf}{k+1} \: \objpose{\worldf}{k}^{-1} \\
     \objmotion{\worldf}{k+1}{k+2} &= \objpose{\worldf}{k+2} \: \objpose{\worldf}{k+1}^{-1} \\
     &= \objpose{\worldf}{k+1} \: \objpose{\worldf}{k}^{-1} \:  \objpose{\worldf}{k+1} \: ^L\mathbf{\Delta} \: \objpose{\worldf}{k+1}^{-1} \\
     &= \objmotion{\worldf}{k}{k+1} \: \objpose{\worldf}{k+1} \: ^L\mathbf{\Delta} \: \objpose{\worldf}{k+1}^{-1}
\label{equ:absolute_motion_interms_of_local_motion_change}
\end{aligned}
\end{equation}
Substituting~\eqref{equ:absolute_motion_interms_of_local_motion_change} into~\eqref{equ:absolute_motion_change} defines the relationship between the change in motion when represented in the local-body and world frames:
\begin{equation}
    ^W\mathbf{\Delta} = \objpose{\worldf}{k+1} \: ^L\mathbf{\Delta} \: \objpose{\worldf}{k+1}^{-1}
\label{equ:absolute_motion_change_interms_of_local_motion_change}
\end{equation}
% and shows that $^W\mathbf{C}$ scales with the object pose $\objpose{\worldf}{k+1}$. Expanding~\eqref{equ:absolute_motion_change_interms_of_local_motion_change} demonstrates $^W\mathbf{C}$ specifically scales with the translation part of the object pose:
% \begin{equation}
%     ^W\mathbf{C}_t = -^W\mathbf{C}_R \: ^Wt_{k+1} + ^W\mathbf{R}_{k+1} \: ^L\mathbf{C}_t + ^Wt_{k+1} 
% \end{equation}
% where $^W\mathbf{R}_{k+1}$ and $^Wt_{k+1}$ are the rotation and translation components of $\objpose{\worldf}{k+1}$ respectively. 
Consequently, the residual and covariance of the object smoothing factor will increase proportionally with the object's pose. To mitigate this effect for our smaller-scale experiments, we set the covariance of the smoothing factor to a sufficiently large value.
The same `scaling' behavior as represented by~\eqref{equ:absolute_motion_change_interms_of_local_motion_change} is also present when evaluating motion errors using $\objmotion{\worldf}{}{}$ instead of $\objmotion{\objf}{}{}$, 
as a small discrepancies in the local motion $\objmotion{\objf}{}{}$ can lead to a large error in $\objmotion{\worldf}{}{}$ due to~\eqref{equ:absolute_motion_change_interms_of_local_motion_change}. 

% \subsection{Camera Model}
% \label{sec:camera_model}
% A calibrated pinhole camera can be described by its intrinsic calibration matrix
% \[
% \mathbf{K} = 
% \begin{bmatrix}
%     f_u & 0 & u_0 & 0 \\
%     0 & f_v & v_0 & 0 \\
%     0 & 0 & 1 & 0
% \end{bmatrix}\text{,}
% \]
% where $f_u$ and $f_v$ are the horizontal and vertical focal lengths, and $u_0$ and $v_0$ define the principle point. The back-projection function, $\pi^{-1}$ obtains a 3D point $\zthreed$ in the camera frame given a measurement $\ztwod = \left[u, v\right]$ on the image plane and associated depth $d$:

% \[
% \begin{aligned}
%    \begin{bmatrix}
%        \zthreed \\
%        1
%    \end{bmatrix}
%    &= \mpoint{\camf_k}{k} = \pi^{-1}(\ztwod, d)
%    = 
%    \begin{bmatrix}
%        (u - u_0)\cdot d/f_u \\
%        (v - v_0)\cdot d/f_v \\
%        d \\
%        1
%    \end{bmatrix}
% \end{aligned}
% \]

%%%%%%%%%%%%%%%%%%%%%%%%%%%%%%%%%%%%%%%%%%%%%%%%%%%%%%%%%%%

\bibliographystyle{IEEEtran}
\bibliography{./IEEEabrv, ./refs/bibliography}

\vspace{-12mm}
\begin{IEEEbiography}[{\includegraphics[width=1in,height=1.25in,clip,keepaspectratio]{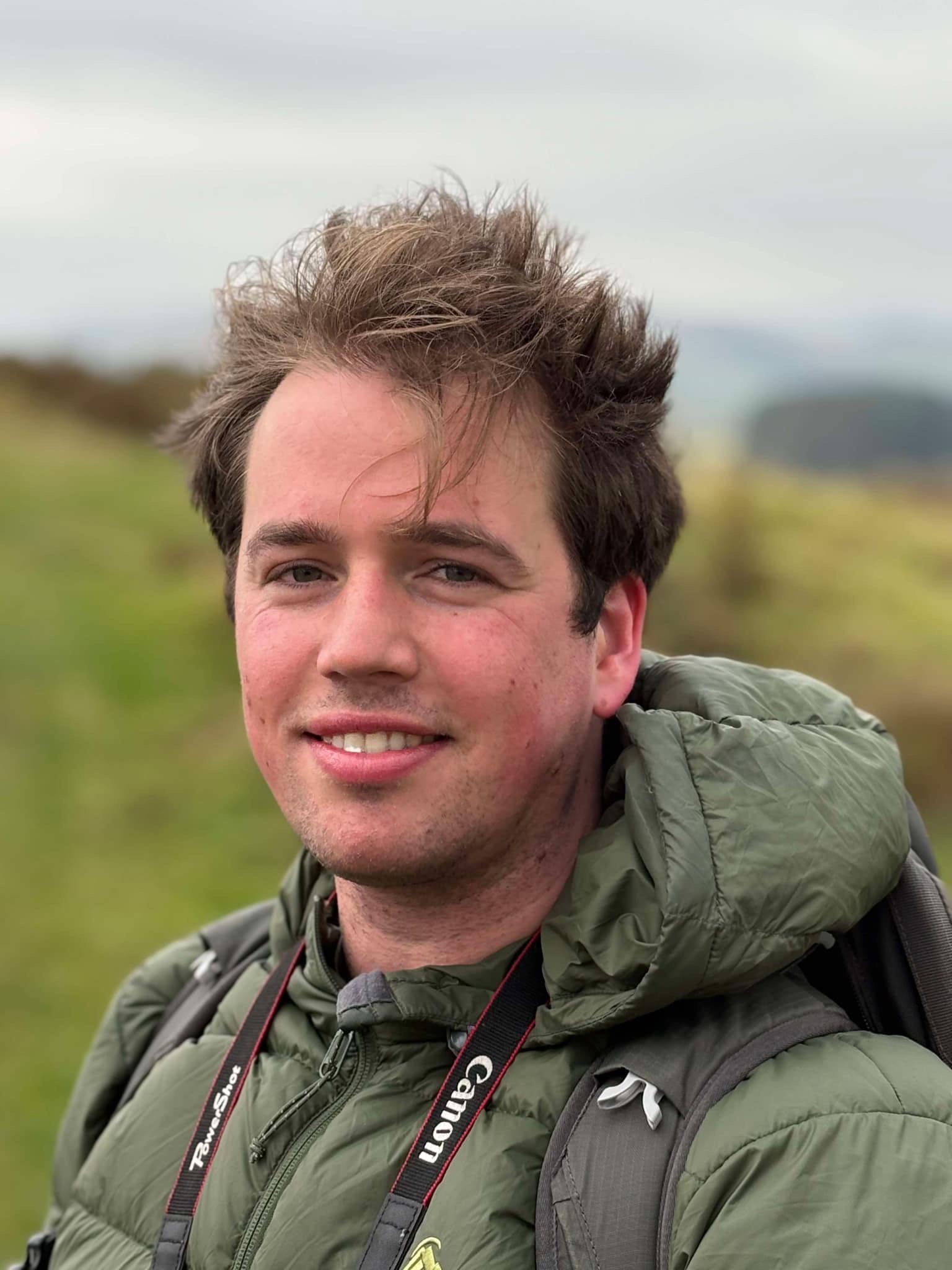}}]{Jesse Morris}
is a 3\textsuperscript{rd}\xspace year PhD student at the Australian Centre For Robotics (ACFR), supervised by Dr Viorela Ila. His thesis focuses on developing estimation frameworks for Dynamic SLAM. He received his bachelors degree in Mechatronic Engineering and Computer Science from the University of Sydney in 2022.
\vspace{-14mm}
\end{IEEEbiography}

\begin{IEEEbiography}[{\includegraphics[width=1in,height=1.1in,clip,keepaspectratio]{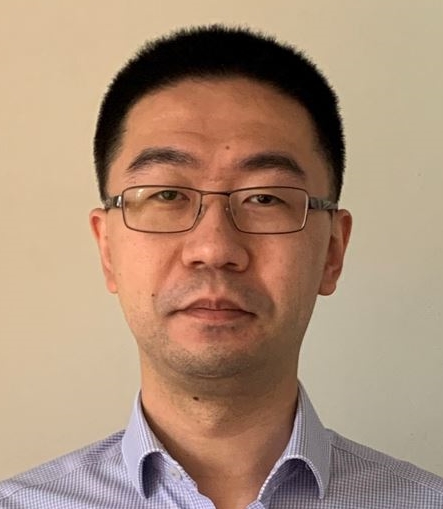}}]{Yiduo Wang}
is a postdoctoral researcher in the ACFR, working with Dr Viorela Ila. He focuses on SLAM, reconstruction and navigation in dynamic environments. He holds a DPhil Engineering Science degree awarded by University of Oxford, where his research focused on large-scale reconstruction. He also holds an MRes Robotics degree awarded by UCL on combining SLAM with semantic segmentation for dynamic environments. 
\vspace{-12mm}
\end{IEEEbiography}

\begin{IEEEbiography}[{\includegraphics[width=1in,height=1.1in,clip,keepaspectratio]{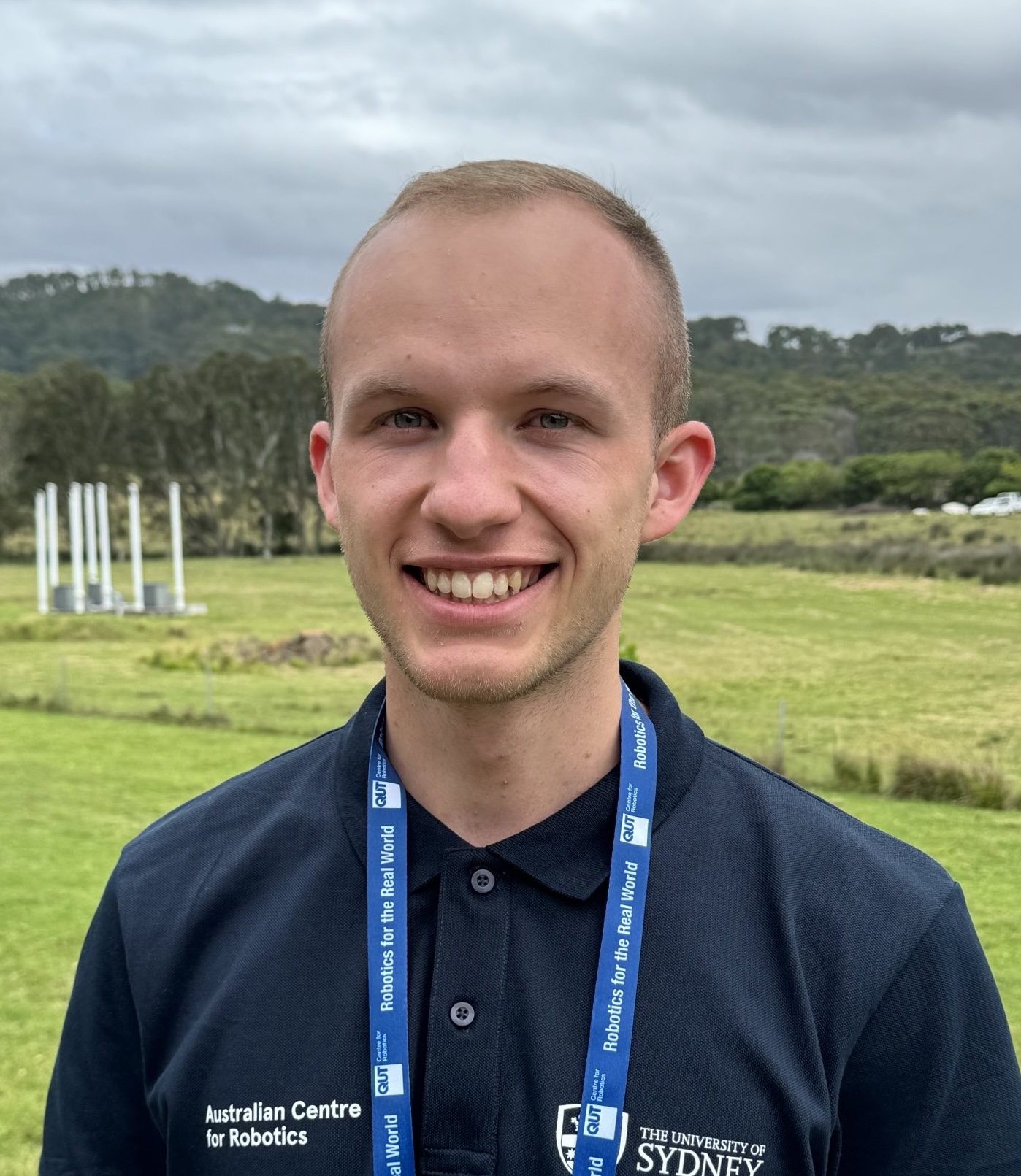}}]{Mikolaj Kliniewski}
is a 2\textsuperscript{nd}\xspace year PhD student at the Australian Centre For Robotics, supervised by Dr Viorela Ila. His thesis focuses on joint planning and estimation in dynamic environments. He received his Bachelor Honours degree in Computer Science from the University of Liverpool in 2023.
\vspace{-12mm}
\end{IEEEbiography}

\begin{IEEEbiography}[{\includegraphics[width=1in,height=1.1in,clip,keepaspectratio]{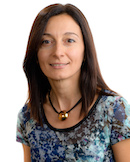}}]{Viorela Ila} received a Ph.D. in Information Technologies from the Universitat de Girona, Spain. She worked at the Institut de Robótica i Informàtica Industrial, Barcelona, and was awarded a MICINN/FULBRIGHT post-doctoral fellowship in 2009, joining Georgia Tech, USA. She joined the ROSACE project in LAAS-CNRS, France, in 2010, and was a Research Scientist at Brno University of Technology, Czech Republic (2012–2014) and a Research Fellow at the Australian National University (2015–2018). Currently a Senior Lecturer at the University of Sydney, her research focuses on robot vision, SLAM and 3D reconstruction, leveraging graphical models, optimization methods and information theory.
\end{IEEEbiography}

\end{document}